\PassOptionsToPackage{numbers,sort&compress}{natbib} 
\documentclass[preprint,a4paper]{elsarticle}
\usepackage{amssymb}
\usepackage{amsthm}
\usepackage{lineno}
\usepackage{amsmath,nccmath}
\usepackage{float}

\usepackage{mathtools}
\usepackage{multirow}
\newcommand{\Same}{\raisebox{1ex}{\(\downarrow\)}}

\usepackage{tikz}

\usepackage{url}
\usepackage{array}
\newcolumntype{P}[1]{>{\centering\arraybackslash}p{#1}}
\newcolumntype{M}[1]{>{\centering\arraybackslash}m{#1}}
\usepackage{subcaption}
\usepackage{placeins}

\usepackage{array,tabularx,booktabs,ragged2e}
\newcolumntype{L}[1]{>{\RaggedRight\arraybackslash}p{#1}}
\newcolumntype{C}[1]{>{\centering\arraybackslash}p{#1}}
\usepackage{booktabs}
\usepackage{subcaption}

\usepackage{graphicx}  

\usepackage[T1]{fontenc}

\journal{Mechanical Systems and Signal Processing}

\begin{document}	
	
\begin{frontmatter}
\title{Transfer learning via interpolating structures}
\author[add1]{T.A.\ Dardeno\corref{mycorrespondingauthor}}
\cortext[mycorrespondingauthor]{Corresponding author}
\ead{t.a.dardeno@sheffield.ac.uk}
\author[add1]{A.J.\ Hughes}
\author[add2]{L.A.\ Bull}
\author[add1]{R.S.\ Mills}
\author[add1]{N.\ Dervilis}
\author[add1]{K.~Worden}
\address[add1]{Dynamics Research Group, School of Mechanical, Aerospace and Civil Engineering, \\ University of Sheffield, Sheffield S1 3JD, UK}
\address[add2]{School of Mathematics and Statistics, University of Glasgow, Glasgow G12 8SQ, Scotland}

\begin{abstract}
Despite recent advances in population-based structural health monitoring (PBSHM), knowledge transfer between highly-disparate structures (i.e., heterogeneous populations) remains a challenge. The current work proposes that heterogeneous transfer may be accomplished via intermediate structures that bridge the gap in information between the structures of interest. A key aspect of the technique is the idea that by varying parameters such as material properties and geometry, one structure can be continuously morphed into another. The approach is demonstrated via a case study involving the parameterisation of (and transfer between) simulated heterogeneous bridge designs (Case 1). Transfer between simplified physical representations of a `bridge' and `aeroplane' is then demonstrated in Case 2, via a chain of finite-element models. The facetious question `When is a bridge not an aeroplane?' has been previously asked in the context of predicting positive transfer based on structural similarity. While the obvious answer to this question is `Always,' the results presented in the current paper show that, in some cases, positive transfer can indeed be achieved between highly-disparate systems.

\end{abstract}


\end{frontmatter}

\section{Introduction} \label{intro}
An exciting prospect for addressing the challenge of transfer between highly-heterogeneous structures involves leveraging the inherent geometry underlying the space of structures. Traditional linear machine-learning methods typically struggle with non-Euclidean data \cite{bronstein2017geometric, nonEuclideanML}, whereas geometric approaches \cite{gopalan2011domain,Boqing2012,masci2015geodesic,monti2017geometric,asif2021graph,simon2021learning} are well-suited for navigating the intricate, curved manifold structures of non-Euclidean spaces. In addition, in areas outside of Structural Health Monitoring (SHM), implementing intermediate steps in the transfer process has been shown to facilitate smoother transitions between vastly different domains or tasks \cite{gopalan2011domain,Boqing2012,rusu2022progressive,sagawa2024gradual,simon2021learning}.

To clarify; one of the ideas in population-based structural health monitoring (PBSHM) is that the structures of a given population can be expressed abstractly in the form of an attributed graph, which allows them to be embedded in a metric space -- a space of graphs \cite{Tsialiamanis2021}. For two structures $S$ and $S'$, given the metric-space structure, one can calculate a {\em distance} $d(S,S')$ between them. If the calculated distance were to be lower than some threshold $\epsilon_d$, the PBSHM framework would dictate that transfer may be attempted. 

An important issue, then, is how transfer can be achieved when the distance between two structures of interest is too large for positive transfer. Suppose the task is to transfer to a new structure $S$, which is data-poor, but there is no structure $S'$ in the current population for which $d(S,S') \le \epsilon_d$. Recall that PBSHM does not distinguish (in its representation space), between real structures and models; as such, a model {\em intermediate} structure $S^*$ may be constructed for which $d(S,S*) \le \epsilon_d$ and $d(S',S*) \le \epsilon_d$. In this situation, transfer may be accomplished in two steps; first from $S'$ to $S^*$ and then from $S^*$ to $S$. Furthermore, for large distances between $S$ and $S'$, multiple intermediate structures may be developed, to enable transfer via a greater number of steps. It is important to note that while transfer is carried out in the feature spaces of the structures, it can be argued that proximity in the structure space is equivalent to proximity in the data space \cite{Tsialiamanis2021}. In transfer-learning terms, the feature spaces of $S$ and $S'$ are the {\em target} and {\em source} domains, respectively. 

Transfer can be considered to be a map between data domains. Geodesic flows, \cite{gopalan2011domain,Boqing2012}, which are derived from differential geometry, identify the shortest path between two domains by leveraging the underlying geometry of the space. Gopalan \emph{et al.}\ \cite{gopalan2011domain} used a geodesic-flows approach in the context of unsupervised domain adaptation for object recognition, representing the source and target domains as subspaces on a Grassmann manifold. The approach in \cite{gopalan2011domain} is influenced by ideas from incremental learning \cite{schlimmer1986incremental}, and involves identifying potential intermediate domains between the source and target and using a finite number of these domains to learn domain transitions. Building upon the work in \cite{gopalan2011domain}, Gong \emph{et al.}\ \cite{Boqing2012} later introduced the geodesic flow kernel, which integrates an infinite series of subspaces along the flow, for improved domain-shift modelling.

In accordance with these principles, a \emph{heterogeneous} transfer approach for PBSHM is introduced herein. With two case studies, it is shown that transfer via intermediate structures can result in greater prediction accuracy compared to transferring directly between the source and desired target. For the examples presented, transfer learning along the chain is performed via statistical pre-processing/alignment with classification using support vector machines (SVM), first with a linear kernel, and then with the geodesic flow kernel \cite{Boqing2012}. For the first case study, highly-simplified and simulated two- and three-span finite element bridges provide the source and target domains, respectively, and demonstrate the technique for parameterisation of geometry and material properties. The second case study provides experimental validation for the approach, and involves transfer between physical representations of a cartoon `bridge' and `aeroplane', utilising interpolating finite-element models developed via parameterisation of geometry.

	\subsection{Research aims of the current work} \label{intro_aims}
	
	The aim of this work is to introduce and evaluate a novel transfer-learning framework that enables knowledge transfer between  heterogeneous systems. The framework integrates concepts from differential geometry and modern domain adaptation (including gradual and geometry-aware transfer) and is evaluated on heterogeneous structural systems within a PBSHM setting, with the goal of broader applicability.
	
	\vspace{0.5\baselineskip}
	The specific research aims are:
	
	\begin{enumerate}
		\item \emph{Develop a sequential transfer framework with intermediate structures/models.} Information is transferred step-by-step across intermediate domains, defined within parametric families of structures, creating a continuous path from source to target that enables transfer across markedly dissimilar structures. Label information is leveraged where available.
		
		\item \emph{Integrate statistical and geometric alignment methods into this framework.} Domain drift is first reduced via a statistical alignment step (e.g., scaling/standardisation, or baseline-informed alignment \cite{PooleNCA} when available). The data are then transformed via the geodesic flow kernel \cite{Boqing2012} to further reduce discrepancies between the domains. 
		
		\item \emph{Evaluate the framework using experimental and finite element (FEM) structures.} Both real test rigs and FEMs can be used as intermediate domains, including cases where FEMs represent parametric states that are not directly realisable but remain valid within a shared family.
	\end{enumerate}

	\subsection{Paper layout} \label{intro_layout}
	The layout of this paper is as follows. Section 2 discusses research related to the ideas presented in the current work. Section 3 highlights the novelty of the current work and its contribution to SHM and wider engineering applications. Section 4 provides an overview of the current work's theoretical basis, including a discussion of the geometric framework for PBSHM and the geodesic flow kernel. Sections 5 and 6 discuss the case studies demonstrating the proposed approach. In Section 5, transfer is demonstrated between simple, simulated heterogeneous bridges. An experimental case study is presented in Section 6, using simple physical representations of a `bridge' and `aeroplane' and finite-element models as interpolating structures. Lastly, conclusions are provided in Section 7.
	
\section{Related work} \label{related_work}
	\subsection{Transfer learning} 
	Machine-learning models use training data to learn data representations and optimise one or more tasks, with the goal of generalisation to unseen test data. The accuracy of these models is highly-dependent on the quantity and quality of the training set. However, fully-labelled real-world data are often difficult to obtain, because collection of high-quality data can be expensive and time consuming, or data may be inaccessible \cite{Alzubaidi2023Survey}. Inadequate training data can lead to poor model performance with poor generalisation and over-fitting to the training set.
	
	Transfer learning methods have been developed to address the data scarcity challenge, where the goal is to pre-train a model on a source domain and task, and use the resulting knowledge to improve performance on a target domain and task. Conventional transfer-learning approaches, where a model trained on one (source) domain is considered directly transferable to a similar (target) domain, only work well when the source and target have highly-similar probability distributions; otherwise, there is a significant risk of \emph{negative transfer}, where knowledge from a source domain or task results in worse performance on the target domain or task than would be achieved by training the target model independently \cite{Wang2019}. 
	
	Transfer learning is becoming an increasingly important tool in SHM and PBSHM. In SHM, transfer learning has been used to address variability in environmental and operational conditions, and in PBSHM, it facilitates information transfer between structures in the presence of missing data and/or labels \cite{Worden2020brief}.

		\subsubsection{Domain adaptation}
		
		In general, domain adaptation (DA) aims to achieve positive transfer by minimising the difference between domain distributions. DA can be categorised according to the availability of target-domain labels: in supervised DA, a small amount of labelled target data is available; in semi-supervised DA, this is complemented by unlabelled target data; and in unsupervised DA (UDA), which is the most widely studied, no target labels are available. Many DA methods are additionally transductive, in that unlabelled target samples are available during adaptation and are used to estimate target-domain structure (e.g., subspaces, moments, or discrepancy measures). Formally, a domain is defined by an input or feature space $\mathcal{X}$, an output or label space $\mathcal{Y}$, and an associated probability distribution $p(x,y)$ \cite{farahani2020brief}. Here, $\mathcal{X}$ is a subset of a $d$-dimensional space, $\mathcal{X} \subset \mathbb{R}^d$, and $\mathcal{Y}$ is either binary ${-1,+1}$ or multi-class ${1,\ldots,K}$, and $p(x,y)$ is the joint probability distribution over $\mathcal{X} \times \mathcal{Y}$. This joint distribution can be decomposed as,
		
		\begin{equation}
			p(x,y) = p(x)\,p(y\mid x)
		\end{equation}
		or, equivalently,
		\begin{equation}
			p(x,y) = p(y)\,p(x\mid y),
		\end{equation}
	
		\noindent where $p(\cdot)$ denotes a marginal distribution and $p(\cdot\mid\cdot)$ a conditional distribution.
		
		Early advancements in DA focussed on statistical alignment under covariate shift, which occurs when marginal probability distributions differ but conditional distributions remain constant across domains, i.e.,
		
		\begin{equation}
			p_{s}(x) \neq p_{t}(x), \qquad p_{s}(y \mid x) = p_{t}(y \mid x).
		\end{equation}
	
		A common strategy in this setting is instance weighting, where training samples are re-weighted so that the effective source distribution better matches the target. Representative approaches include re-weighting likelihood functions \cite{shimodaira2000covariate}, direct density-ratio estimation \cite{Sugiyama2007}, and kernel-based matching of covariate distributions in reproducing-kernel Hilbert space (RKHS) \cite{gretton2009covariate}. Closely-related feature-based methods map source and target data to a space where a discrepancy measure (often maximum mean discrepancy (MMD)) is minimised, e.g., transfer component analysis (TCA) and its extensions for joint alignment \cite{pan2009tca,long2013jda,long2013adaptation}. Other classical approaches target specific distributional structure, such as aligning second-order statistics (e.g., CORAL) \cite{Sun2017CORAL}. In parallel, DA has also been formulated as an optimal transport (OT) problem, where a cost-minimising transport plan aligns samples across domains \cite{Courty2017}; although often considered under the DA umbrella, OT more fundamentally reflects the geometry of probability distributions, and will be revisited below in the context of geometric approaches to transfer learning.
		
		More recent work has extended these ideas in several directions. Deep learning approaches integrate distribution alignment into a neural network, including moment-based alignment objectives \cite{Long2015,Long2017}, adversarial learning of domain-invariant features \cite{Ganin2016,Tzeng2017}, and deep variants of classical statistical matching (e.g., covariance and higher-order moment matching) \cite{Sun2016,Zellinger2017}. Beyond these standard templates, alternative objectives have also been explored (e.g., covariance-based visual alignment, distributed balancing for multi-site healthcare data, and entropy-based alignment strategies) \cite{Karn2023,Tong2025,Perez2025}.
		
		Another family of DA approaches focusses on exploiting the structure of unlabelled target data with self-training. In this setting, the model iteratively generates pseudo-labels for unlabelled samples using its own predictions, retrains on the expanded labelled set, and progressively adapts to the target distribution \cite{Yarowsky1995,Nigam2000}. While conceptually related to the idea of leveraging predicted labels, the transfer approach proposed herein differs in that pseudo-labels are propagated across domains in a sequential manner, rather than within a single target domain.
		
		In recent years, DA has been used to facilitate transfer for SHM applications. Gardner \emph{et al.}\ demonstrated the applicability of classical DA methods for transfer between structures on numerical and experimental case studies \cite{Gardner2020da}, and Poole \emph{et al.}\ introduced Normal Condition Alignment (NCA) to mitigate environmental and operational variability by aligning condition statistics \cite{PooleNCA}. Complementary SHM work has incorporated deep transfer learning to learn transferable representations from vibration data \cite{Azimi2020structural} and generative domain adaptation to align simulated and experimental time-series responses \cite{Ge2024domain}. Other recent studies have explored population-based transfer mechanisms for damage detection, including clustering based on principal angles within federated learning settings \cite{Cheema2025}.

		\subsubsection{Gradual domain adaptation}
		
		Direct adaptation is often successful with small distribution shifts; however, when the distance between the source and target is large, features that are predictive for the source task are often not suitable for the target task, which can result in negative transfer \cite{Wang2019}. Gradual domain adaptation (GDA) \cite{He2024} addresses this challenge by introducing intermediate domains that bridge the gap between the source and target, allowing knowledge to be transferred incrementally across domains of increasing similarity. In the standard GDA setting, intermediate domains are assumed to be observed rather than constructed. In addition to labelled source data and unlabelled target data, unlabelled datasets from one or more intermediate domains are available. This idea has been implemented in several ways. For example, gradual self-training uses pseudo-labels generated at one step to supervise the next, thereby propagating knowledge forward until the target is reached \cite{Zou2019confidence, kumar2020gda}, whereas other approaches frame the problem as a curriculum, and the model progresses from easier to harder tasks (or increasingly similar domains) \cite{Zhang2017curriculum}. More recent work has combined adversarial learning with GDA, showing that intermediate domains can stabilise adversarial alignment \cite{Wulfmeier2018incremental}.
		
		Recent formalisation of GDA clarifies why dividing a large distribution shift into a multi-step problem can facilitate transfer. In gradual self-training, an important factor is the amount of shift that is accumulated along the chain, and how pseudo-label errors propagate \cite{He2024, Wang2022GDA}. Let $\mu_0,\mu_1,\dots,\mu_T$ denote the sequence of domain distributions over the input space $\mathcal{X}$, where $\mu_0$ is the source and $\mu_T$ is the target \cite{He2024}. The total shift in distribution space can be quantified via the $p$-Wasserstein distance $W_p$, 
		
		\begin{equation}
			L_{\mathrm{path}}=\sum_{t=1}^{T} W_p(\mu_{t-1},\mu_t)
			\label{eq:gda-path-length}
		\end{equation}
	
		\noindent This distance provides validation for the heuristic: GDA tends to result in accurate transfer when adjacent domains are close \cite{He2024, Wang2022GDA}. Because $W_p$ is a metric, the inequality,
		
		\begin{equation}
			W_p(\mu_0,\mu_T)\le L_{\mathrm{path}}
			\label{eq:gda-triangle}
		\end{equation}

		\noindent holds, indicating that if the chain of intermediate domains deviates from the optimal path, $L_{\mathrm{path}}$ increases. This motivates prioritising intermediates that minimise $L_{\mathrm{path}}$ \cite{He2024, Wang2022GDA}.
		
		When observed intermediate domains are sparse, intermediate states can instead be artificially generated. In \cite{Abnar2021GDA}, virtual samples are created from intermediate distributions by interpolating representations of examples from source and target domains. In \cite{sagawa2024gradual}, normalising flows are integrated into a GDA framework to allow interpolation when intermediate domains are sparse. Similarly, in \cite{He2024}, intermediate domains are generated along an optimal-transport path. In \cite{Zhuang2024GDA}, samples and labels are transported along a path from source to target using a Wasserstein gradient-flow. 
		
		Recent GDA formalisation makes the path viewpoint explicit by treating domains as points in a space of distributions and measuring the cumulative shift along a chain, which naturally motivates a geometric interpretation of adaptation as movement along a (near-)geodesic trajectory. In this sense, several transport- and flow-based GDA methods (e.g., \cite{gopalan2011domain,Boqing2012}) have considerable conceptual overlap with geometry-based transfer.

		\subsubsection{Geometric and topological approaches to transfer learning}
		
		Some transfer learning approaches combine statistical and geometric perspectives, treating domains as manifolds or points on a manifold and aligning their structure to enable transfer. Many of these methods can be understood within the DA framework, where geometry provides a means of interpolating or aligning source and target domains. Other methods extend more broadly, using geometry to guide representation learning or to capture invariant features across domains. This viewpoint also overlaps with gradual domain adaptation (GDA): when intermediate domains are generated or selected to lie along a smooth trajectory from source to target, the adaptation process can be interpreted as following a path on an underlying manifold of domains.
		
		Early work focussed on representing domains as linear subspaces treated as points on the Grassmann manifold. The Grassmann manifold $\mathrm{Gr}(d,D)$ is the space of all $d$-dimensional linear subspaces of the $D$-dimensional Euclidean space $\mathbb{R}^D$. In practice, each domain is first reduced to a subspace, often using Principal Component Analysis (PCA), and that subspace is treated as a single point on the manifold. Gopalan \emph{et al.}\ \cite{gopalan2011domain} introduced an approach, which, after projection to PCA space, involves interpolating between source and target by sampling intermediate subspaces along the geodesic between them. The geodesic flow kernel (GFK) \cite{Boqing2012} extended this idea by integrating over infinitely many intermediate subspaces to obtain a closed-form kernel. Arguing that constraining the path to the geodesic may be overly restrictive, particularly when multiple and diverse sources of variation are present, Caseiro \emph{et al.}\ \cite{Caseiro2015} proposed Subspaces by Sampling Spline Flow (SSF), which generalised geodesics to smooth spline curves on the Grassmann manifold via rolling maps, enabling nonlinear interpolation between domains. Subspace Alignment \cite{fernando2013unsupervised} took a different approach, directly learning a linear mapping to reduce the discrepancy between source and target subspaces. Hybrid methods have been developed to address geometric and statistical shifts simultaneously, for example, Joint Geometrical and Statistical Alignment (JGSA) \cite{zhang2017joint}, and Manifold Embedded Joint Geometrical and Statistical Alignment (MEJGSA) \cite{sanodiya2022manifold}.
			
		Geometry-based methods have also been applied beyond Grassmann manifolds, particularly on the manifold of symmetric positive definite (SPD) matrices. Covariance descriptors (i.e., covariance matrices used as features) are widely used in pattern recognition, neuroscience, and SHM. When full-rank, they naturally lie on the SPD manifold because they are symmetric and positive definite. Treating them in Euclidean space distorts their geometry, motivating approaches that respect the SPD structure. Yair \emph{et al.}\ \cite{Yair2019parallel} proposed a transfer framework based on parallel transport, which maps features from different domains to a shared reference point in a way that preserves the SPD manifold geometry, helping to retain the information encoded in their covariance structure during alignment. More recently, Ju and Guan \cite{Ju2025deep} introduced a framework that combines a deep SPD network with losses based on optimal transport. In the context of electroencephalogram (EEG) image classification, they found that their method improved transfer performance between different EEG recording sessions.
		
		In addition to geometric methods, transfer learning has also been informed by topological data analysis, where the focus shifts from distances and alignments to the shape and connectivity of data. Weeks and Rivera \cite{Weeks2021domain} applied persistent homology within a domain-adversarial framework, investigating whether aligning topological persistence diagrams across domains, rather than only aligning feature distributions, can improve transfer performance. More recently, a homology consistency constraint has been used to maintain topological structure during fine-tuning of vision-language models, supporting better generalisation in few-shot and domain generalisation scenarios \cite{Zhang2024homology}.
		
		Transfer motivated by geometric structure has been formalised in PBSHM. Tsialiamanis \emph{et al.}\ \cite{Tsialiamanis2021} modelled populations as spaces of structures with associated feature bundles, where each structure corresponds to a point on the base space (often represented as an attributed graph derived from an irreducible element model) and fibres are formed from the diagnostic features of each structure. This framework uses concepts from fibre bundles and gauge theory to represent environmental variability and enables transfer across both homogeneous and mildly heterogeneous populations, with graph neural networks proposed to learn consistent feature mappings. A limitation of representing structures as isolated points in a graph-based space is that, while distances between graphs can be defined, this does not in itself provide a useful notion of continuity for structures in a population. In the present work, a topological view is instead adopted, in which the base space is described by parametric families (open sets) to enable continuous variation between fibres and motivate transfer as local transport between neighbouring fibres \cite{WordenISMA}. In SHM transfer under domain shift, the GFK has previously been combined with statistical alignment methods (in this case, NCA), in earlier conference publications related to this work \cite{DardenoEWSHM, DardenoISMA} (and has been integrated with statistical matching and Bayesian inference for corrosion-state estimation in \cite{JIA2026102672}, in an SHM setting). In the present work, the principal contribution is the construction and exploitation of substantiated intermediate domains: intermediate steps correspond to actual structures or numerical models generated within a parameterised configuration space (geometry, material properties, support conditions, etc.). This strategy yields a sequential transfer pathway in which each hop is physically interpretable and corresponds to a controlled change in the underlying structure, enabling transfer across highly-heterogeneous structures by decomposing a large domain shift into a series of smaller, learnable shifts.

\section{Novelty and contribution} \label{novelty}

This work proposes a physically-grounded sequential transfer framework for PBSHM, where information is propagated along a chain of intermediate structures defined in configuration space. Each intermediate corresponds to a specific \emph{state of an engineered system} drawn from a shared parametric family, constructed using real structures and/or finite element models (FEMs). Endpoint FEMs can be validated against measurements, while intermediate FEMs can be introduced to maintain continuity even when a directly realisable intermediate does not exist. For example, when transferring between a two-support bridge and a three-support bridge, a third support can be introduced gradually by varying stiffness and mass parameters; intermediate FEM states need not correspond to a realisable structure at every step, but they remain well-defined within the modelling assumptions and are mechanically interpretable.

The proposed transfer approach shares with GDA the principle of using intermediate domains, but differs in what constitutes an intermediate and how the chain is constructed. In standard GDA, intermediate domains are typically assumed to be observed, meaning that data are available from a sequence of intermediate environments between source and target \cite{He2024,Wang2022GDA}. When such intermediates are sparse or absent, recent methods construct virtual intermediates using representation interpolation \cite{Abnar2021GDA}, normalising flows \cite{sagawa2024gradual}, optimal transport \cite{He2024}, or Wasserstein gradient flow \cite{Zhuang2024GDA}. In contrast, the present work specifies the chain in \emph{configuration space}, so that intermediates correspond to \emph{engineered system states} (real or model-based) rather than defining intermediates purely in data or representation space. Likewise, the proposed approach differs from geometry-based transfer methods in how geometric tools are used. These approaches often construct abstract intermediate objects (e.g., a geodesic flow between subspaces) and apply them directly between a single source and target. Here, geometric alignment (via the GFK) is applied locally between consecutive (physically-grounded) intermediate domains.

The framework is designed for SHM settings where labels are scarce, i.e., a fully-labelled source is available and the target and intermediate domains may be only partially labelled (often with limited or no damage labels). The method does not require labelled data beyond the source; however, when baseline (healthy) measurements are available in an intermediate or target domain, they can be exploited for statistical normalisation, calibration, or consistency checks. 

Relative to other transfer approaches in the PBSHM framework, the novelty is a chain-based mechanism for handling \emph{highly-heterogeneous} transfer. Gardner \emph{et al.}\ \cite{Gardner2020da} considered both mildly-dissimilar and topologically-dissimilar structures, and showed that, in some cases, methods such as TCA, JDA, and ARTL can transfer successfully. In contrast, the present work targets settings where the source and target are sufficiently far apart that a single adaptation step is unreliable. 

Poole \emph{et al.}\ \cite{PooleNCA} introduced Normal-Condition Alignment (NCA), a statistical alignment step that estimates an affine normalisation using baseline data, motivated by class imbalance and partial DA issues that arise in SHM. In the present work, a statistical alignment step is applied at each hop to improve comparability across consecutive domains; using baseline-only information in the normalisation (i.e., NCA) is one practical option when baseline measurements are available. In \cite{Tsialiamanis2021}, PBSHM was formalised within a fibre-bundle framework, treating each structure as a point on the base space with an associated feature fibre, and used graph neural networks to learn mappings across populations. 

In the current work, structures on the PBSHM base space are interpreted as belonging to parametric families, so that local neighbourhoods (e.g., open balls under an appropriate metric) can be associated with each structure and used to motivate continuity between systems. Intermediate states are instantiated within such families so that transfer can be performed sequentially between adjacent, closely-related domains.

\vspace{0.5\baselineskip}
In summary, the contributions of this work are:
\begin{enumerate}
	
	\item \textbf{Intermediate domains defined in configuration space:} This work introduces a physically-grounded notion of intermediate domains as \emph{engineered system states} defined in configuration space within a shared parametric family, realised via real structures and/or FEMs, including transitional FEM states when directly realisable intermediates do not exist.
	
	\item \textbf{Sequential statistic and geometric alignment:} This work proposes a \emph{multi-hop} transfer approach that composes consecutive GFK alignments at each hop, replacing unreliable single-step transfer when structural mismatch is large.
	
	\item \textbf{Operationalising continuity in PBSHM framework:} This work operationalises continuity between structures by chaining overlapping neighbourhoods in the PBSHM base space to form a continuous parametric path for sequential adaptation; while the current application is PBSHM, the resulting framework applies to any population that can be parameterised in a mechanically meaningful way.
	
\end{enumerate}

It should be noted that the focus of this work is on demonstrating the intermediate-structures approach, rather than optimising the domain adaptation tool applied at each hop. More sophisticated alignment methods could be substituted within the same framework when appropriate.

\section{Background theory} \label{theory}
This section describes the extended geometric framework for PBSHM that has enabled the development of the proposed transfer approach. Also included in this section is a brief introduction to the geodesic flow kernel and the associated equations required for its formulation; for a more comprehensive discussion of geodesic flows and the geodesic flow kernel, interested readers are directed to \cite{gopalan2011domain,Boqing2012}.	

	\subsection{A geometric framework for PBSHM} \label{PBSHM}
	
\begin{figure}[h!]
		\vspace{0.5cm}
		\centering
		\includegraphics[width=0.95\textwidth]{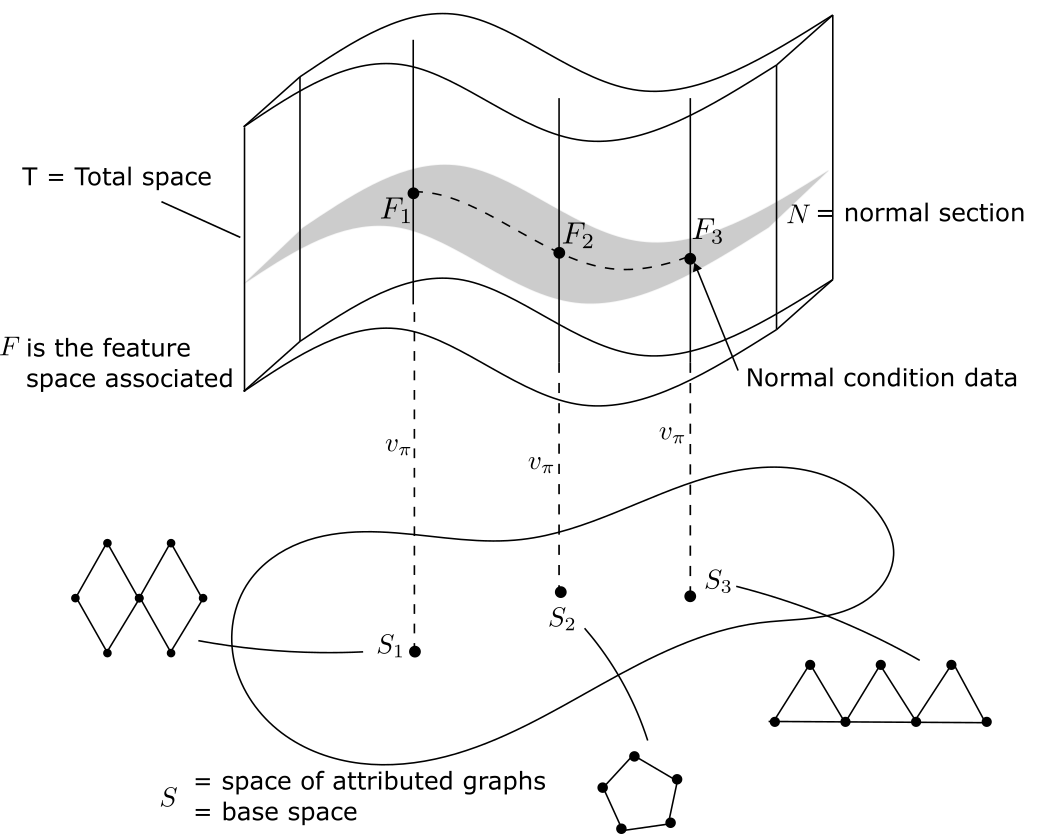}
		\caption{Schematic of geometrical model of PBSHM in terms of a fibre bundle.}
		\label{fig:FB}
	\end{figure}

	In the `foundations' series of papers, which laid out the basis for a population-based approach to SHM, the paper \cite{Tsialiamanis2021} proposed a geometrical model of the theory in which two spaces featured prominently. The first space contained abstract representations of the structures of interest in the population, while the second aggregated all of the structure feature spaces which would contain the SHM data; the two spaces to be linked by a projection function and sections which mapped between the structures and their individual data sets. This picture is shown in Figure \ref{fig:FB}. 
	
	The geometry here strongly resembled that of a fibre bundle; a construct from differential geometry that had earned an important place in modern theoretical physics. In the terminology of fibre bundles, the space of structures was the {\em base space} $S$, while the feature spaces combined to form the {\em total space} $T$. A projection operator from $T$ to $S$ served to associate data with specific structures. This picture motivated discussion of several important features of PBSHM, including the similarity measures on $S$ and transfer learning. Assignments of specific data points to structures was made meaningful by the use of {\em sections} of the bundle, with the most important being the {\em zero section}, which assigned normal-condition data to the structure. Even the idea of environmental and operational variations could be accommodated in the picture. The problem -- and a serious one at that -- was that the geometrical viewpoint could not be made mathematically rigorous. The obstruction to a mathematical formulation resulted from the fact that the base space was not a topological space (although it was a metric space); it was composed of a discrete set of points with no notion of continuity between points. In most applications of fibre bundles, both the base and total spaces were actually assumed to be {\em differentiable manifolds}, with smoothness properties going beyond continuity. This problem rendered the bundle picture unsuitable for any analysis beyond analogy. 
	
	Despite the clear problem with the fibre bundle picture, the strength of the analogy suggested that some rigorous form of geometrical model might well be fruitful if only the main objection could be removed. Two ways to solve this problem suggested themselves:
	
	\begin{enumerate}
		\item Modify the geometrical object of interest -- in this case a fibre bundle -- to some other mathematical object, where the graph-like nature of the base space is not an obstacle to
		rigour. 
		\item Modify the base space to allow some notion of continuity and thus of topology.
	\end{enumerate}  
	
	The first option is explored in separate work. The second option is the focus of the current paper, and mathematical formalism for both options will be addressed in forthcoming papers (see also \cite{WordenISMA}). One realisation of the second option could be to embed the space of structures into a continuous space of graph embeddings \cite{xu2021understanding}, which would confer topological structure on the base space. However, this approach would not guarantee that points along a path in the embedding space correspond to physically-meaningful intermediates. Here, continuity is instead introduced via parametric families of structures. The idea is fairly simple; although individual structures usually exist in isolation, one can regard them as representing specific choices of building blocks, each contingent on geometrical and material parameters. By allowing continuous variations in the parameters, one can then construct families of abstract `deformations' of the original structure. These `parametric families' -- as mentioned earlier -- allow a continuous open ball of structures to be built around each structure in the original population; if the ranges of the parameters are made large enough, the open balls overlap and extend a local notion of continuity to the entire base space. This is the strategy adopted in this paper. 
	
	One benefit of the approach is crucial for the analysis in this paper. In the original graph space, while there is a notion of distance between the points, {\em there are no points between the points}; there are no places for intermediate structures. The parametric families overcome this problem; furthermore, they allow more useful structure. In particular, the families allow the construction of continuous paths between structures and thus the specification of a shortest path -- a {\em geodesic}; the idea of a geodesic will be fundamental in the analysis to follow.

	\subsection{Geodesics and geodesic flows} \label{geodesics}
	In flat Euclidean space, the  `acceleration'\footnote{In this context, acceleration need not be the second derivative of the displacement with respect to time. More generally, it denotes the rate of change of the tangent vector (i.e., the direction of travel) along the curve. The kinematic terminology is inherited from classical mechanics, where this mathematical machinery was originally developed to describe particle trajectories; when the geometry was later abstracted to Riemannian manifolds, the language carried over \cite{Arnold1989}.} of a particle moving along a path is the second derivative of its position, and a straight line is the path along which this acceleration is zero. On a curved space (i.e., a Riemannian manifold), which is equipped with a metric that defines distances and angles at every point, this straightforward notion of acceleration breaks down because the tangent spaces at different points along the path are distinct; there is no natural way to compare or subtract velocity vectors at different points \cite{Chavel}. This issue is addressed by the \emph{Levi-Civita connection}, the unique connection compatible with the metric of the manifold, which provides a consistent rule for differentiating vector fields along a path, thus enabling the acceleration of a curve to be defined as $D_t\dot{\gamma}$, the covariant derivative of the velocity vector along the curve \cite{Lee}. A geodesic is a curve whose acceleration vanishes, $D_t\dot{\gamma} \equiv 0$. In Euclidean space, $D_t$ reduces to the ordinary time derivative and the geodesic reduces to a straight line. A consequence of this definition is that geodesics are locally length-minimising. The geodesic is uniquely determined by an initial point and initial tangent direction, providing a canonical path on the manifold \cite{Chavel}.

	\begin{figure}[h!]
		\centering
		\includegraphics[width=0.65\textwidth, trim=100pt 100pt 100pt 100pt, clip]{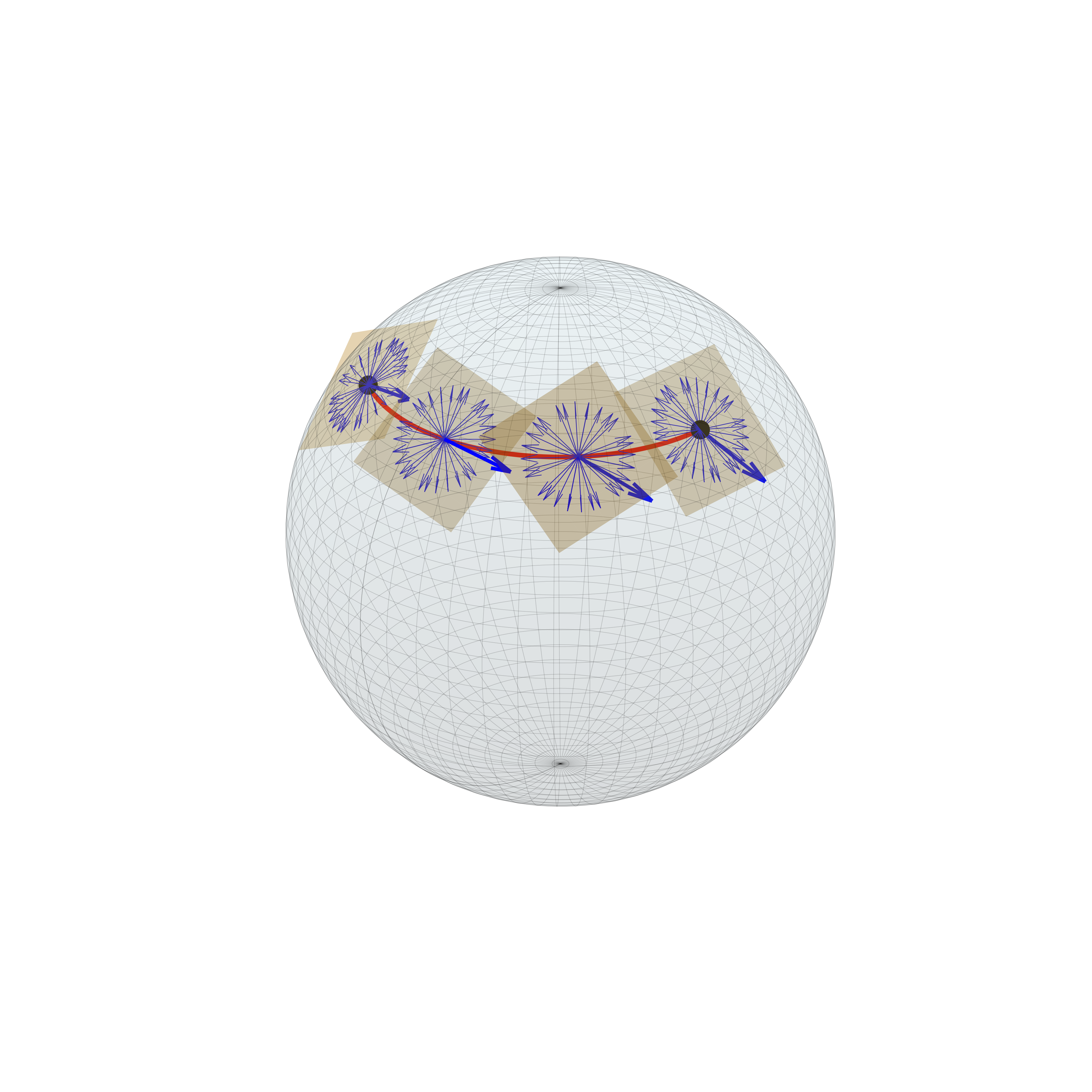}
		\caption{A geodesic (red curve) on a sphere, with tangent spaces (planes) shown at successive points along the path. The blue arrows show the tangent directions evolving continuously along the geodesic, illustrating the geodesic flow.}
		\label{fig:geodesic}
	\end{figure}
	
	The exponential map at a given point $p$ maps each tangent vector $\xi$ to the point on the manifold reached by travelling along the geodesic from $p$ in direction $\xi$; equivalently, straight lines through the origin of the tangent space correspond to geodesics on the manifold emanating from $p$ \cite{Chavel}. Geodesic flows occur in the tangent bundle of a manifold, where the tangent bundle can be understood as the collection of all tangent spaces at every point on the manifold. Each tangent space represents all possible directions and velocities of motion at a specific point, providing a way to describe how motion evolves in the context of the manifold's geometry. In other words, the geodesic flow describes how a point on the manifold and its associated tangent vector evolve continuously over time \cite{Chavel}. This is illustrated in Figure \ref{fig:geodesic} above, which shows the evolution of tangent spaces along a geodesic on a sphere. Specifically, the red curve is the geodesic, the planes are the tangent spaces at successive points, and the blue arrows show the corresponding tangent directions evolving continuously along the path. A point in the tangent bundle specifies both a location on the manifold and a direction of travel; the geodesic flow carries this point continuously along the corresponding geodesic, sweeping out a smooth trajectory parameterised by $t \in [0,1]$ from a source to a target. This continuous flow of intermediate domains is exploited by the geodesic flow kernel (GFK) introduced in Section \ref{GFK}. Furthermore, it provides a conceptual parallel to the intermediate structures framework introduced in the current work. 
		
	\subsection{Geodesic flow kernel (GFK)} \label{GFK}
	The geodesic flow kernel (GFK) \cite{Boqing2012}, characterises incremental changes in geometrical and statistical properties between the source and target domains via integration of all subspaces along the flow. To construct the kernel, PCA subspaces are computed and their appropriate dimensionality determined. The principal angles of the subspaces are then used to develop the geodesic flow. The geodesic flow kernel is then constructed and embedded into a kernel-based classifier \cite{Boqing2012}. 
	
	Let $ \mathbb{G}( \text{d}, \text{D} ) $ represent the Grassmann manifold, which is the collection of all $d$-dimensional subspaces of $\mathbb{R}^{\text{D}}$. Let $\boldsymbol{S}_1,\boldsymbol{S}_2 \in \mathbb{R}^{\text{D} \times \text{d}} $ signify the principal component analysis (PCA) \cite{PCA1987} bases of the source and target data, respectively. Then, let $\boldsymbol{R}_1 \in \mathbb{R}^{\text{D} \times (\text{D}-\text{d})}$ and  $\boldsymbol{Q} \in \mathbb{R}^{\text{D} \times \text{D}} $ define the orthogonal complement and orthogonal completion of $\boldsymbol{S}_1$, respectively. The cosine-sine decomposition of $\boldsymbol{Q} ^{\intercal} \boldsymbol{S}_2$ is given by,
	
	\begin{equation}
		\boldsymbol{Q}^{\intercal} \boldsymbol{S}_2 = \begin{bmatrix}
			\boldsymbol{V}_1 & 0 \\
			0 & \boldsymbol{\tilde{V}}_2 
		\end{bmatrix}
		\begin{bmatrix}
			\bold{\Gamma} \\
			-\bold{\Sigma} 
		\end{bmatrix}
		\boldsymbol{V}^{\intercal}
	\end{equation}
	
	\noindent 
	where $\boldsymbol{V}_1$, $\boldsymbol{\tilde{V}}_2$, and $\boldsymbol{V}$ are orthogonal matrices that rotate/align the subspaces onto a common basis, such that $\boldsymbol{S}_1^{\intercal} \boldsymbol{S}_2 = \boldsymbol{V}_1 \bold{\Gamma} \boldsymbol{V}^{\intercal}$ and $\boldsymbol{R}_1^{\intercal} \boldsymbol{S}_2 = -\boldsymbol{V}_2 \bold{\Sigma} \boldsymbol{V}^{\intercal}$ \cite{gopalan2011domain,Boqing2012}. The arccosine and arcsine of matrices $\bold{\Gamma}$ and $\bold{\Sigma}$ are used to compute the principal angles, ${\theta}$, respectively \cite{gopalan2011domain,Boqing2012}, which are then used to develop the geodesic flow. Via the canonical Euclidean metric on the Riemannian manifold, the geodesic flow is parameterised as $ \bold{\Phi} : t \in [0,1] \rightarrow \bold{\Phi}(t) \in \mathbb{G}(\text{d},\text{D})$, with the constraints that $ \bold{\Phi}(0) = \boldsymbol{S}_1$ and  $\bold{\Phi}(1) = \boldsymbol{S}_2$ \cite{gopalan2011domain,Boqing2012}. For other $t$, $\bold{\Phi}(t)$ can be given as \cite{gopalan2011domain,Boqing2012},
	
	\begin{equation}
		\bold{\Phi}(t) = \boldsymbol{Q} \begin{bmatrix}
			\boldsymbol{V}_1 \bold{\Gamma}(t) \\
			-\boldsymbol{\tilde{V}}_2\bold{\Sigma}(t)
		\end{bmatrix}
	\end{equation}
	
	Now, assume two $D$-dimensional feature vectors $\boldsymbol{x}_i$ and $\boldsymbol{x}_j$, whose projections into the space defined by $\bold{\Phi}(t)$ are calculated for continuous time $t$ from 0 to 1 \cite{Boqing2012}. These projections are then concatenated to form the infinite-dimensional feature vectors $\boldsymbol{z}_i^\infty$ and $\boldsymbol{z}_j^\infty$ \cite{Boqing2012}. Via the kernel trick, the inner product between these vectors gives the geodesic flow kernel, $\boldsymbol{G}$,
	
	\begin{equation}
		\langle \boldsymbol{z}_i^\infty, \boldsymbol{z}_j^\infty \rangle = \int_{0}^{1} \left( \boldsymbol{\Phi}(t)^T \boldsymbol{x}_i \right)^T \left( \boldsymbol{\Phi}(t)^T \boldsymbol{x}_j \right) dt = \boldsymbol{x}_i^T \boldsymbol{G} \boldsymbol{x}_j
	\end{equation}
	
	\noindent 
	where $\boldsymbol{G} \in \mathbb{R}^{\text{D} \times \text{D}} $ is a positive semidefinite matrix \cite{Boqing2012}. The matrix $\boldsymbol{G}$ can be written in closed form \cite{Boqing2012},

	\begin{equation}
		\boldsymbol{G} = \boldsymbol{Q} 
		\begin{bmatrix}
			\boldsymbol{V}_1 & 0 \\
			0 & -\boldsymbol{\tilde{V}}_2
		\end{bmatrix}
		\begin{bmatrix}
			\bold{\Lambda}_1 & \bold{\Lambda}_2 \\
			\bold{\Lambda}_2 & \bold{\Lambda}_3 
		\end{bmatrix}
		\begin{bmatrix}
			\boldsymbol{V}_1^{\intercal} & 0 \\
			0 & -\boldsymbol{\tilde{V}}_2^{\intercal} 
		\end{bmatrix}
		\boldsymbol{Q}^{\intercal}
	\end{equation}
	
	\noindent 
	where $\bold{\Lambda}_1$, $\bold{\Lambda}_2$, and $\bold{\Lambda}_3$ are diagonal matrices with elements \cite{Boqing2012},
	
	\begin{equation}
		\lambda_{1i} = 1 + \frac{\sin(2\theta_i)}{2\theta_i}, \quad \lambda_{2i} = \frac{\cos(2\theta_i) - 1}{2\theta_i}, \quad \lambda_{3i} = 1 - \frac{\sin(2\theta_i)}{2\theta_i}
	\end{equation}
	
\begin{figure}[h!]
		\vspace{0.5cm}
		\centering
		\includegraphics[width=0.95\textwidth]{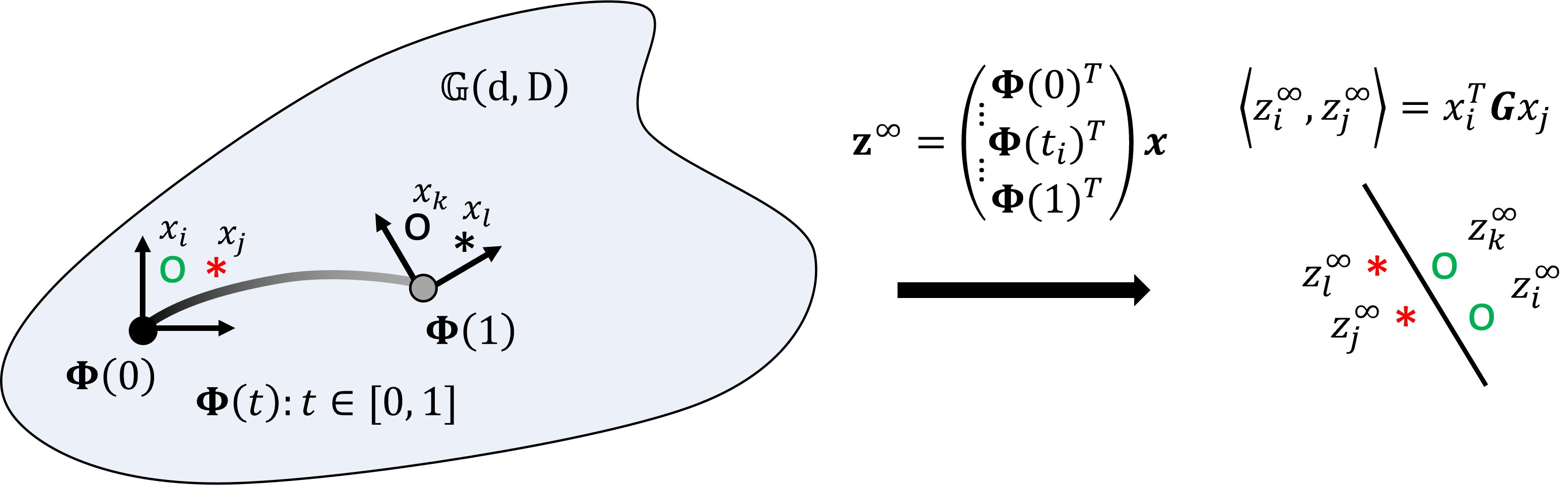}
		\caption{Geodesic flow kernel, adapted from \cite{Boqing2012}.}
		\label{fig:GFK}
	\end{figure}

	\noindent This process is shown in Figure \ref{fig:GFK}. It is important to note that the geodesic flow kernel measures similarity by considering information from both the source and target domains. As such, the GFK is insensitive to smooth domain shifts, and can therefore provide better transfer compared to traditional methods. The GFK is used in this work to facilitate transfer between two highly-disparate structures by embedding it into a support vector machine (SVM) classifier, as discussed in Sections \ref{Case-1} and \ref{Case-2}.
	
\vspace{12pt} 
\section{Case 1: Transfer between heterogeneous bridges} \label{Case-1}
The intent of this work is to demonstrate how in some cases, it may be possible to treat highly-disparate structures as differing only in their values for a certain set of parameters. Models of these structures can be generated, and varying these parameters within a given interval can result in a continuous and gradual morphing of one structure into another. A subset of these models can be used to facilitate information transfer, by incrementally transferring along the chain. This section describes a case study that demonstrates the intermediate-structures approach using data from entirely simulated structures: specifically, highly-simplified FEMs of two and three-span bridges.

	\subsection{Generation of finite element models}
 
	Simple FEMs of two- and three-span bridges were generated via PyMAPDL using tetrahedral elements. Both models included a 32-metre steel deck with a rectangular cross-section (7-metre width by 0.7-metre height). The concrete supports were 3 metres in width, 1 metre in thickness, and 5 metres in height. The two-span bridge had a single support located at the centre of the deck. The three-span bridge differed only in that it had an additional support located at 75\% of the deck length. Both supports had the same material properties and boundary conditions.
	
	Boundary conditions were implemented consistently at the deck ends and at the column bases. Both ends of the deck and the bases of the concrete supports were connected to ground via sets of linear translational springs ($k = 1 \times 10^{12}$ N/m) that constrained motion in the X, Y, and Z directions. Likewise, for each deck end and each column base, a small number of nodes on the underside of each column were selected, and at each such node springs of stiffness $k = 1 \times 10^{12}$ N/m were applied to connect the supports to ground and constrain them in the X, Y, and Z directions. The dimensions and material properties of these bridge models is summarised in Table \ref{tab:Case1_Properties}. 
	
	\begin{table}[h]
		\normalsize
		\renewcommand{\arraystretch}{1.25}
		\caption{Model Properties for Case 1.}
		\label{tab:Case1_Properties}
		\centering
		\begin{tabular}{lcccccc}
			\toprule
			Component & L (m) & W (m) & H (m)
			& E (Pa)
			& $\rho$ (kg/m$^3$)
			& $k$ (N/m) \\
			\midrule
			Deck     & 32 & 7 & 0.7
			& $2\times10^{11}$
			& 7850
			& $1\times10^{12}$ \\
			Supports & 3  & 1 & 5
			& $1.9\times10^{10}$
			& 2392
			& $1\times10^{12}$ \\
			\bottomrule
		\end{tabular}
	\end{table}

\begin{figure}[h!]
		\centering
		\subfloat[\label{fig:source-bridge}]{\includegraphics[width=0.45\textwidth, trim = {30cm 20cm 30cm 15cm}, clip]{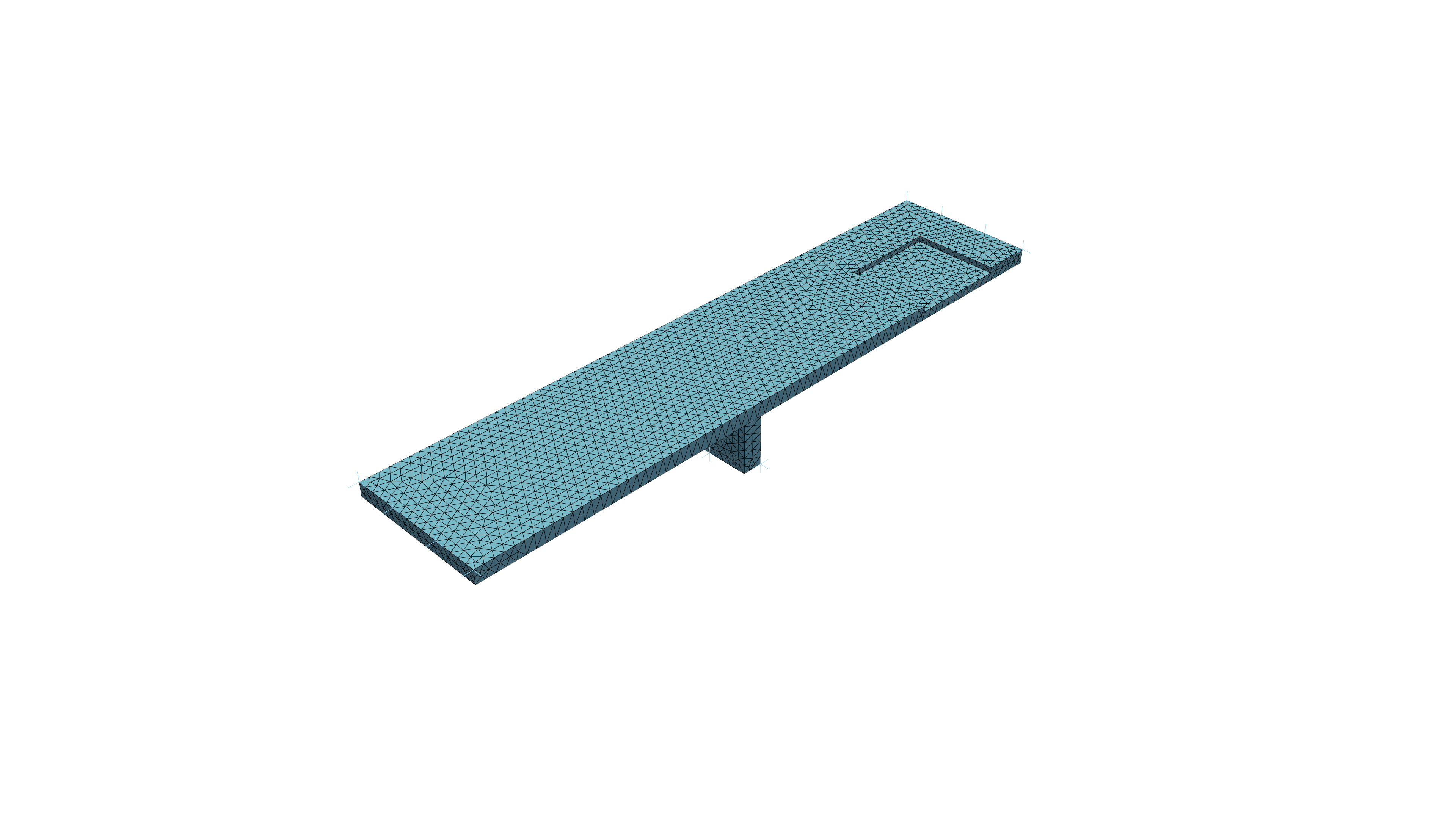}} 
		\hspace{1em} 
		\subfloat[\label{fig:target-bridge}]{\includegraphics[width=0.45\textwidth, trim = {30cm 20cm 30cm 15cm}, clip]{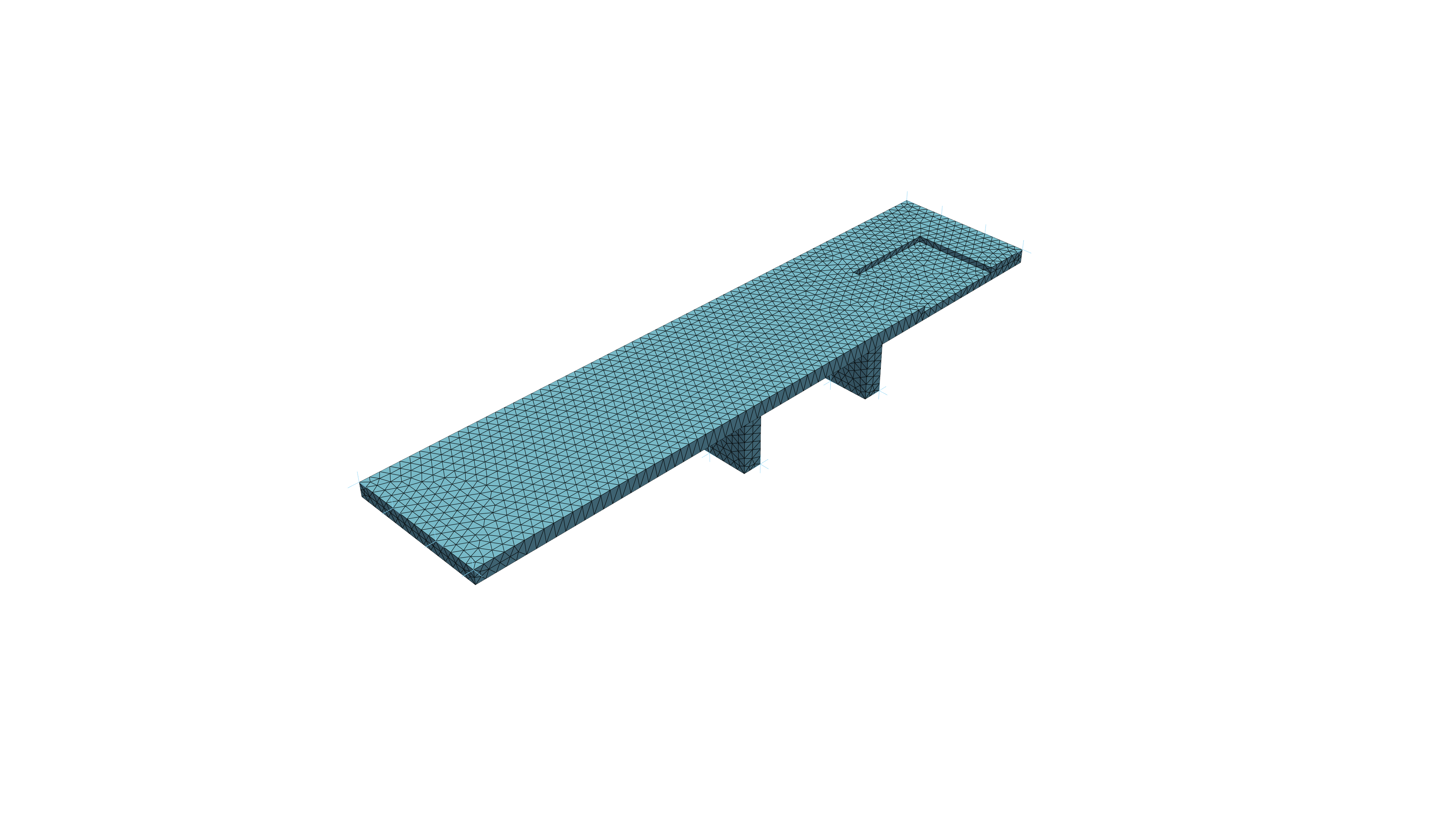}} 
		\caption{Two-span source (a) and three-span target bridges (b) in the damaged state.}
	\end{figure}

	Deck damage was represented as a localised partial-depth material loss. The damage patch was located on the top surface of the deck, centred at 85\% from the left-hand end of the deck. The patch extended 70\% of the deck width from the front edge and 40\% into the depth of the deck. The source and target bridges (shown in the damaged state) are shown in Figures \ref{fig:source-bridge} and \ref{fig:target-bridge}.

	The two-span bridge was defined as the source structure, $ \boldsymbol{S}_1 $, and the three-span bridge as the target structure, $ \boldsymbol{S}_2 $. It was predicted that $ \boldsymbol{S}_1 $ and $ \boldsymbol{S}_2 $ would be too far apart in the representation space for reliable transfer under statistic alignment alone. Therefore, a series of interpolating structures were generated from the parametric families of two- and three-span bridges, to facilitate transfer. 
	
\begin{figure}[ht!]
		\centering
		\definecolor{bridgeBlue}{RGB}{121,166,183}    
		\definecolor{bridgeBlueLight}{RGB}{203,221,228} 
		
		\begin{tikzpicture}[scale=0.3, line cap=round, line join=round]
			\def\L{32}      
			\def\H{1.0}     
			\def\Hs{5.0}    
			\def\Ws{1.5}    
			
			\draw[fill=bridgeBlue, draw=bridgeBlue] (-\L/2,0) rectangle (\L/2,\H);
			
			\draw[fill=bridgeBlue, draw=bridgeBlue]
			(-\Ws/2,-\Hs) rectangle (\Ws/2,0);
			
			\draw[bridgeBlueLight, very thin] (-1.5,-\Hs) -- (10,-\Hs);
			
			\foreach \alpha/\xpos/\opa in {0.25/2/0.25,
				0.50/4/0.45,
				0.75/6/0.70,
				1.00/8/1.00} {
				\draw[fill=bridgeBlue, draw=bridgeBlue,
				fill opacity=\opa, draw opacity=\opa]
				(\xpos-\Ws/2,-\Hs) rectangle (\xpos+\Ws/2,0);
			}
			
			\draw[->, thick, bridgeBlue]
			(0,-\Hs-1.5) -- (8,-\Hs-1.5)
			node[midway, below, text=black] {increasing $\alpha$};
			
		\end{tikzpicture}
		\caption{Schematic of model generation process. The first support remained fixed while the second support moved to the right and materialised as $\alpha$ increased.}
		\label{fig:case-1-IS}
	\end{figure}
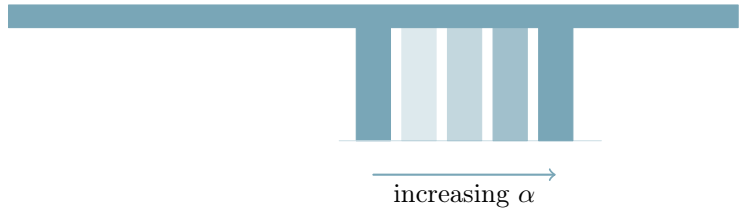

	The family of bridge configurations was parameterised with respect to the position and material properties of the second support, which were driven by a scalar variable $\alpha \in [0,1]$. At $\alpha = 0$, the second support was absent, corresponding to the two-span source structure, $ \boldsymbol{S}_1 $. As $\alpha$ increased, the second support was translated linearly along the deck towards its final position corresponding to the target three-span bridge, $ \boldsymbol{S}_2 $. At the same time, its mass and stiffness were increased linearly from zero to the full concrete properties by scaling the Young’s modulus and mass density in proportion to $\alpha$.  At $\alpha = 1$, the second support reached its target position and was fully materialised, corresponding to $ \boldsymbol{S}_2 $. For the transfer-learning experiments, $\alpha$ was discretised into a finite set of values $\{\alpha_i\}_{i=1}^N$, which can be interpreted as parameter draws along the continuous path between the source and target bridges. Each incremental $\alpha_i$ therefore corresponds to a distinct interpolating structure with its own dynamics. A total of 16 interpolating structures were generated. The model generation process is shown in Figure \ref{fig:case-1-IS}.

	While the present study adopts a simple linear progression in the support parameters, in practice, this choice leads to a subset of low--$\alpha$ configurations that do not correspond to realisable geometries. For small values of $\alpha$, the moving support overlaps with the midspan support. Very low $\alpha$ configurations were therefore excluded as interpolating structures. This particular behaviour is an artefact of the particular parametrisation adopted here, rather than a limitation of the intermediate-structure transfer approach itself. (Discontinuities can exist in the mathematical interpretation, for example, if the source and target domain have different topologies. However, as long as the data representations are sufficiently similar, transfer may still be possible). Future work will include exploring alternative approaches to parameter growth, for example, nonlinear mappings such as logistic curves, and asymmetric parameter growth, for example, different rates of change for material properties versus the support position.

	\subsection{Transfer approach and results}
	
	In this case study, all samples (healthy and damaged) are assumed available for each structure, but only partial label information is available beyond the source. The source structure is assumed fully labelled. For each intermediate domain and the final target domain, a subset of healthy samples is assumed labelled, while damaged labels are not assumed. The unlabelled samples are used transductively during sequential transfer via pseudo-labelling. For all structures, eigensolutions were computed to obtain the first 15 natural frequencies, which were then used to construct normal- and damage-condition datasets. The natural frequencies were assembled into a feature vector, which was then replicated 100 times. For each repetition, each modal frequency was perturbed with independent Gaussian noise with a standard deviation equal to 0.8\% of its mean value. In this case study, the natural frequencies served as features in a two-class classification problem with the goal of distinguishing between healthy and damaged states.
	
	Prior to transfer, NCA was applied to the data to remove linear domain shift using known healthy data. (The normalisation parameters were estimated using the labelled healthy subset only, and the resulting affine transform was applied to all samples (healthy and unlabelled/damaged) in the corresponding domains.) With the exception of the source $ \boldsymbol{S}_1 $, which was assumed fully labelled, 20\% of the healthy samples were assumed labelled. To characterise how the feature subspaces evolved along the chain, alignment between leading principal components of the normalised data was monitored at every hop. At each hop, PCA was applied separately to the current source and target domains, yielding orthonormal bases for their respective subspaces. Two sets of dot products were then computed: (1) between the leading principal directions of the current source and current target, and (2) between the leading principal directions of the current source and those of the original source (with the final target included as the last point), $ \boldsymbol{S}_1 $. Because these principal directions were unit-norm, the resulting dot products were equal to the cosines of the angles between the associated one-dimensional subspaces, i.e., the cosines of the corresponding principal angles.
	
\begin{figure}[h!]
		\vspace{0.5cm}
		\centering
		\includegraphics[width=0.95\textwidth]{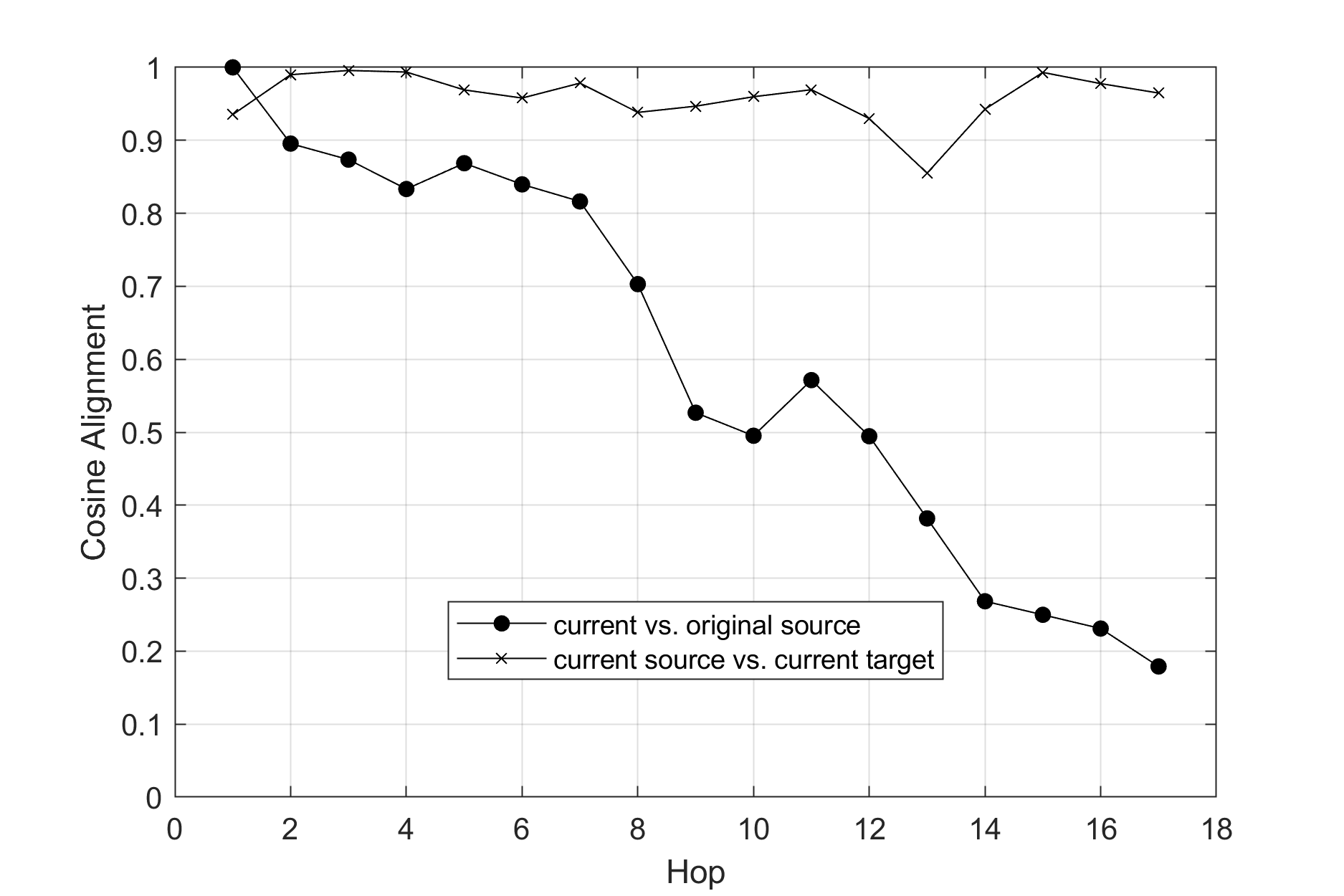}
		\caption{Cosine alignment along the transfer path for Case 1. The alignment between the current source and target is shown in black with $\mathbf{\times}$ markers, with values close to 1 indicating that adjacent structures in the chain are consistently similar. The alignment between the current source and the original source is shown in black with $\bullet$ markers, with decreasing values showing progressively less similarity as the chain moves farther away from the original source.}
		\label{fig:case-1-dot}
	\end{figure}

	Plotting these cosine alignments as a function of hop index provided a geometric view of the transfer chain. Values close to one indicated that the dominant subspaces were nearly collinear (small principal angles), whereas lower values indicated larger rotations. The alignment between the current source and current target summarised how similar the local domains were at each hop, while the alignment between the current source and the original source quantified how far the chain had moved away from the initial subspace. Sharp drops in either alignment highlighted hops at which the subspaces rotated more abruptly; in practice, such locations indicated regions where additional intermediate structures were required to obtain a smoother evolution of the subspaces and, consequently, more stable transfer behaviour. The cosine alignment curves are shown in Figure \ref{fig:case-1-dot}.

	Figure \ref{fig:case-1-dot} shows that the curve tracking the alignment between the current and original source was shown to be nearly linearly monotonically decreasing. On the other hand, the curve tracking the alignment between the current source and current target was shown to be a relatively flat line, with most values above 0.9. This behaviour is close to what would be expected in an ideal case where the feature subspaces evolve linearly between the source and target. As such, the alignment curves provide an \emph{a priori} indication that a relatively few intermediates will be sufficient to achieve positive transfer.
		
	To explore the effect of different path resolutions, multiple chains of intermediate structures were constructed as ordered sequences of structure indices, with up to six intermediate states. For each admissible number of intermediates, 60 random chains were generated by selecting distinct intermediate structures and sorting them, yielding a pool of candidate chains. At each hop, classification was performed using two approaches. In the first approach, PCA was fitted on the source domain (using all available samples, labelled and unlabelled) and a linear SVM was trained on the projected source data and then used to classify the projected target data. In the second approach, a geodesic flow kernel was constructed from the source and target PCA subspaces, and a linear SVM was trained and applied in the resulting embedded space. 
	
	The subspace dimension was fixed at $d=5$ for all runs, and was not tuned per chain or method. This value was selected as a moderate regularisation and to satisfy method-specific dimensionality constraints. (The GFK requires the subspace dimension to be no greater than half the ambient feature dimension and, in practice, no greater than the rank supported by the available samples. In this case, the maximum allowable subspace dimension was $d=7$.) For consistency with the GFK approach, the same subspace dimension was used for the linear runs.	
	
	For each chain, label propagation was performed incrementally from source to target over 100 Monte Carlo realisations (where the added noise was randomised as well as the indices for the labelled healthy data). At each hop, pseudo-labels for the unlabelled samples were generated from the current classifier predictions (with the labelled healthy subset retained as fixed labels) and propagated to the next hop, where the source from one hop became the target for the next, and so on, until the final target $ \boldsymbol{S}_2 $ was reached. Transfer accuracy at the final hop was summarised as mean and standard deviation over the 100 runs for each chain, and performance was compared as a function of the number of intermediate structures, with the best-performing chain identified at each level. After the optimal chain was determined for each number of intermediates, the selected chains were re-evaluated over 1000 additional runs using a different starting seed (i.e., an independent set of noise and labelled healthy datasets) to more accurately characterise variability.
	
\begin{figure}[h]
		\centering
		\includegraphics[width=\textwidth]{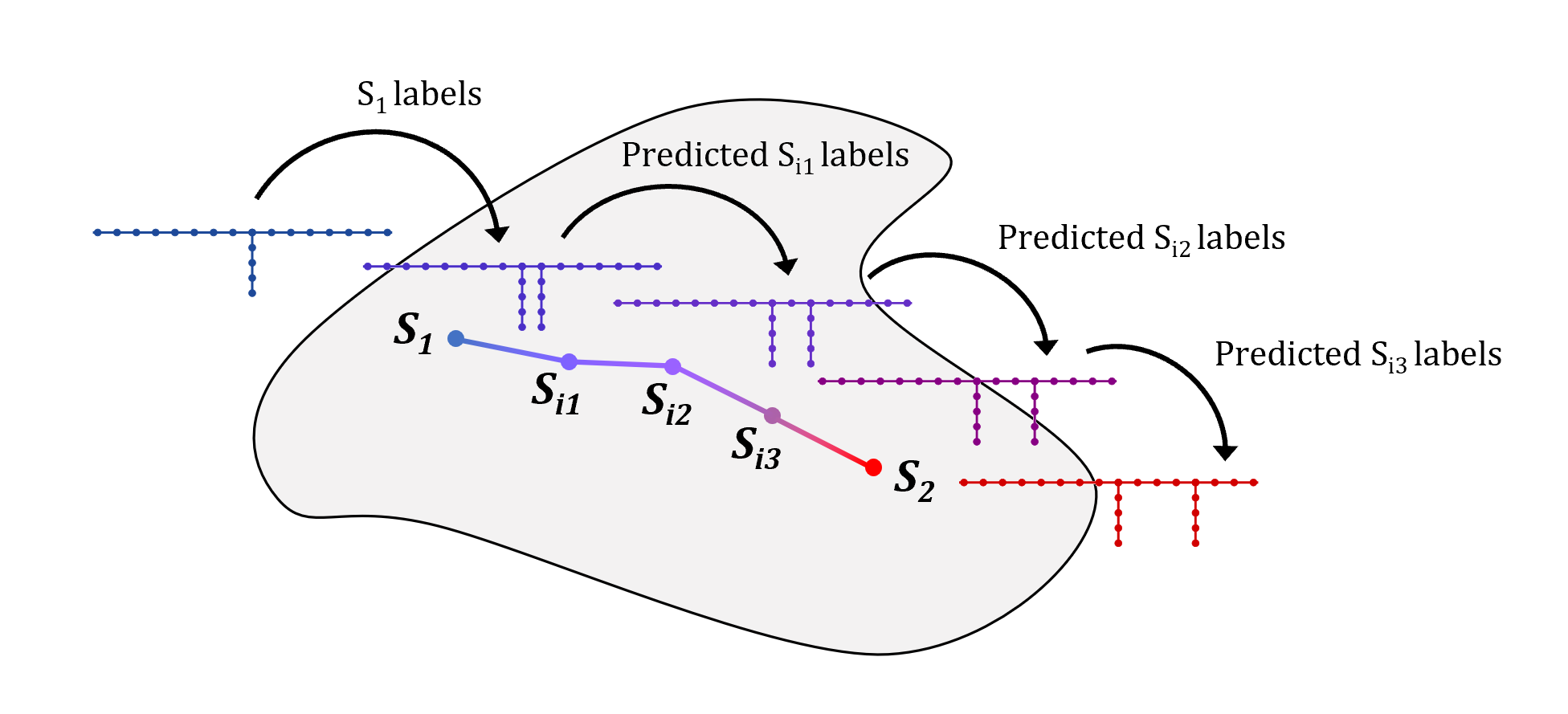}
		\caption{Heterogeneous transfer approach via intermediate structures.}
		\label{fig:intermediate_structures1}
	\end{figure}

	Note that this approach of selecting possible chains was label-informed (i.e., it relied on knowing the labels of the unlabelled data with the goal of selecting chains with the highest target-domain classification performance) and therefore not intended as deployable in practice. The intent was exploratory, to investigate the sensitivity of the transfer approach to the placement and number of intermediate domains for both the linear and GFK approaches, and to provide an idealised upper bound on transfer performance for the given selection of intermediates. A chain comprised of all the intermediates, which was considered acceptable based only on the continuity of cosine alignments of adjacent intermediates and not on target classification performance, was also employed. The transfer approach is shown pictorially in Figure \ref{fig:intermediate_structures1}, and results are shown in Table \ref{tab:best_chains}.
	
	\begin{table}[ht]
		\centering
		\caption{Best chains per number of intermediates $K$ for linear and GFK classification approaches. Values are mean $\pm$ standard deviation over 1000 noise realisations.}
		\label{tab:best_chains}
		\begin{tabular}{c
				c l
				c l}
			\toprule
			& \multicolumn{2}{c}{Linear} & \multicolumn{2}{c}{Geodesic Flow Kernel} \\
			\cmidrule(lr){2-3}\cmidrule(lr){4-5}
			$K$ & Accuracy (\%) & Best chain & Accuracy (\%) & Best chain \\
			\midrule
			0  & $44.44 \pm 0.00$  & (no chain)                 & $46.76 \pm 9.51$  & (no chain) \\
			1  & $44.70 \pm 4.08$  & [1 10 18]                  & $67.46 \pm 25.24$ & [1 11 18] \\
			2  & $57.16 \pm 22.11$ & [1 5 13 18]                & $94.75 \pm 15.83$ & [1 9 13 18] \\
			3  & $74.03 \pm 26.96$ & [1 4 12 13 18]             & $98.82 \pm 7.57$  & [1 4 12 13 18] \\
			4  & $93.59 \pm 17.85$ & [1 8 11 13 14 18]           & $99.82 \pm 3.11$  & [1 8 11 13 14 18] \\
			5  & $96.54 \pm 12.93$ & [1 5 8 11 13 16 18]         & $99.89 \pm 2.37$  & [1 5 8 11 13 16 18] \\
			6  & $98.64 \pm 8.83$  & [1 4 8 10 12 13 14 18]      & $99.91 \pm 1.99$  & [1 4 8 10 12 13 14 18] \\
			16 & $99.93 \pm 2.07$  & (all intermediates)        & $100.00 \pm 0.00$ & (all intermediates) \\
			\bottomrule
		\end{tabular}
	\end{table}

	Table~\ref{tab:best_chains} shows that when no intermediates were used ($K=0$), transfer from source to target performed poorly for both approaches, with mean accuracies of $44.44\%$ for the linear classifier and $46.76\%$ for the GFK. The linear approach exhibited effectively zero variability across noise realisations in this setting. This behaviour is consistent with a collapse of the linear classifier to predicting a single class for all unlabelled target samples; because $20\%$ of the healthy target data were treated as labelled and excluded from the reported accuracy, a constant prediction yields a deterministic accuracy equal to the prevalence of that predicted class within the remaining unlabelled target set.
	
	Introducing a single intermediate ($K=1$) had little effect on the linear approach ($44.70 \pm 4.08\%$), but increased GFK performance to $67.46 \pm 25.24\%$, albeit with substantial variability. With two intermediates ($K=2$), the GFK approach began to substantially improve ($94.75 \pm 15.83\%$), while the linear approach improved modestly ($57.16 \pm 22.11\%$).
	
	As $K$ increased further, both approaches benefitted from additional intermediates, but the linear method required more intermediates than GFK to achieve consistently high performance. In this case study, the linear classifier first exceeded $90\%$ mean accuracy at $K=4$ ($93.59 \pm 17.85\%$) and continued to improve with additional intermediates, reaching $96.54 \pm 12.93\%$ at $K=5$ and $98.64 \pm 8.83\%$ at $K=6$. In contrast, GFK reached very high mean accuracy by $K=3$ ($98.82 \pm 7.57\%$) and remained consistently above $99.0\%$ for larger chains. Near-perfect performance was achieved using a chain comprised of all available intermediates ($K=16$) for both approaches ($99.93 \pm 2.07\%$ for linear and $100.00 \pm 0.00\%$ for GFK).
		
	It is interesting that the two methods shared many of the best-performing chains (these were determined independently for both methods), suggesting that there exist \emph{geometrically favourable paths} for transfer. For example, at $K=3$ both methods selected the same chain, $[1\ 4\ 12\ 13\ 18]$, which achieved $74.03 \pm 26.96\%$ for the linear approach and $98.82 \pm 7.57\%$ for GFK.

	These results are consistent with the alignment curves in Figure \ref{fig:case-1-dot}. Successful transfer was associated with chains along which the dominant feature subspaces evolved in a relatively smooth way between source and target, even if this evolution involved rotations and occasional areas of difficult geometry. NCA alone cannot resolve rotational differences between the source and target subspaces. In the proposed approach, the classifiers are re-trained at each hop, but at a given hop the linear kernel still relies on a single PCA subspace and a single linear decision boundary to relate the current source and target, so it likely requires more intermediates to break a strongly rotating, curved relationship into smaller, locally linear steps. In contrast, the GFK implicitly interpolates intermediate subspaces between the source and target using a computed geodesic. This makes GFK naturally more robust to rotations and local irregularities, and explains why it can achieve near-perfect performance with relatively few intermediates.
	
\vspace{12pt} 
\section{Case 2: Transfer between a `bridge' and `aeroplane'} \label{Case-2}
The intermediate structures approach was validated using experimental data at the chain endpoints (i.e., the original source and final target). The data were obtained from further-simplified physical representations of the cartoon bridge and aeroplane described in \cite{DardenoISMA}, while FEMs were used to construct the transfer path. While the notion of transferring between a bridge and an aeroplane is admittedly far-fetched, this case study serves as a useful proof-of-concept, as the cartoon structures are connected via overlapping parametric families, which allows a transfer path to be constructed despite their dissimilarity.

	\subsection{Experiments} \label{experiments}	
\begin{figure}[h!]
		\centering
		\subfloat[\label{fig:exp-bridge}]{\includegraphics[width=0.95\columnwidth, trim = {2cm 9cm 6cm 12cm}, clip]{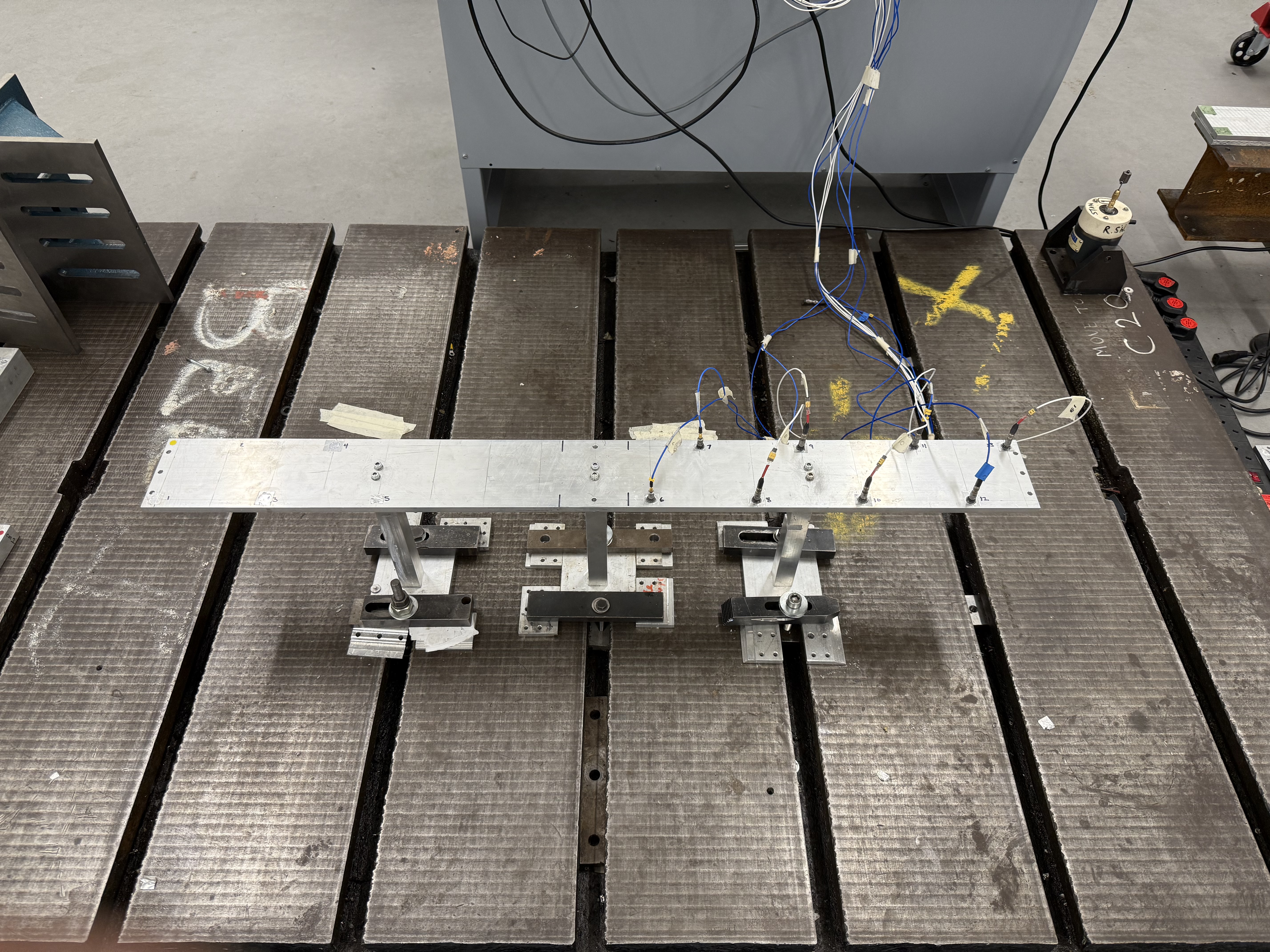}} \\
		\subfloat[\label{fig:exp-aeroplane}]{\includegraphics[width=0.95\columnwidth, trim = {5cm 2cm 6cm 3cm}, clip]{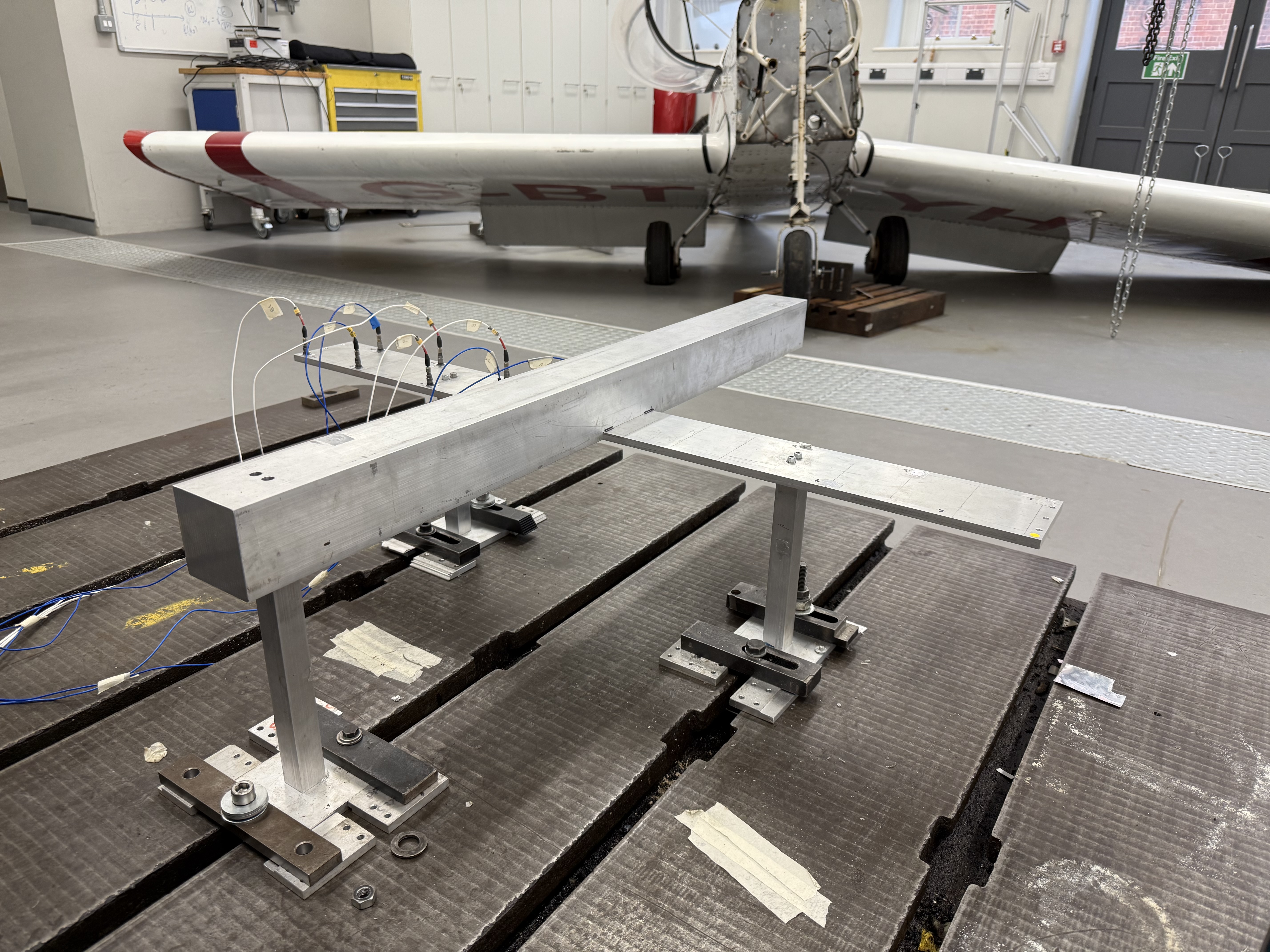}} 
		\caption{Experimental setup for bridge (a) and aeroplane (b) structures.}
	\end{figure}

	This case study used two aluminium test structures: a three-support `bridge' and an `aeroplane' (shown in Figures \ref{fig:exp-bridge} and \ref{fig:exp-aeroplane}). The bridge consisted of a 1.0 m × 0.10 m × 0.01 m aluminium deck supported by three 25.4 mm × 25.4 mm × 230 mm columns mounted on 10 mm foot pads that were rigidly clamped to a heavy steel base fixture. The aeroplane configuration used the same deck and supports but included an aluminium fuselage (0.0762 m × 1.00 m × 0.0762 m) attached to the centre of the deck. The middle support remained laterally centred as in the bridge but was repositioned such that its centre was located 38 mm behind the front edge of the fuselage (to very loosely represent landing gear), with no other geometric changes. 
	
\begin{figure}[h!]
		\centering
		\subfloat[\label{fig:D1}]{\includegraphics[height=5cm, trim = {20cm 8cm 6cm 12cm}, clip]{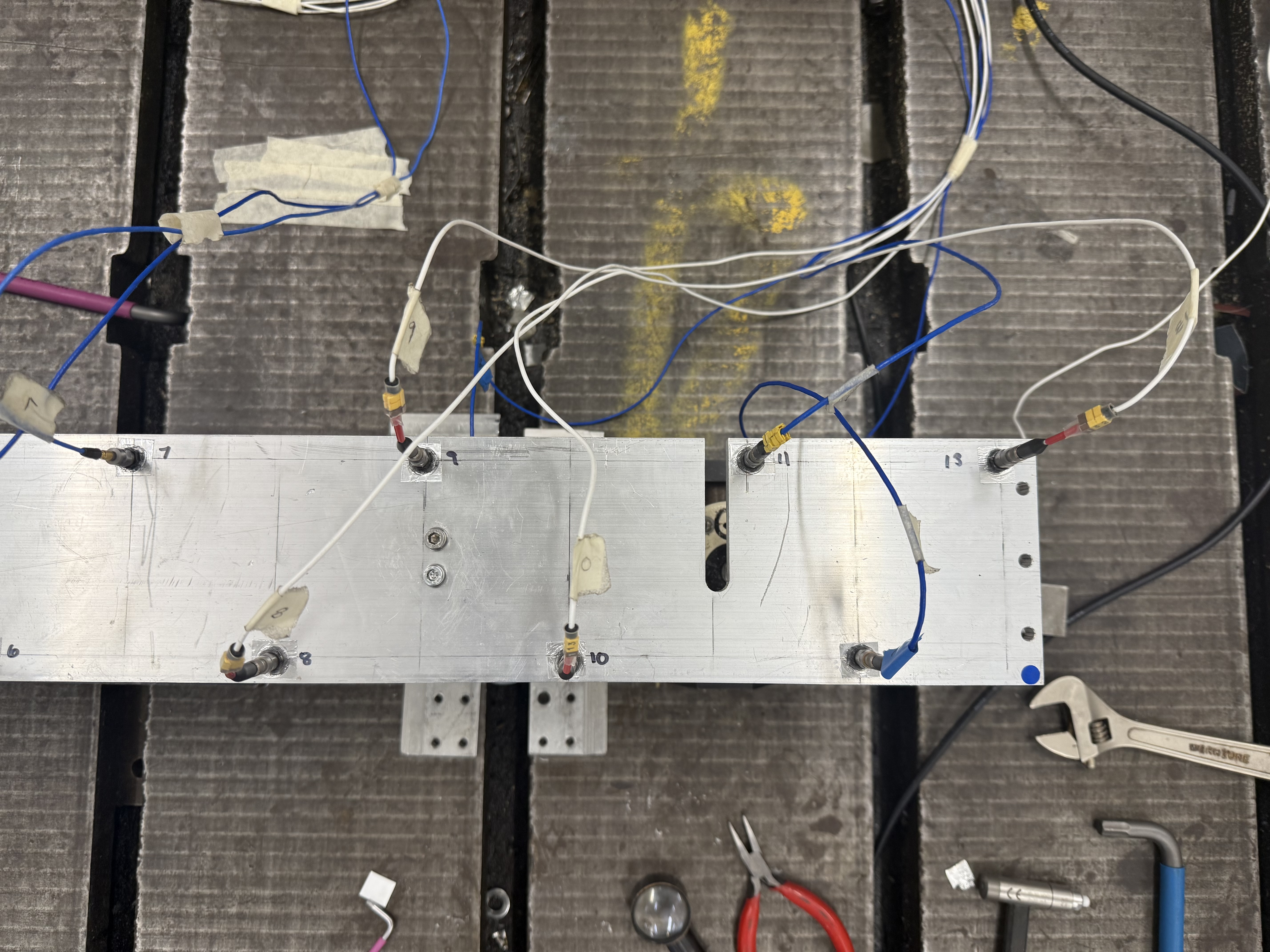}} 
		\hspace{1em} 
		\subfloat[\label{fig:D2}]{\includegraphics[height=5cm, trim = {10cm 10cm 50cm 50cm}, clip]{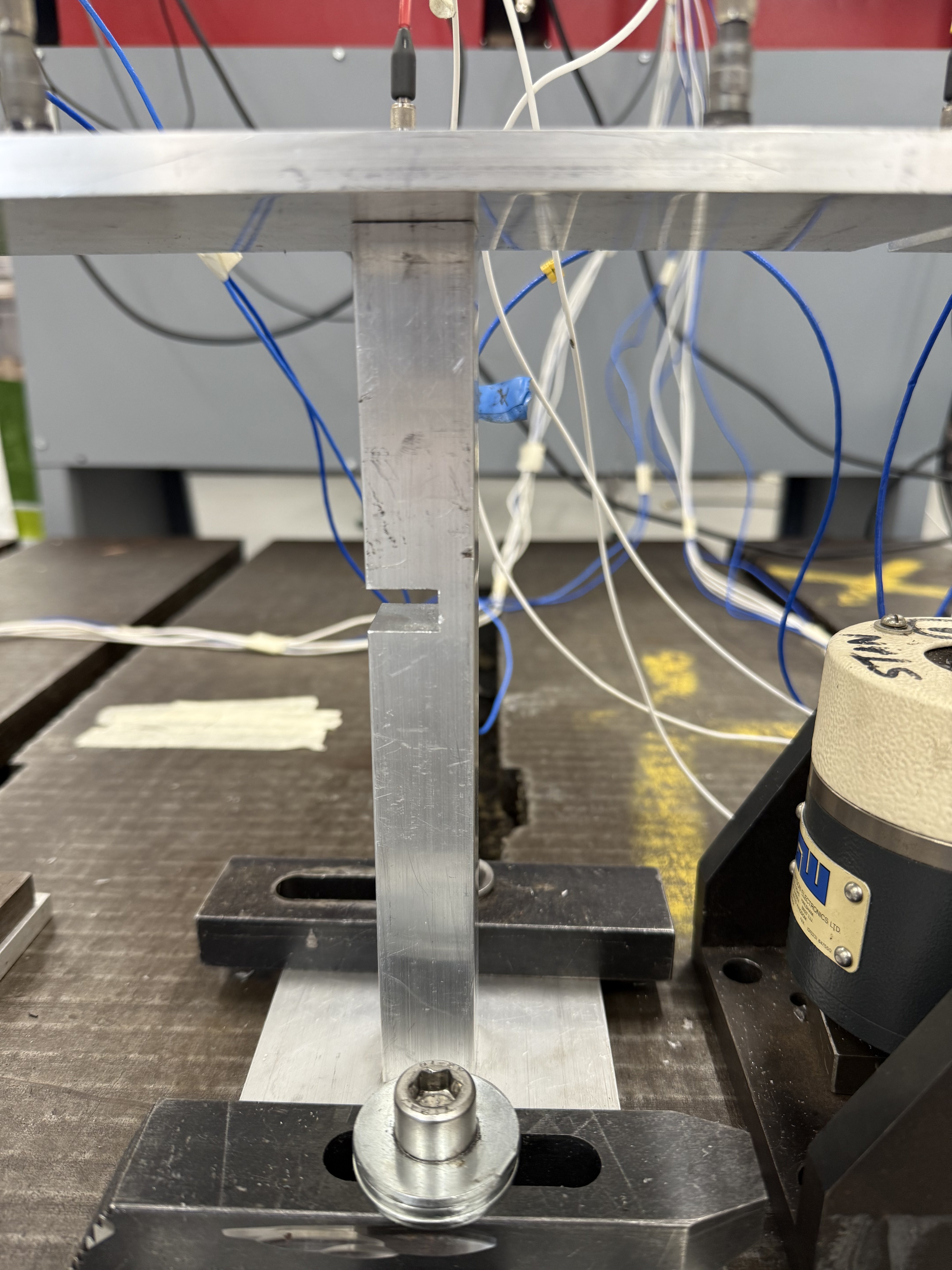}} 
		\caption{Damage to the deck (D1) (a) and (b) column (D2).}
	\end{figure}

\begin{figure}[h!]
		\centering
		\vspace{0.5cm}
		\subfloat[\label{fig:shape1}]{\includegraphics[width=0.33\linewidth, trim = {8cm 8cm 8cm 7cm}]{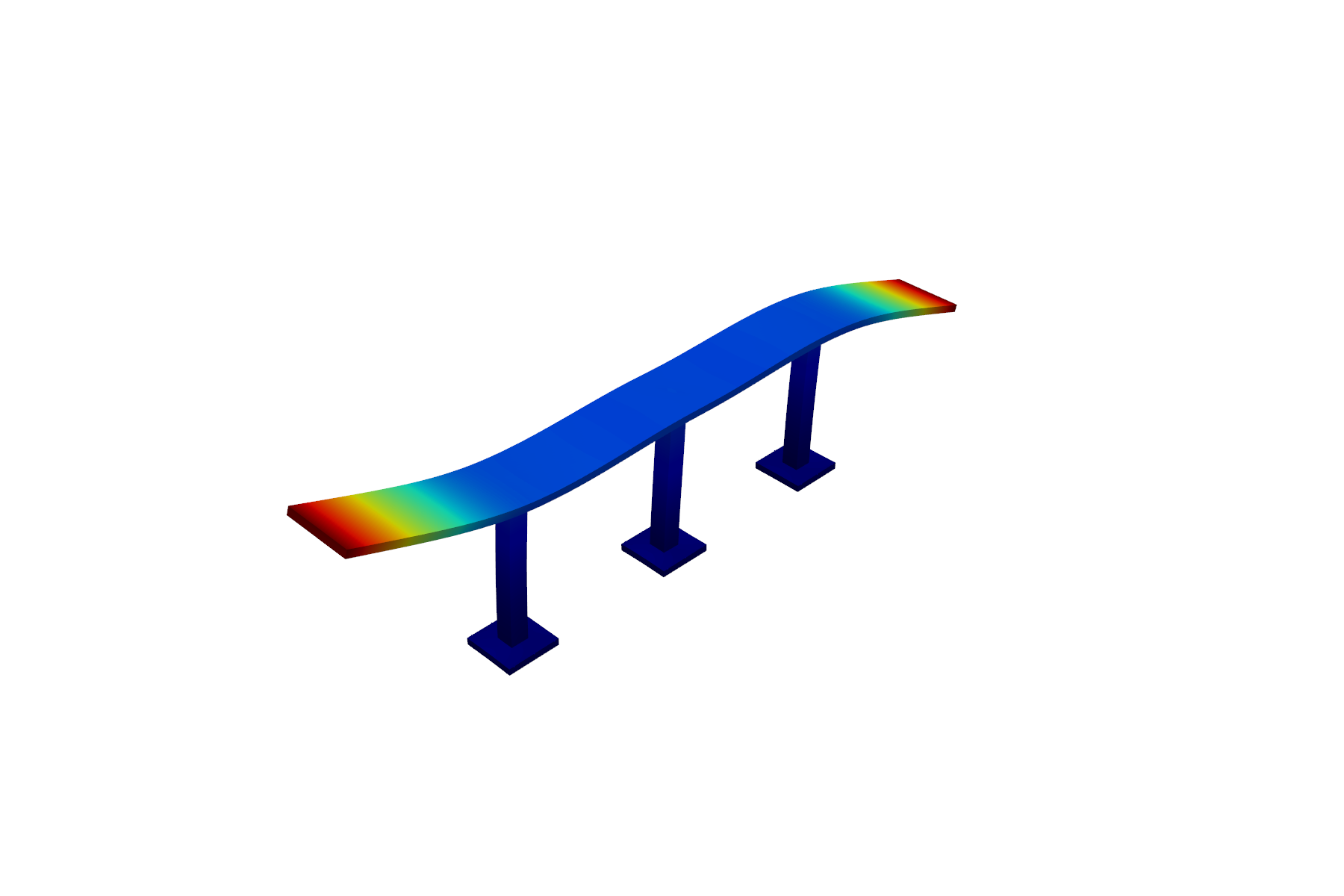}} 
		\subfloat[\label{fig:shape2}]{\includegraphics[width=0.33\linewidth, trim = {8cm 8cm 8cm 7cm}]{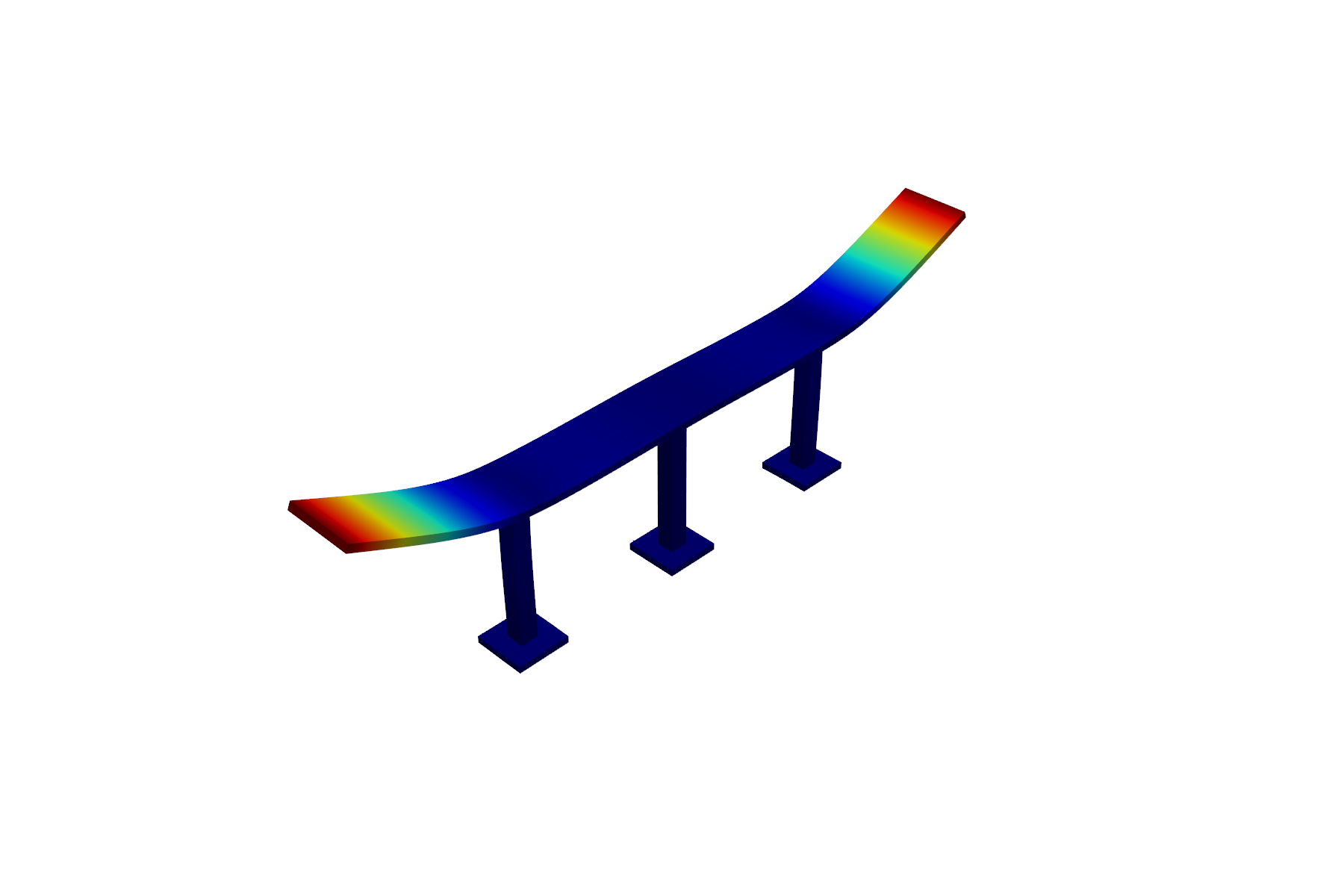}} 
		\subfloat[\label{fig:shape3}]{\includegraphics[width=0.33\linewidth, trim = {8cm 8cm 8cm 7cm}]{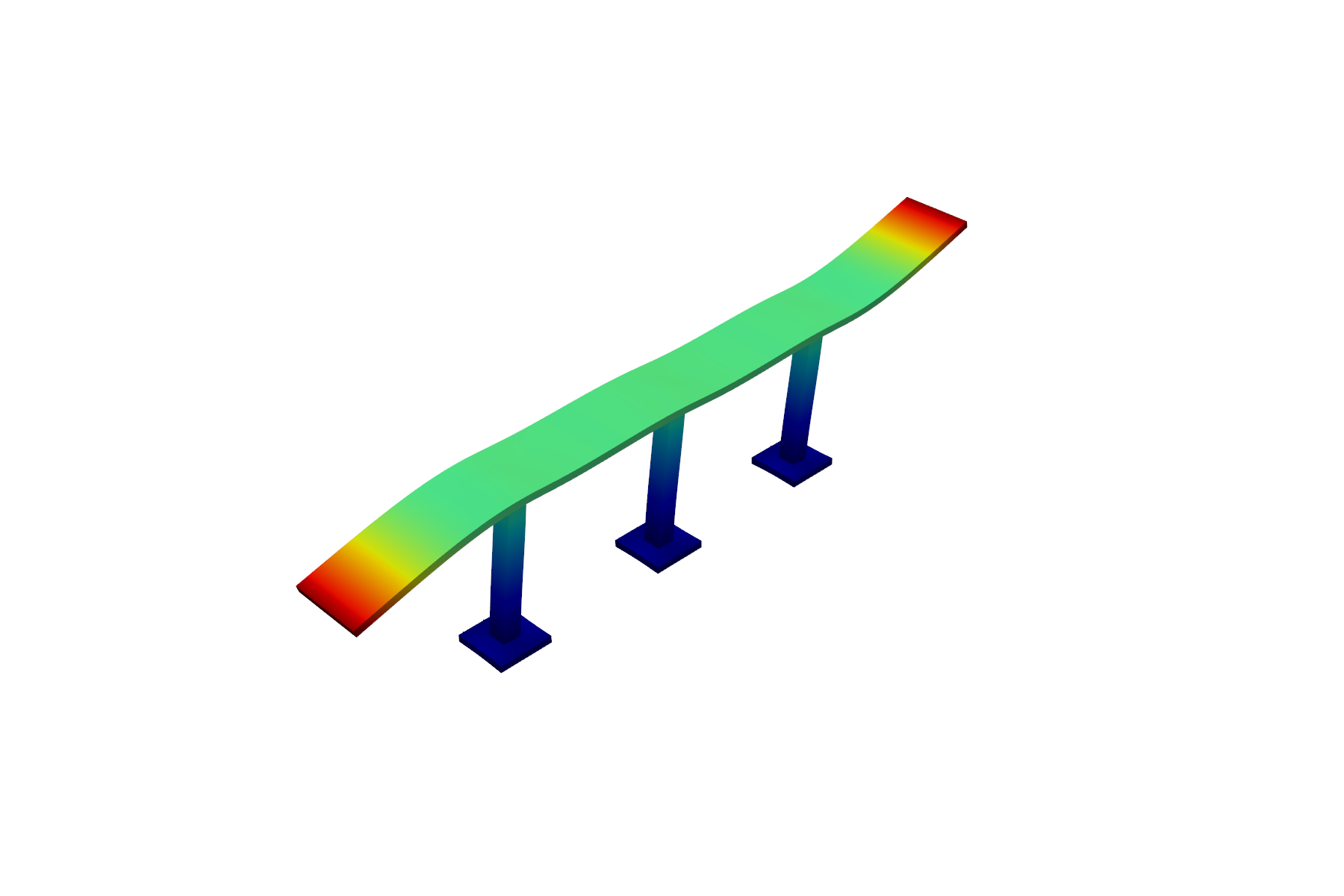}} \\
		\vspace{0.75cm}
		\subfloat[\label{fig:shape4}]{\includegraphics[width=0.33\linewidth, trim = {8cm 8cm 8cm 7cm}]{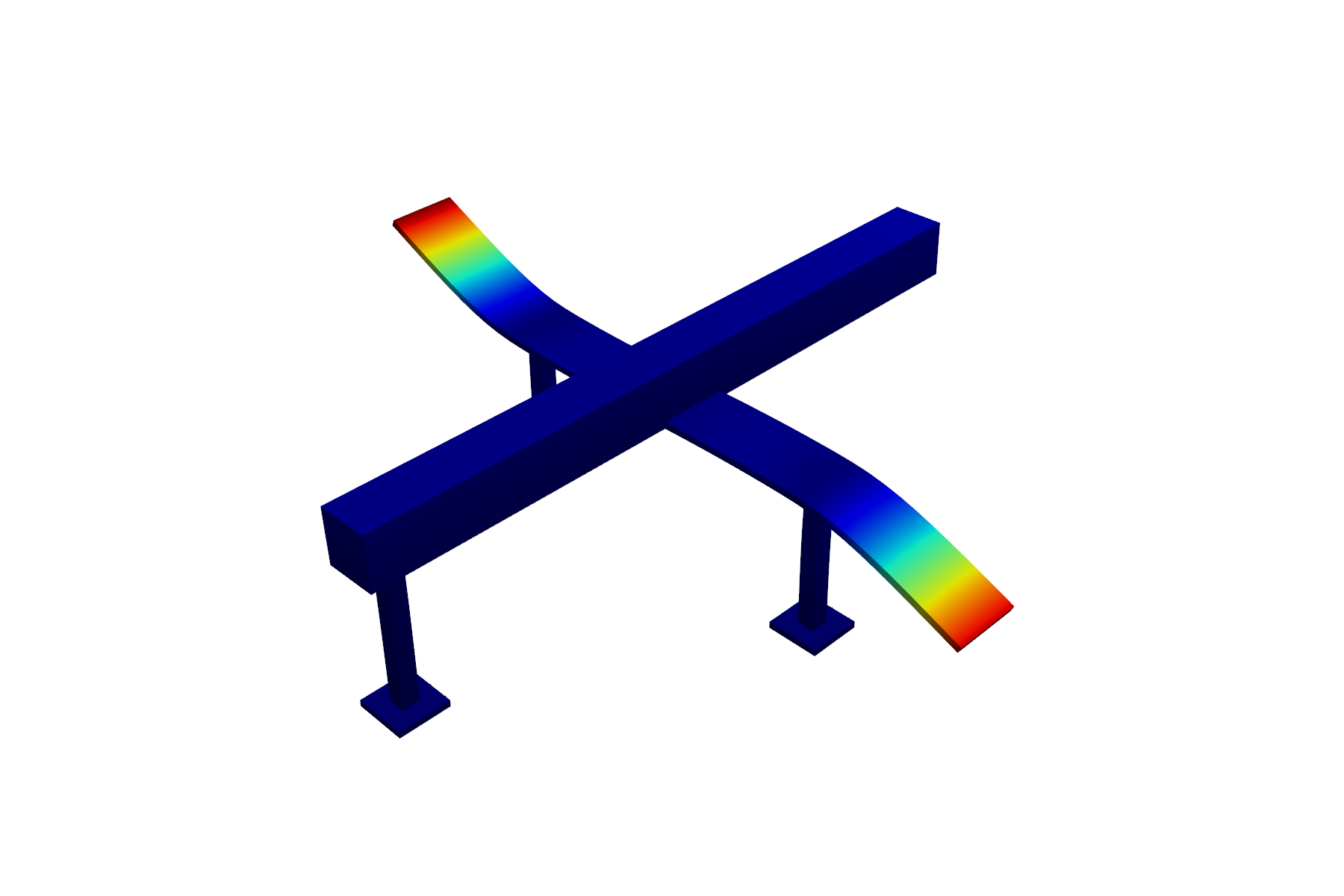}} 
		\subfloat[\label{fig:shape5}]{\includegraphics[width=0.33\linewidth, trim = {8cm 8cm 8cm 7cm}]{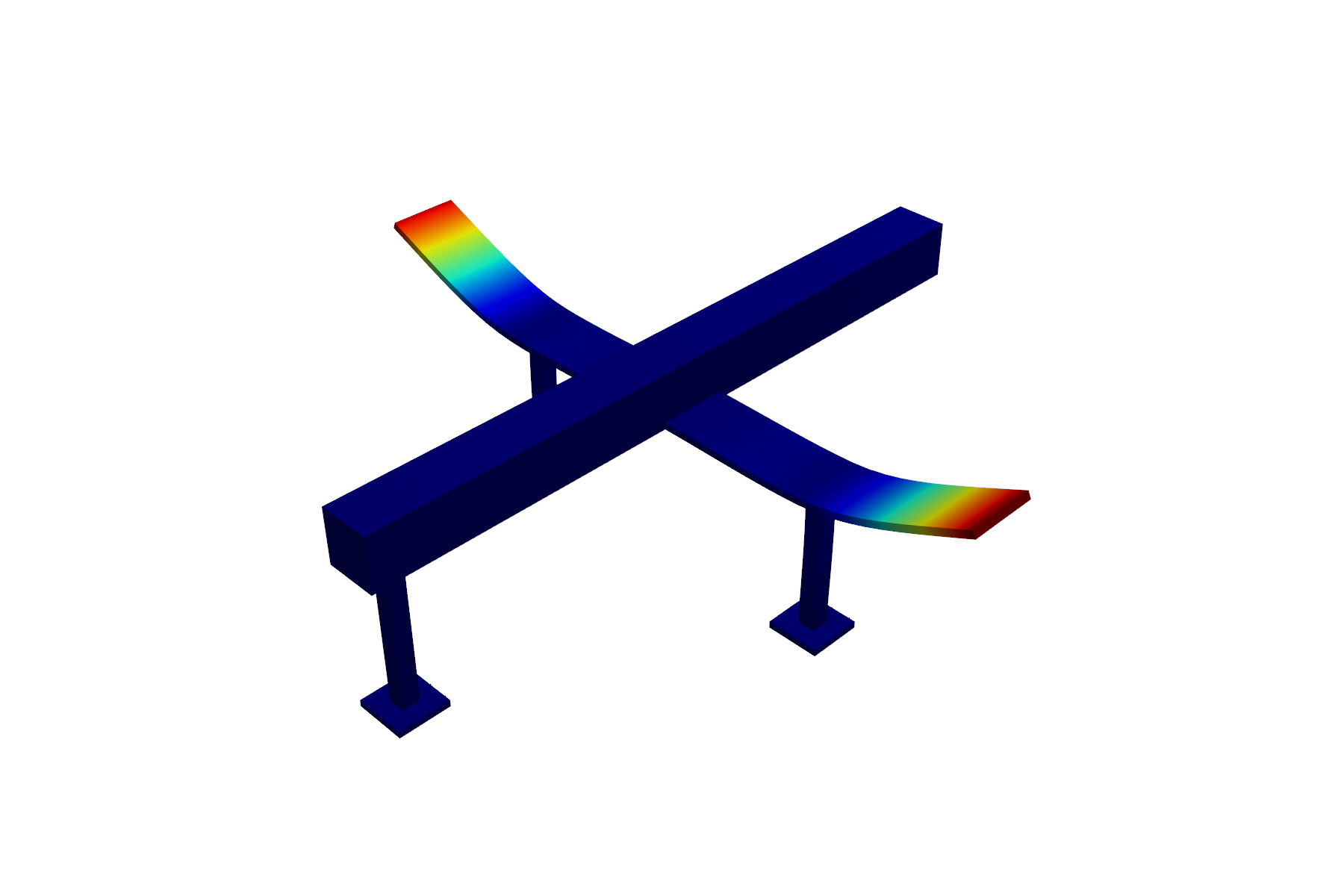}} 
		\caption{Lower bending modes for the bridge and aeroplane. Bridge modes at approximately (a) 100 Hz, (b) 116 Hz, and (c) 150 Hz. Aeroplane repeated root modes at approximately 117 Hz (d), (e).} 
	\end{figure}

	Two damage cases were considered. For the first damage case (D1), a slot that was 65 mm in length and 10 mm in width was machined fully through the deck thickness, located approximately 130 mm from the right-hand edge of the deck. For the second damage case, a 10-mm high slot was machined 18 mm into the side of the far-right column, spanning the full column width. It was located approximately 80 mm from the underside of the deck. The damaged deck and column are shown in Figures \ref{fig:D1} and \ref{fig:D2}. The intent was to have a strong effect on several of the lower bending modes; and these locations coincided with high strain areas for these mode shapes. The finite-element mode shapes in the band of interest (approximately 75 to 180 Hz) are shown in Figures \ref{fig:shape1} to \ref{fig:shape5}. (The FEMs and model-generation process are discussed below in Section \ref{FEMs}.)
	
\begin{figure}[h!]
		\centering
		{\includegraphics[width=0.9\textwidth]{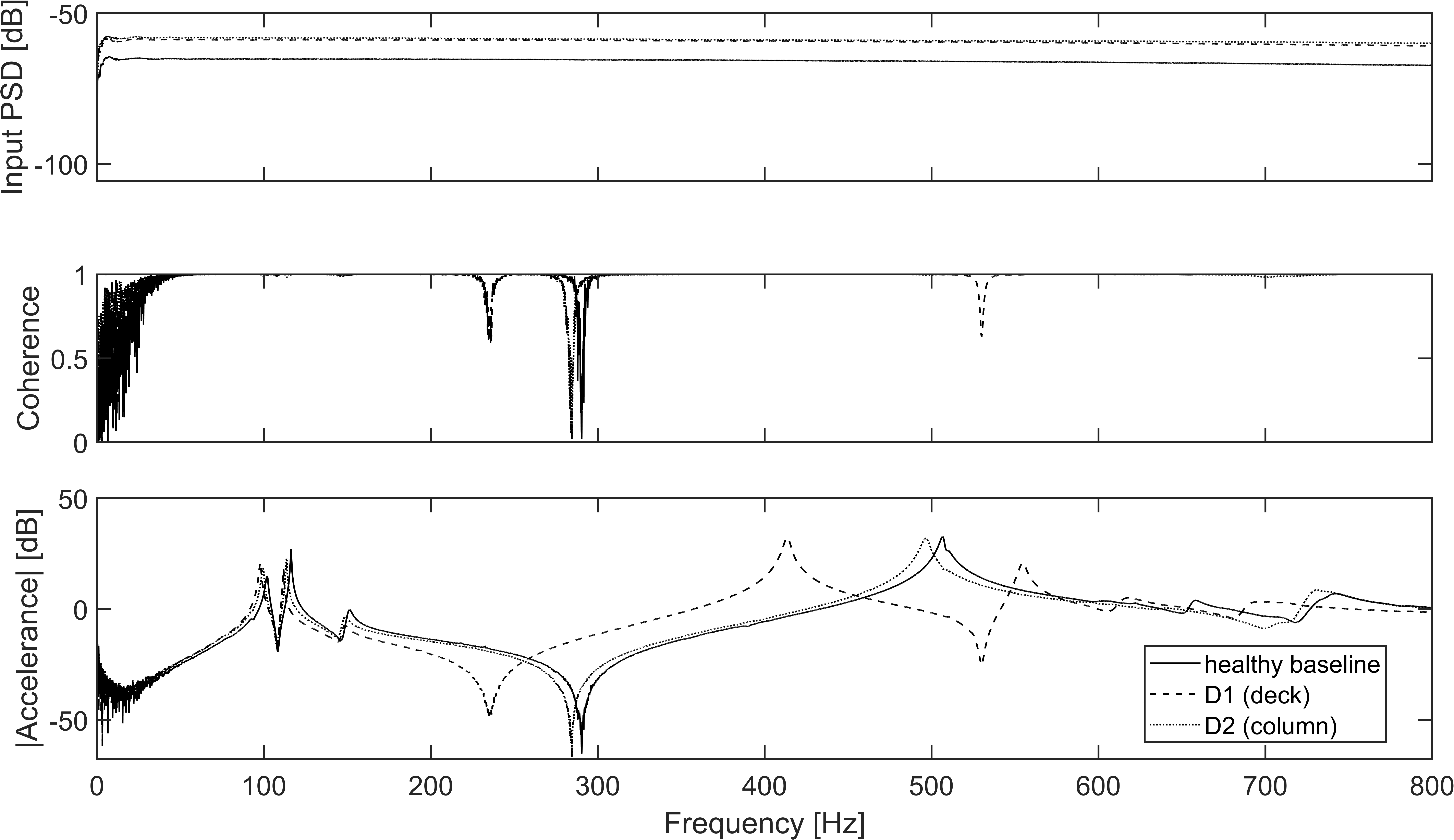}} 
		\caption{Representative input spectra, coherence, and FRFs for healthy and damaged states for the bridge configuration.}
		\label{fig:repMeas}
	\end{figure}

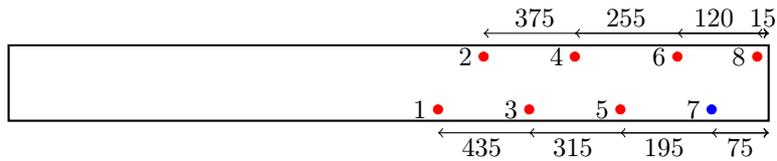
\begin{figure}
		\centering	
		\begin{tikzpicture}[scale=0.4, xscale=-1]
			\draw[thick] (0,0) rectangle (25,2.5);
			\foreach \x/\y in {
				10.875/2.125,
				9.375/0.375,
				7.875/2.125,
				6.375/0.375,
				4.875/2.125,
				3.000/0.375,
				0.375/0.375
			} \filldraw[red] (\x,{2.5-\y}) circle (0.15);
			
			\foreach \x/\y in {
				1.875/2.125
			} \filldraw[blue] (\x,{2.5-\y}) circle (0.15);
			
			\foreach \x/\y/\n/\xoff/\yoff in {
				10.875/2.125/1/ 0.6/ 0,
				9.375/0.375/2/ 0.6/ 0,
				7.875/2.125/3/ 0.6/ 0,
				6.375/0.375/4/ 0.6/ 0,
				4.875/2.125/5/ 0.6/ 0,
				3.000/0.375/6/ 0.6/ 0,
				1.875/2.125/7/ 0.6/ 0,
				0.375/0.375/8/ 0.6/ 0
			} \node[scale=1] at (\x+\xoff, {2.5-\y+\yoff}) {\n};
			
			\def\ytop{2.9}
			\def\ybot{-0.4}
			
			\foreach \x/\d/\ypos/\xoff in {
				10.875/435/\ybot/4,
				7.875/315/\ybot/2.5,
				4.875/195/\ybot/1,
				1.875/75/\ybot/0
			} {
				\draw[<->] (0,\ypos) -- (\x,\ypos);
				\node at (\x/2+\xoff,{\ypos-0.5}) {\d};
			}
			
			\foreach \x/\d/\ypos/\xoff in {
				9.375/375/\ytop/3,
				6.375/255/\ytop/1.5,
				3.000/120/\ytop/0.3,
				0.375/15/\ytop/0
			} {
				\draw[<->] (0,\ypos) -- (\x,\ypos);
				\node at (\x/2+\xoff,{\ypos+0.5}) {\d};
			}
		\end{tikzpicture}
		\caption{Sensor layout for all analyses. Accelerometers are shown in red, except for the drive point location which is shown in blue. All units are in mm, and measured from the right-hand edge of the deck/wing.}
		\label{fig:sensor-layout} 
	\end{figure}

	For each test, an impact hammer was used for excitation at a single point near the right-hand end of the deck/wing. Out-of-plane accelerations were collected at multiple locations using a consistent sensor layout (Figure \ref{fig:sensor-layout}). FRFs were computed over a consistent bandwidth for both structures to allow direct comparison in subsequent analyses. Representative FRFs at the drive-point location for the bridge configuration in the healthy, D1, and D2 states are plotted along with the corresponding coherence and input spectra in Figure \ref{fig:repMeas}. 
	
\begin{figure}[h!]
		\centering
		{\includegraphics[width=0.9\textwidth]{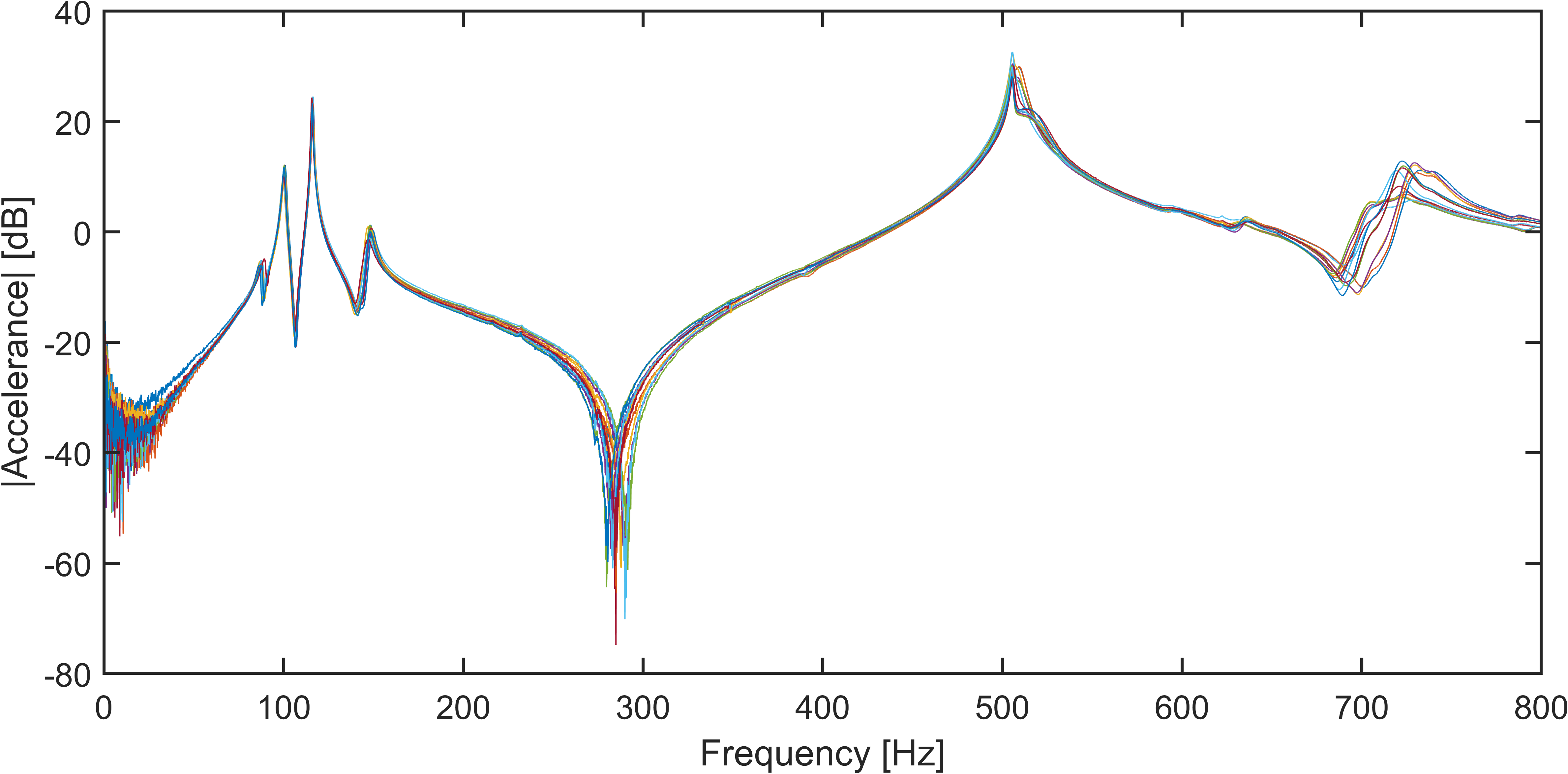}} 
		\caption{FRFs from a variability study on the bridge, where four different decks, and four different columns, were removed and reinstalled using consistent torque for the bolts.}
		\label{fig:var}
	\end{figure}

	Note that the damaged deck and column were separate to their healthy counterparts. This was considered to be acceptable because all components (healthy and damaged) were cut from the same stock material, and a torque wrench was used to tighten all bolts in a consistent manner. To ensure that this replacement procedure would not introduce excessive variability in the data, a variability study was conducted. This study was conducted using four healthy decks and four healthy columns with the structure in the bridge configuration. For each deck and each column, baseline data were collected, the component was removed, reattached, and the bolts re-tightened using the torque wrench, after which data were collected again. For the decks, the procedure consisted solely of removing and reattaching the deck. For the columns, the procedure required removing the deck, detaching the column and footpad assembly, reattaching the column assembly, and then reinstalling the deck before re-tightening all bolts. These results are shown in Figure \ref{fig:var}. Note that this study was completed several months prior to the primary tests. While the approach was largely the same, the variability tests used a slightly different setup and accelerometer layout, leading to some minor differences between the primary and variability datasets. 
		
	The resulting FRFs were consistent across repetitions. For the primary analyses, the band of interest contained three modes at approximately 100, 116, and 150 Hz. At these modes, differences between repeated measurements did not exceed 1.1, 0.75, and 2.9 Hz respectively. The differences between these modes for the healthy and D1 bridges was 2.6 Hz for each, and the differences between these modes for the healthy and D2 bridges were 4.0, 4.6, and 3.1 Hz, respectively. With the exception of the mode at around 150 Hz, which was more highly damped and less participatory than the other two, the variability caused by removing and reinstalling the columns and decks was substantially smaller than the frequency shifts introduced by the damage cases. Peak amplitudes at the modes of interest varied by 2–3 dB across repeats, consistent with impact-to-impact variability and sensitivity to measurement setup (e.g., occasional cable contact with the deck). As such, the primary analyses focussed on frequency shifts to indicate damage (and used normalisation to reduce sensitivity to amplitude differences).

	For the following analysis, a consistent frequency band covering the three modes at approximately 100, 116, and 150 Hz was selected for both the bridge and aeroplane configurations. Within this band, the FRF magnitudes at the sensor locations in Figure \ref{fig:sensor-layout} were used directly as features (following normalisation and projection to PCA space), rather than attempting to extract modal parameters. This avoided the risk that modal parameter estimation via curve-fitting might miss or inconsistently identify modes across structures and damage states.
	
	\subsection{Generation of finite element models} \label{FEMs}
	A finite element model of the bridge was generated via PyMAPDL using tetrahedral elements. The baseline model used nominal aluminium properties, with Young’s modulus $E = 69\,\text{GPa}$ and mass density $\rho = 2700\,\text{kg/m}^3$, which were then updated to align with test FRFs up to approximately 200 Hz. 
	
	Multiple stiff ground springs ($9\times10^{8}$ N/m) were applied to the foot pads at the bottom of each support in the X, Y, and Z directions to simulate the clamped boundary condition. Point masses (7.5 g each) were applied at the accelerometer locations to simulate the added mass of the sensors (these were estimated from the weight of the sensors and some extra mass related to the glue and other factors), and several dashpots were applied at the spring locations to simulate the damping effects of the connections, and laterally to the deck to simulate the damping effects of the cables. The aeroplane model was generated in the same manner, the only exception being the inclusion of a fuselage and translated middle support. The dimensions and (updated) material properties of the bridge and aeroplane models is shown in Table \ref{tab:Case2_Properties}.

	\begin{table}[h]
		\normalsize
		\renewcommand{\arraystretch}{1.25}
		\caption{Geometry and material properties for the bridge and aeroplane.}
		\label{tab:Case2_Properties}
		\centering
		\begin{tabular}{llccccc}
			\toprule
			Structure & Component & L (m) & W (m) & H (m) & E (Pa) & $\rho$ (kg/m$^3$) \\
			\midrule
			\multirow{3}{*}{Bridge}
			& Deck      & 1.00   & 0.10   & 0.010 & $6.8\times10^{10}$ & 2830 \\
			& Supports  & 0.0254 & 0.0254 & 0.220 & \multirow{2}{*}{\Same} & \multirow{2}{*}{\Same} \\
			& Foot pads & 0.0762 & 0.0762 & 0.010 &  &  \\
			\midrule
			\multirow{4}{*}{Aeroplane}
			& Wings     & 1.00   & 0.10   & 0.010 & $6.8\times10^{10}$ & 2830 \\
			& Supports  & 0.0254 & 0.0254 & 0.220 & \multirow{3}{*}{\Same} & \multirow{3}{*}{\Same} \\
			& Foot pads & 0.0762 & 0.0762 & 0.010 &  &  \\
			& Fuselage & 0.0762 & 1.00 & 0.0762 &  &  \\
			\bottomrule
		\end{tabular}
	\end{table}

\begin{figure}[ht!]
		\centering
		\begin{subfigure}{.3\textwidth}
			\centering
			\includegraphics[width=\linewidth]{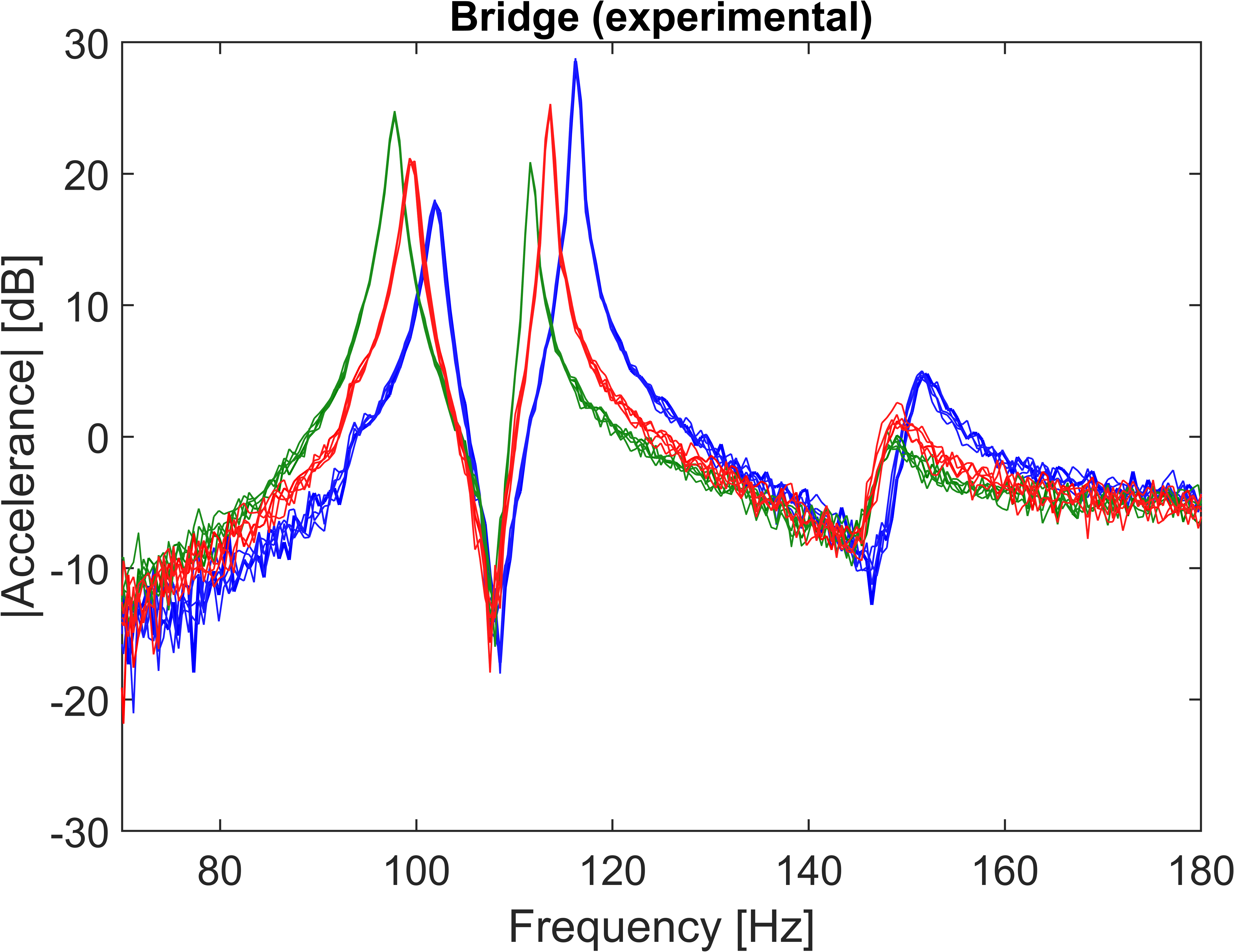}
		\end{subfigure}%
		\begin{subfigure}{.3\textwidth}
			\centering
			\includegraphics[width=\linewidth]{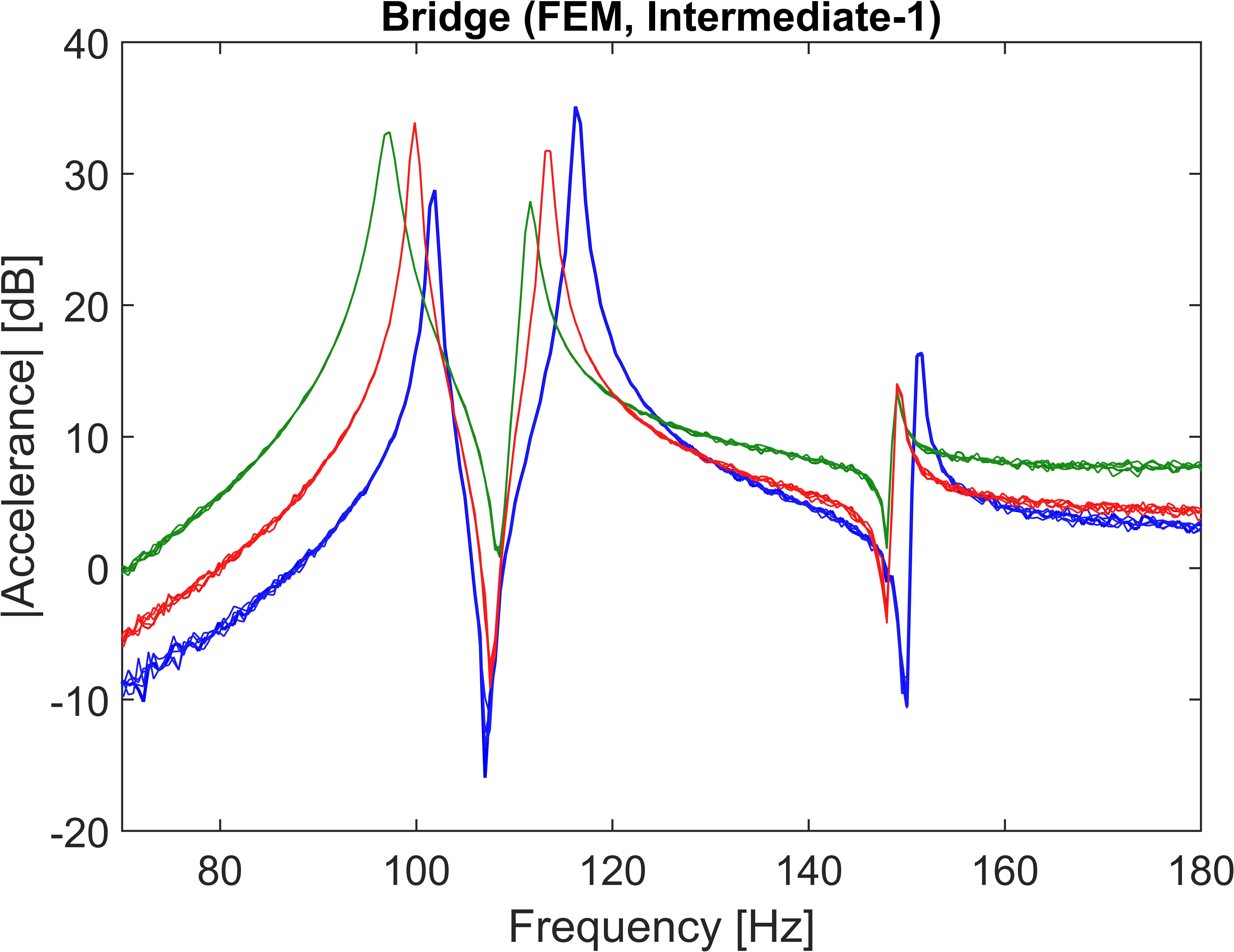}
		\end{subfigure} 
		\begin{subfigure}{.3\textwidth}
			\centering
			\includegraphics[width=\linewidth]{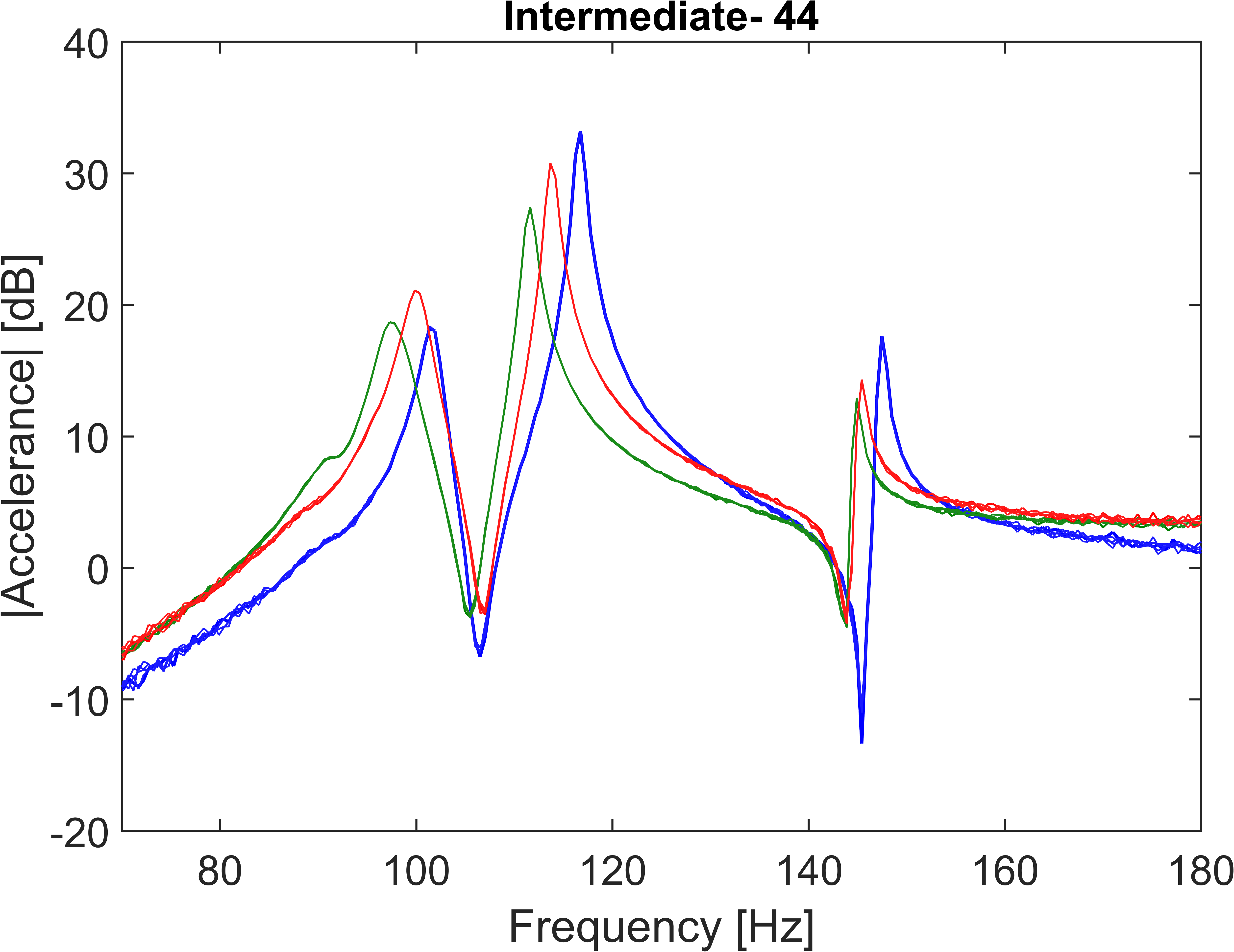}
		\end{subfigure} \\
		\vspace{1pt}
		\begin{subfigure}{.3\textwidth}
			\centering
			\includegraphics[width=\linewidth]{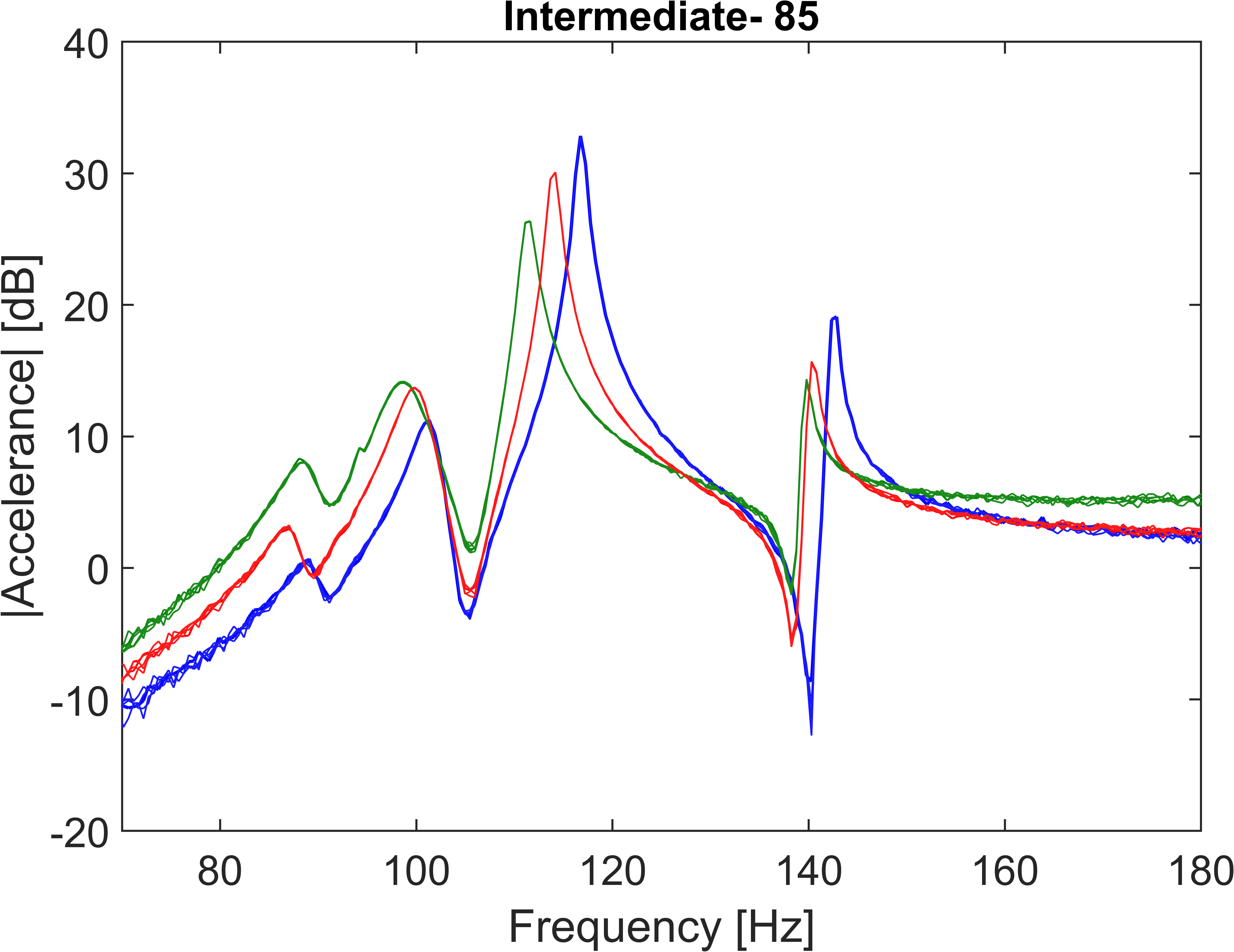}
		\end{subfigure} 
		\vspace{1pt}
		\begin{subfigure}{.3\textwidth}
			\centering
			\includegraphics[width=\linewidth]{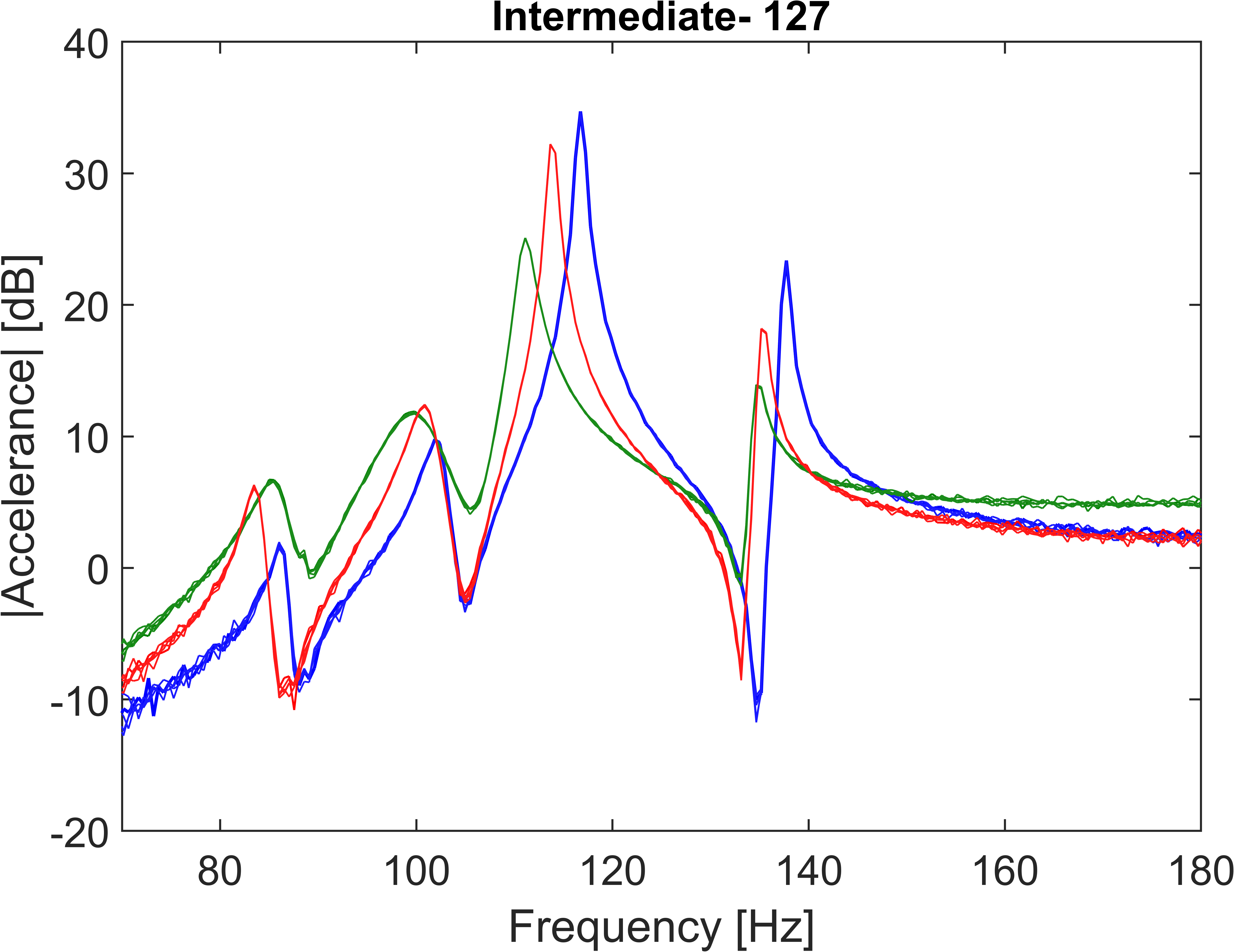}
		\end{subfigure} 
		\vspace{1pt}
		\begin{subfigure}{.3\textwidth}
			\centering
			\includegraphics[width=\linewidth]{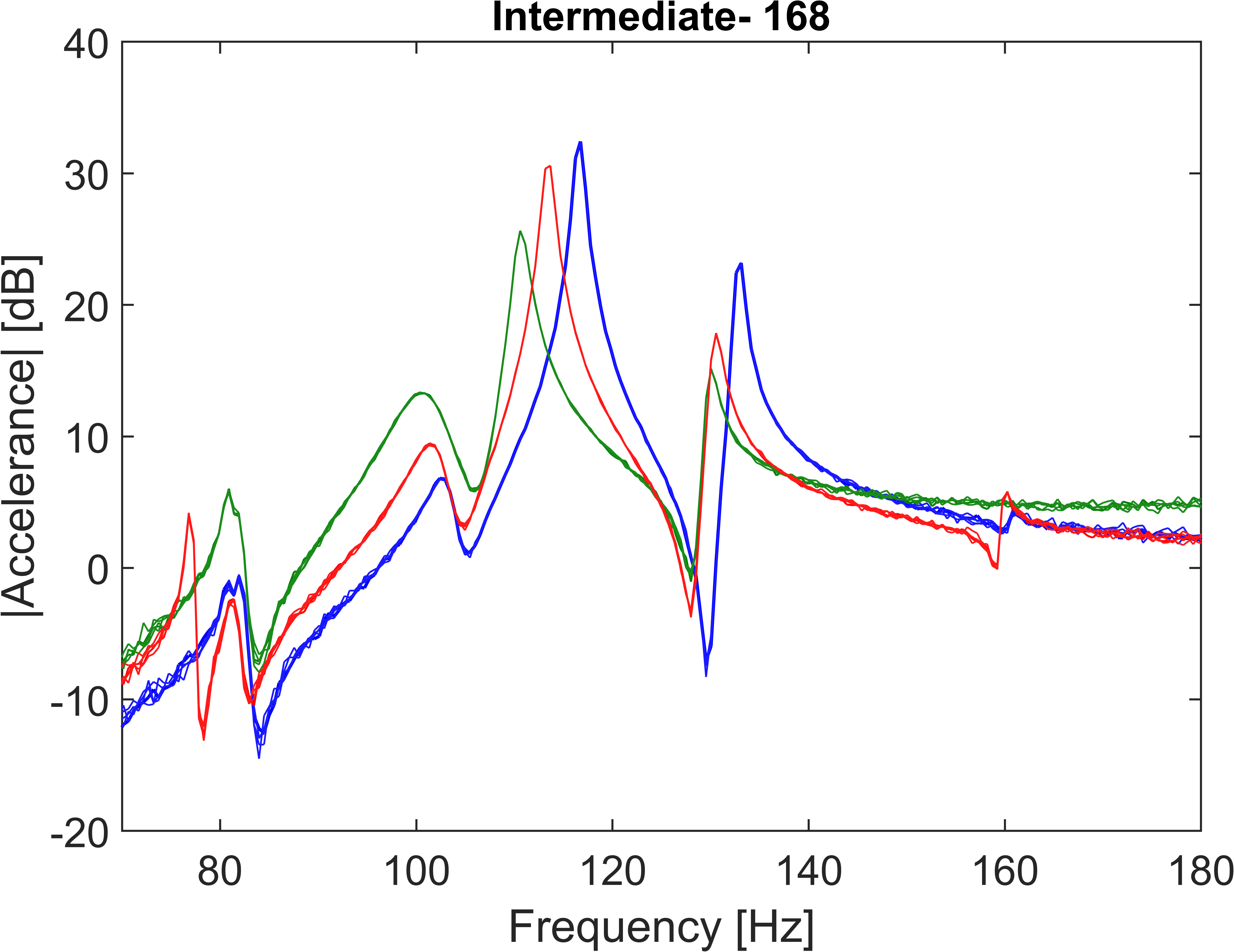}
		\end{subfigure} \\
		\vspace{1pt}
		\begin{subfigure}{.3\textwidth}
			\centering
			\includegraphics[width=\linewidth]{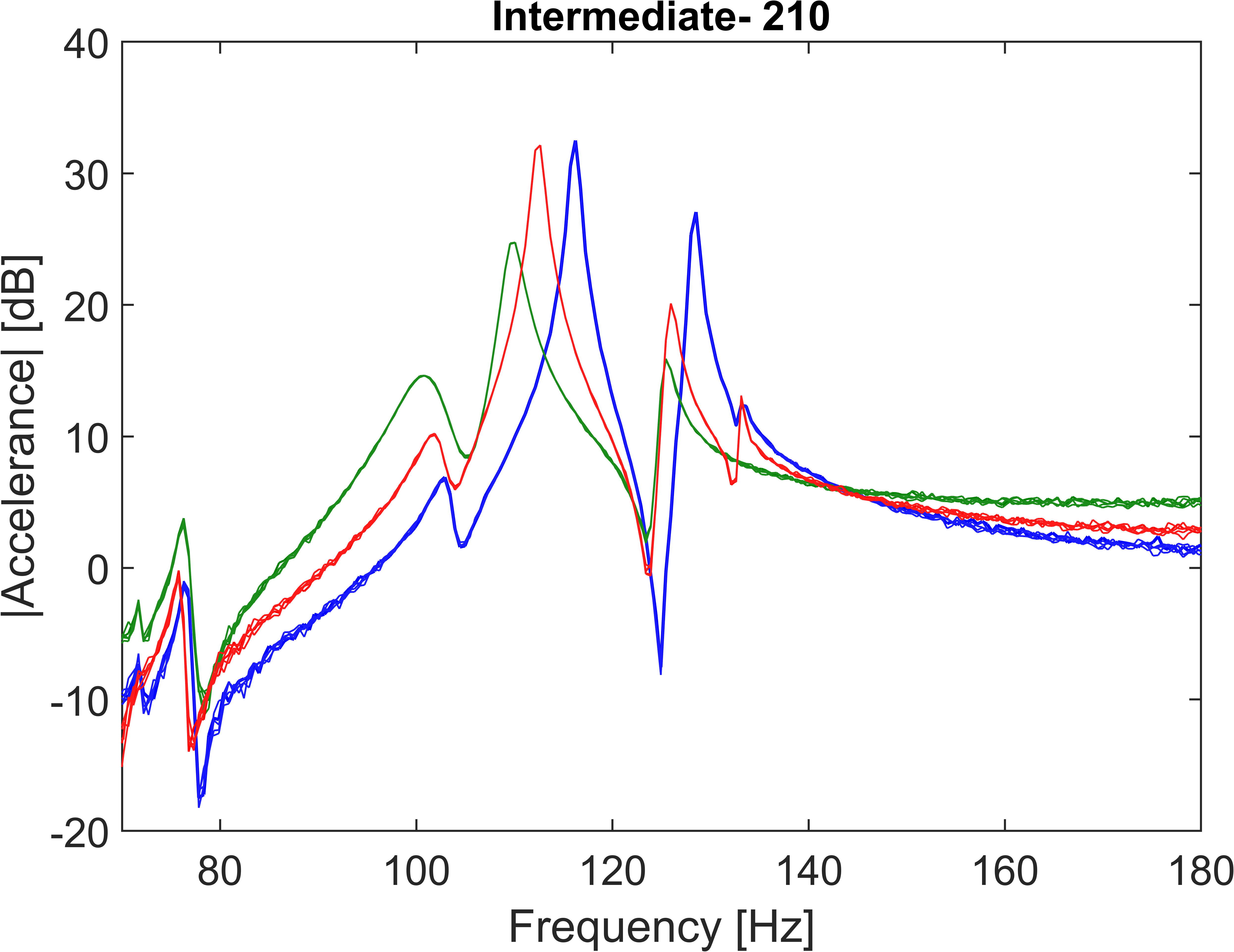}
		\end{subfigure}
		\vspace{1pt}
		\begin{subfigure}{.3\textwidth}
			\centering
			\includegraphics[width=\linewidth]{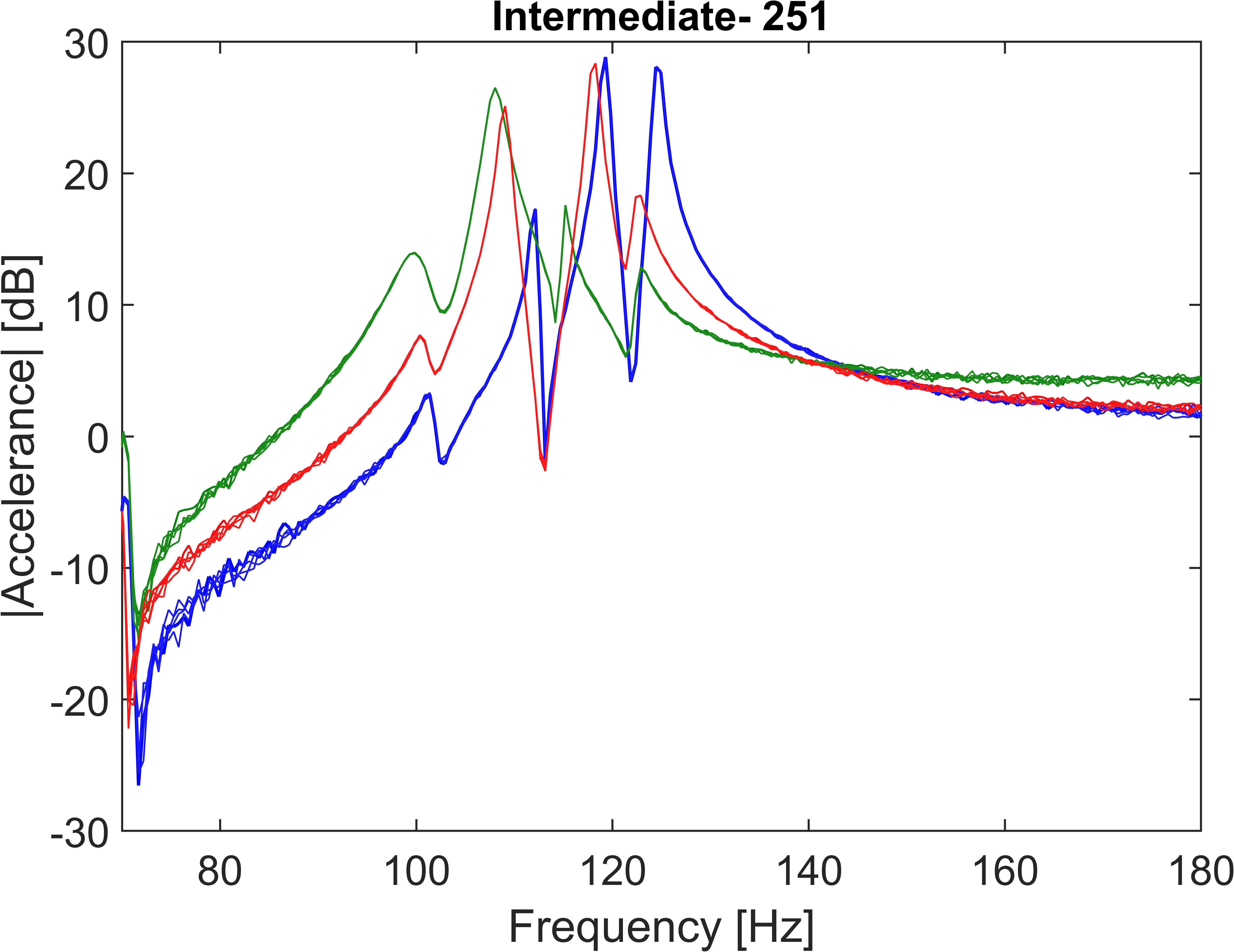}
		\end{subfigure}
		\vspace{1pt}
		\begin{subfigure}{.3\textwidth}
			\centering
			\includegraphics[width=\linewidth]{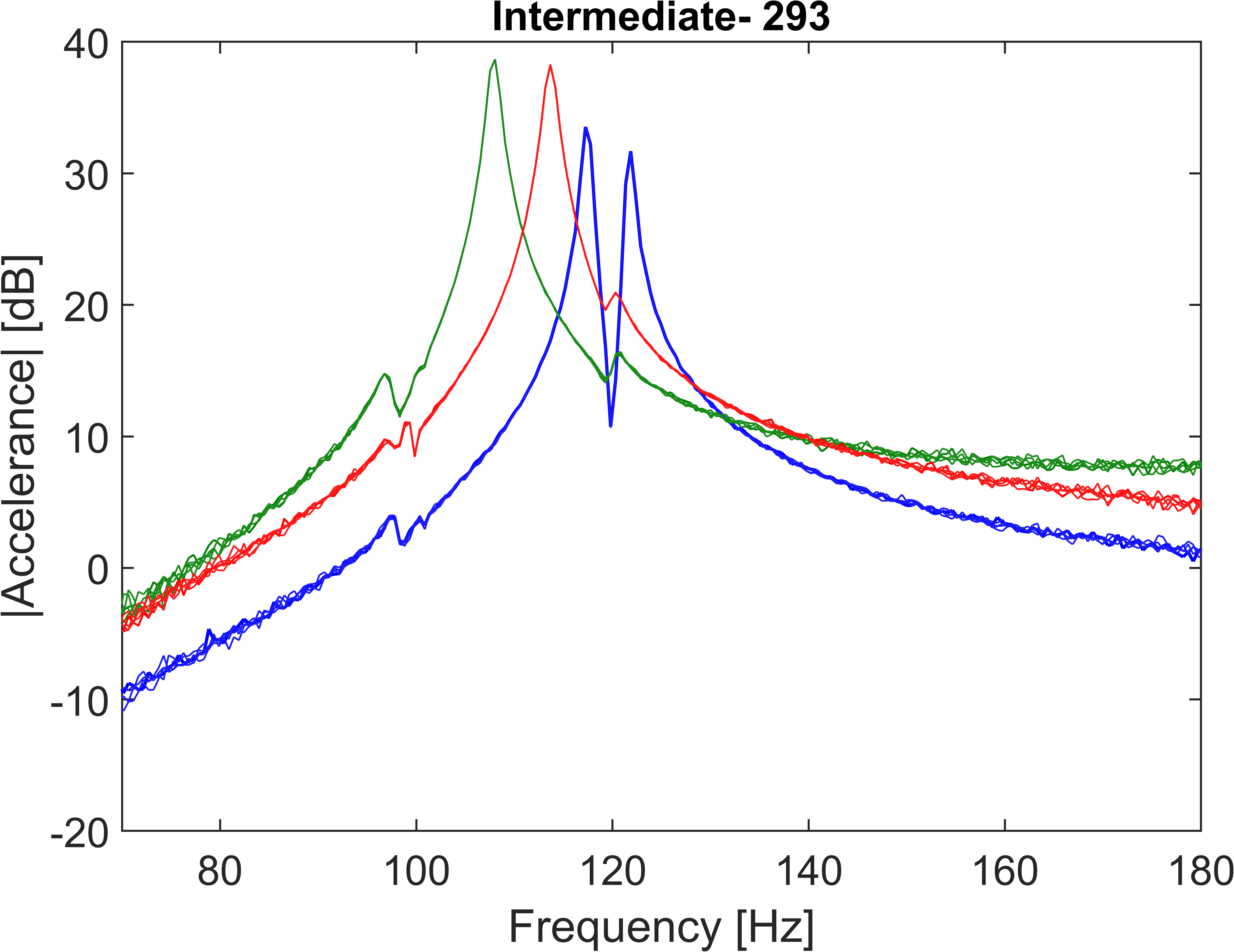}
		\end{subfigure} \\
		\vspace{1pt}
		\begin{subfigure}{.3\textwidth}
			\centering
			\includegraphics[width=\linewidth]{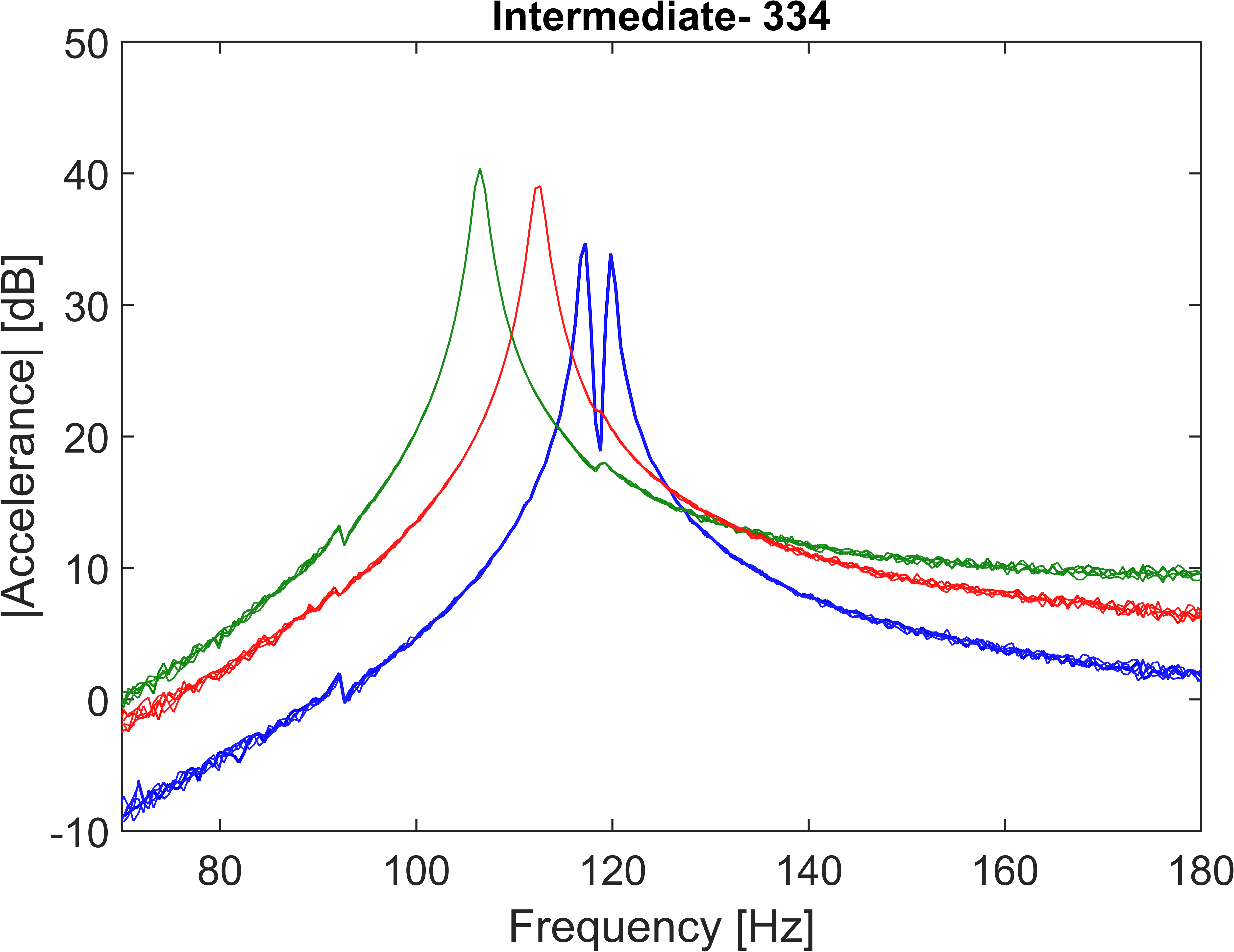}
		\end{subfigure} 
		\begin{subfigure}{.3\textwidth}
			\centering
			\includegraphics[width=\linewidth]{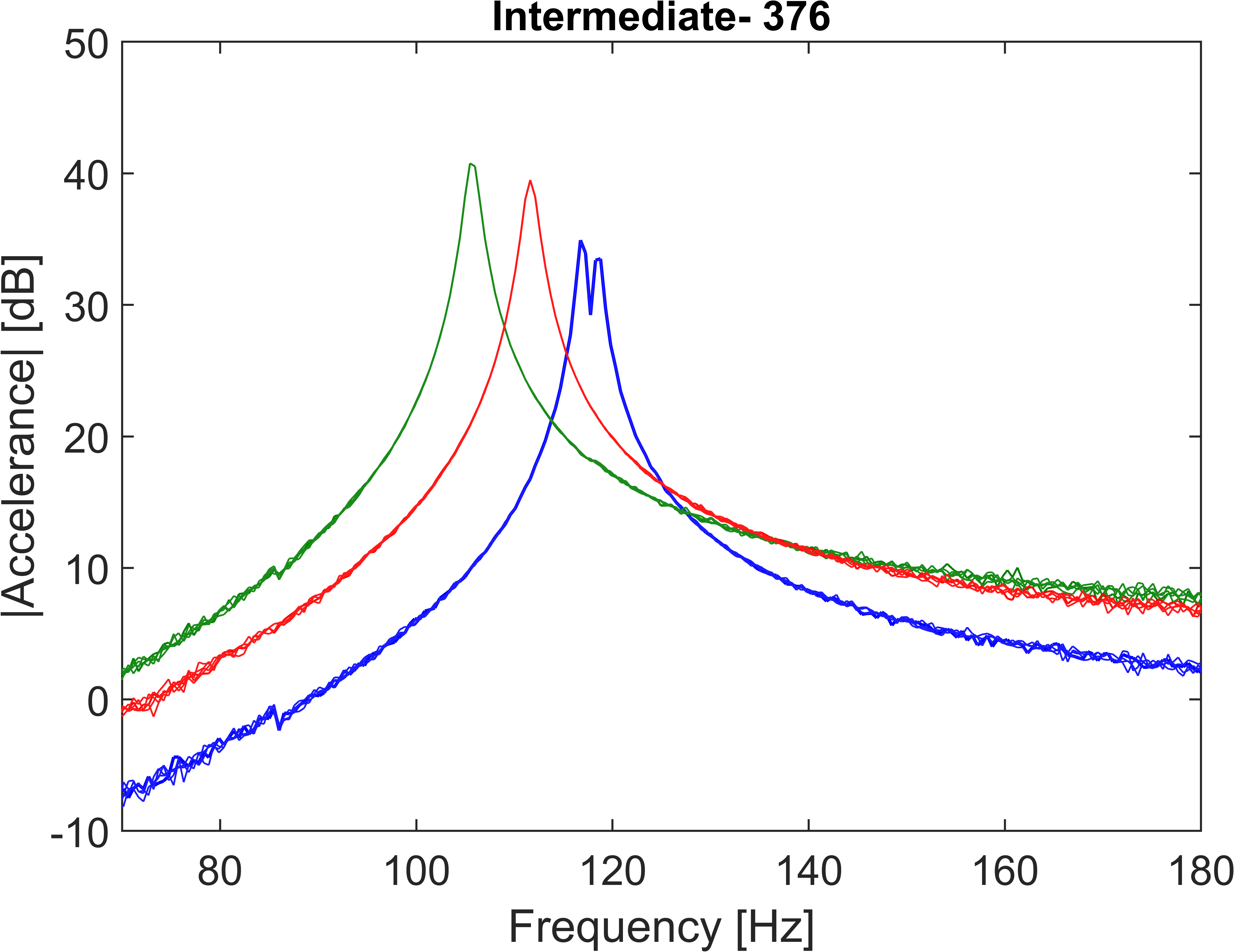}
		\end{subfigure}
		\vspace{1pt}
		\begin{subfigure}{.3\textwidth}
			\centering
			\includegraphics[width=\linewidth]{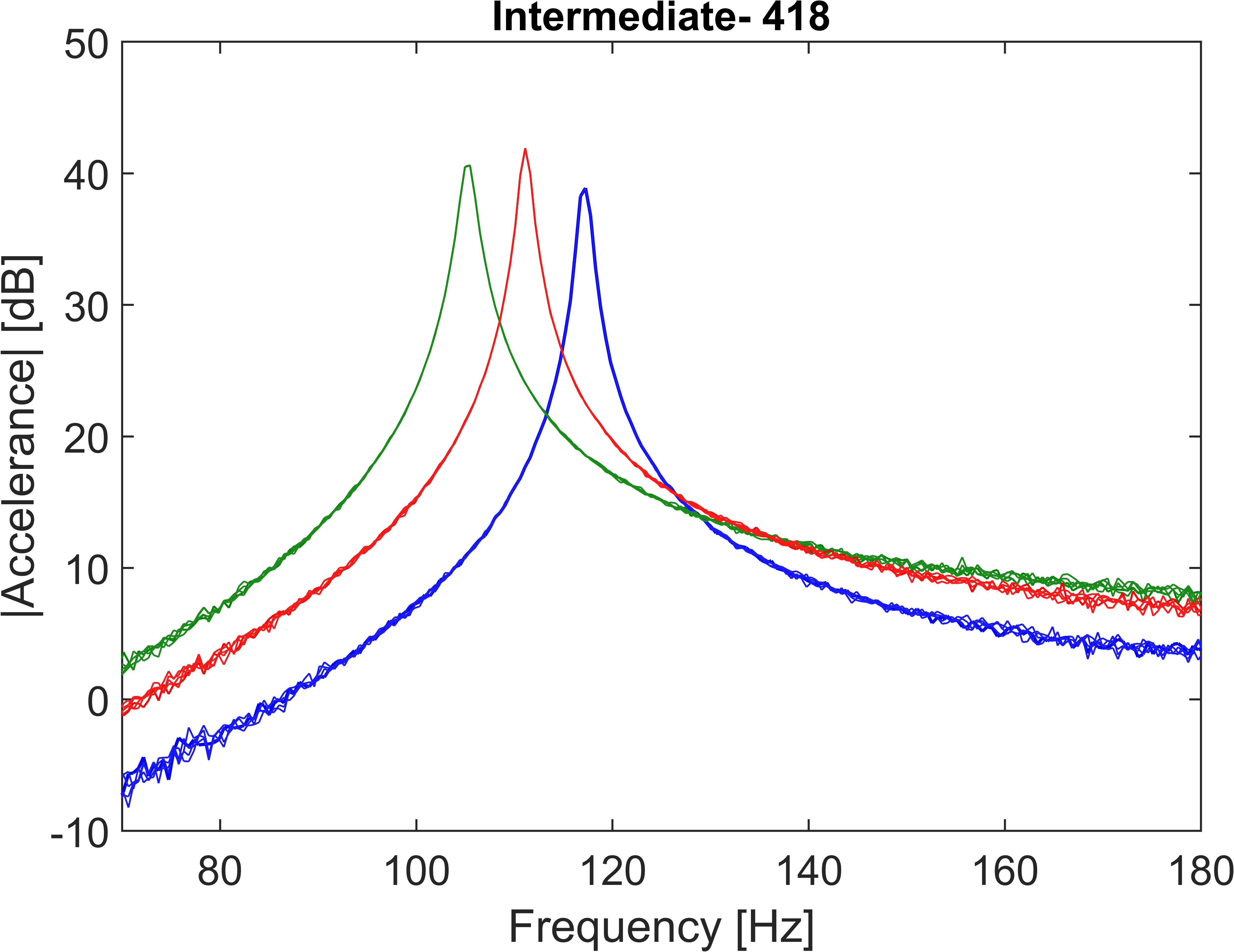}
		\end{subfigure} \\
		\vspace{1pt}
		\begin{subfigure}{.3\textwidth}
			\centering
			\includegraphics[width=\linewidth]{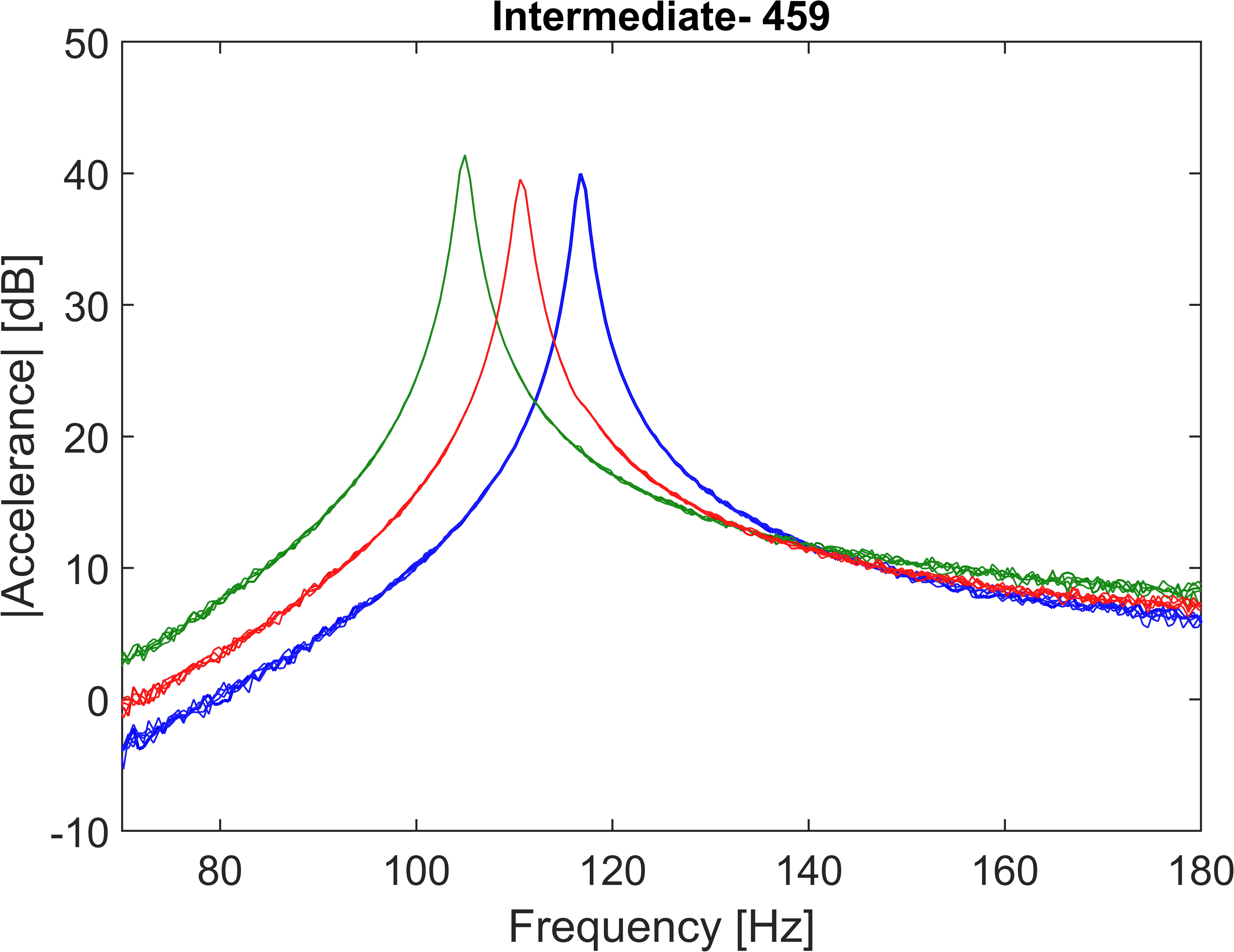}
		\end{subfigure} 
		\vspace{1pt}
			\begin{subfigure}{.3\textwidth}
			\centering
			\includegraphics[width=\linewidth]{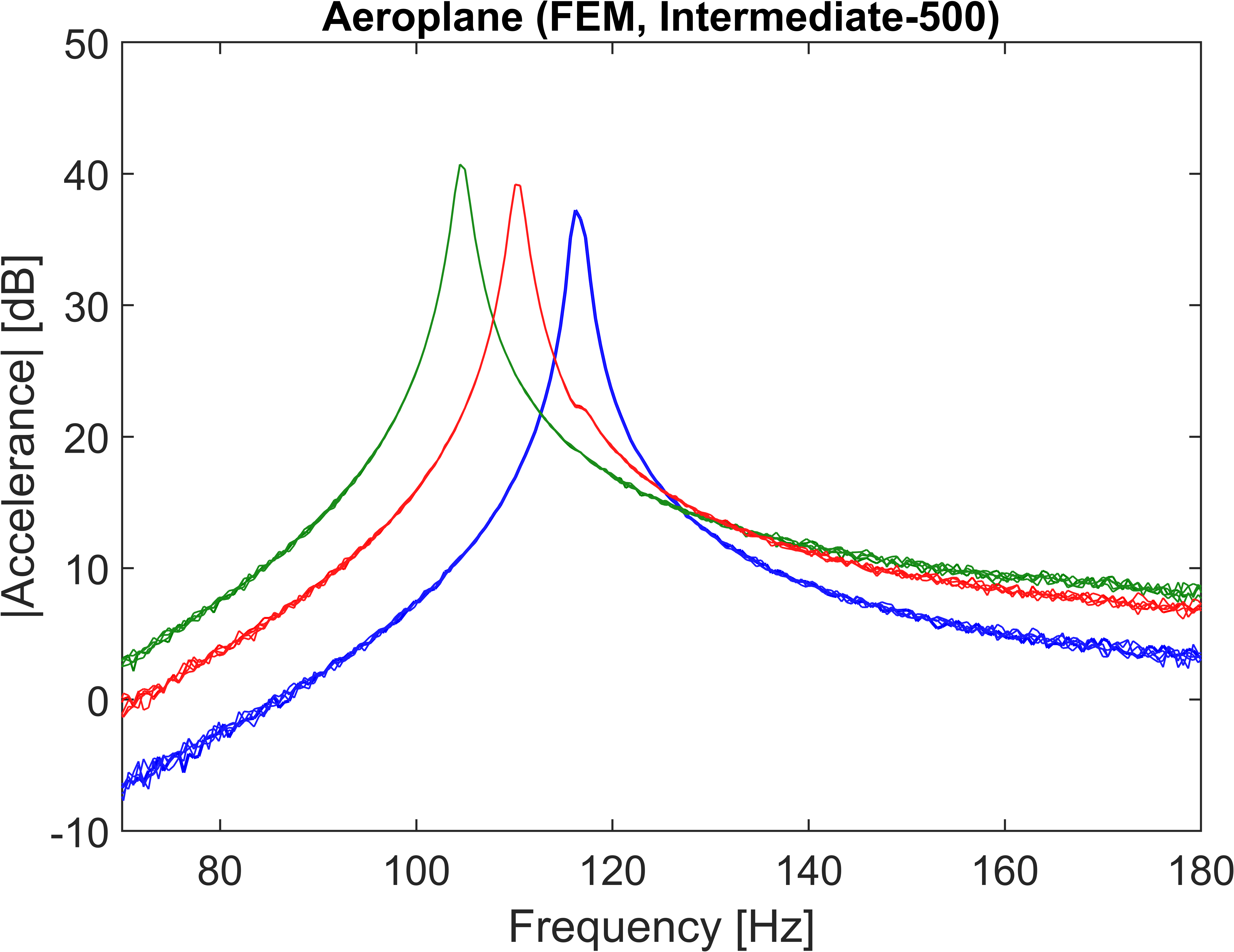}
		\end{subfigure}
		\vspace{1pt}
		\begin{subfigure}{.3\textwidth}
			\centering
			\includegraphics[width=\linewidth]{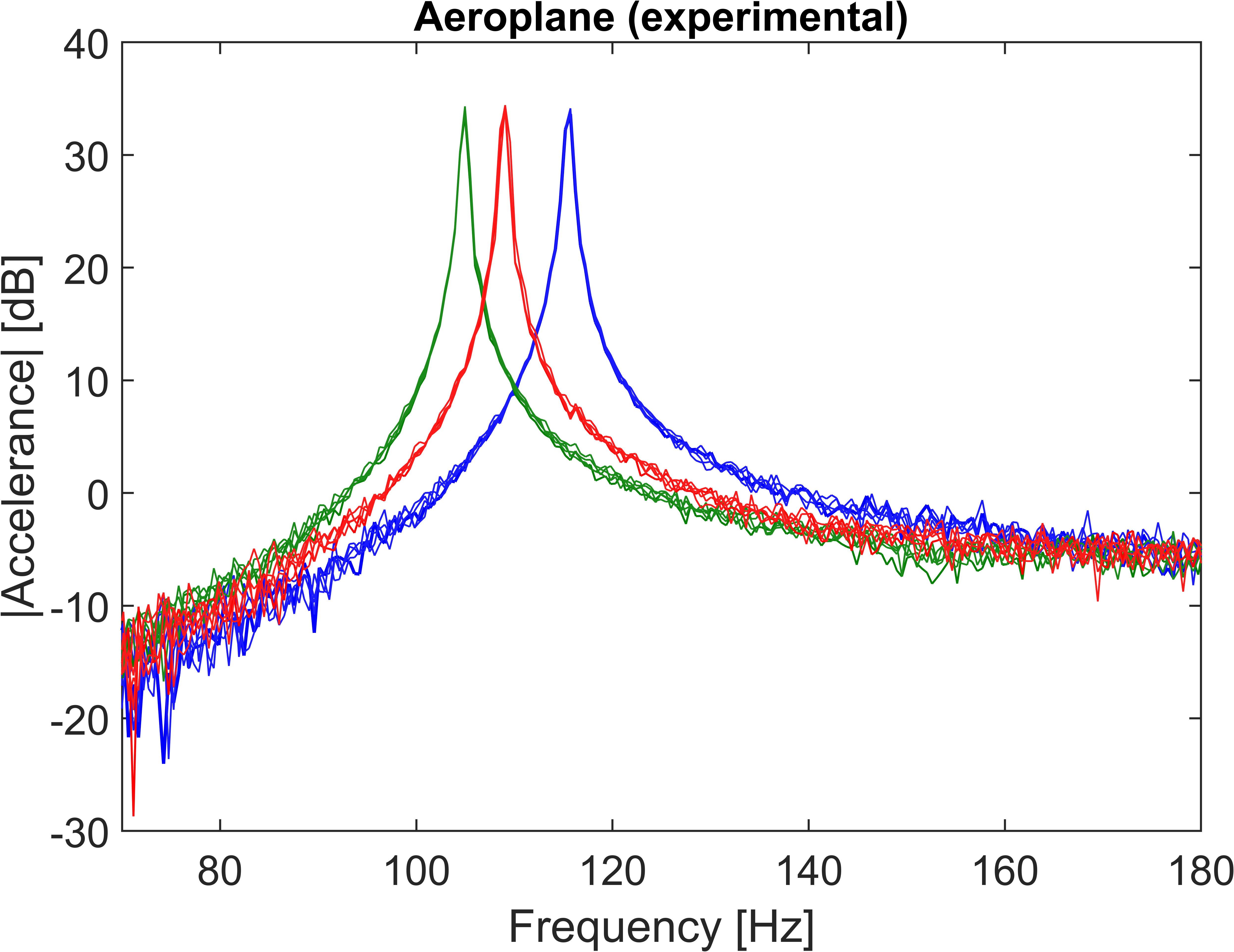}
		\end{subfigure}
		\vspace{1pt}
			\caption{FRFs at drive-point location for experimental bridge, subset of FEM intermediates, and experimental aeroplane. The blue, green, and red curves correspond to the healthy, D1, and D2 classes, respectively.}
			\label{fig:frfs-case-2}
	\end{figure}

\begin{figure}[ht!]
		\centering
		
		\begin{minipage}[t]{0.48\textwidth}
			\centering
			\begin{subfigure}{\linewidth}\centering
				\includegraphics[width=\linewidth, trim={8cm 8cm 6cm 7cm}, clip]{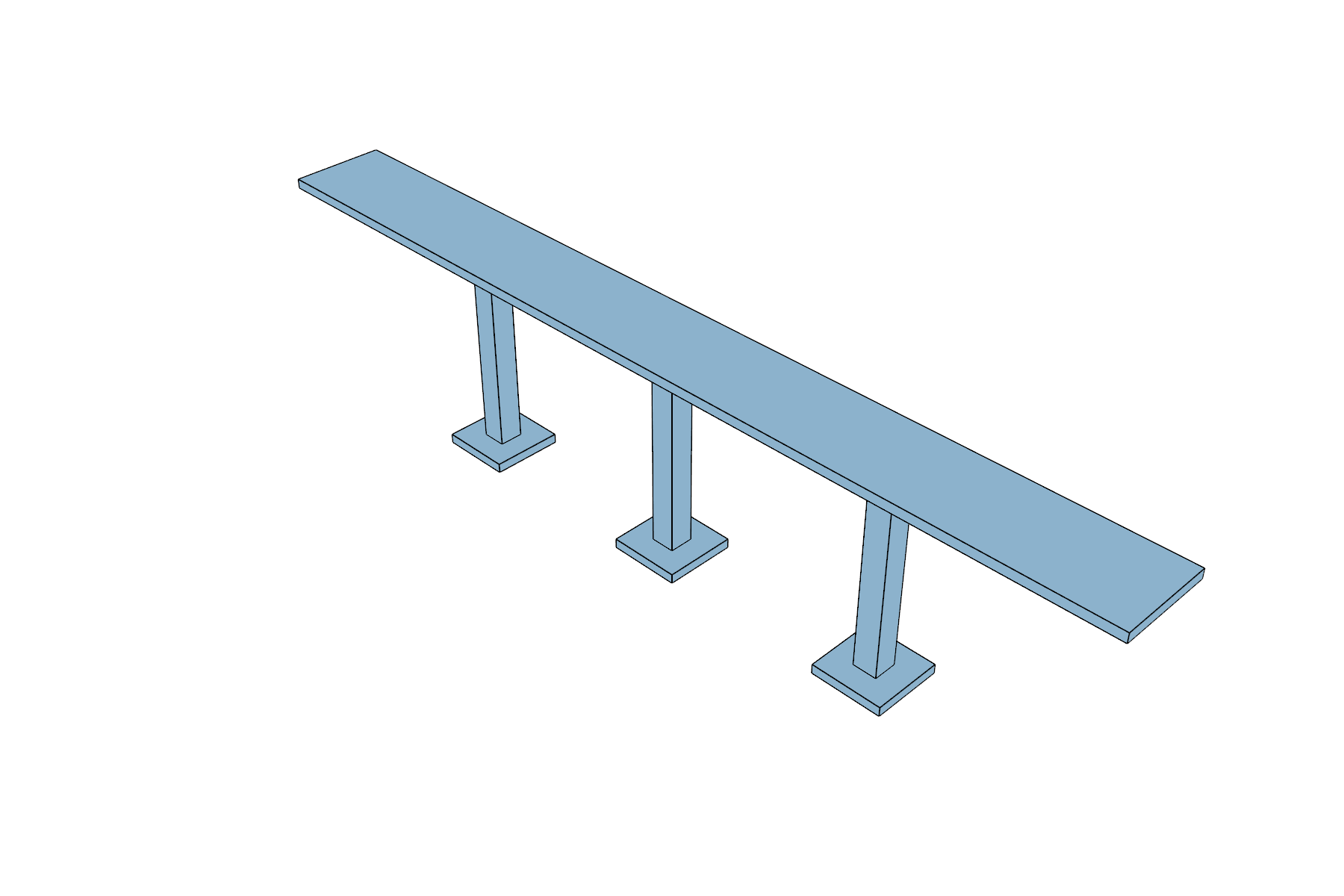}
			\end{subfigure}\par\vspace{3pt}
			\begin{subfigure}{\linewidth}\centering
				\includegraphics[width=\linewidth, trim={8cm 8cm 6cm 7cm}, clip]{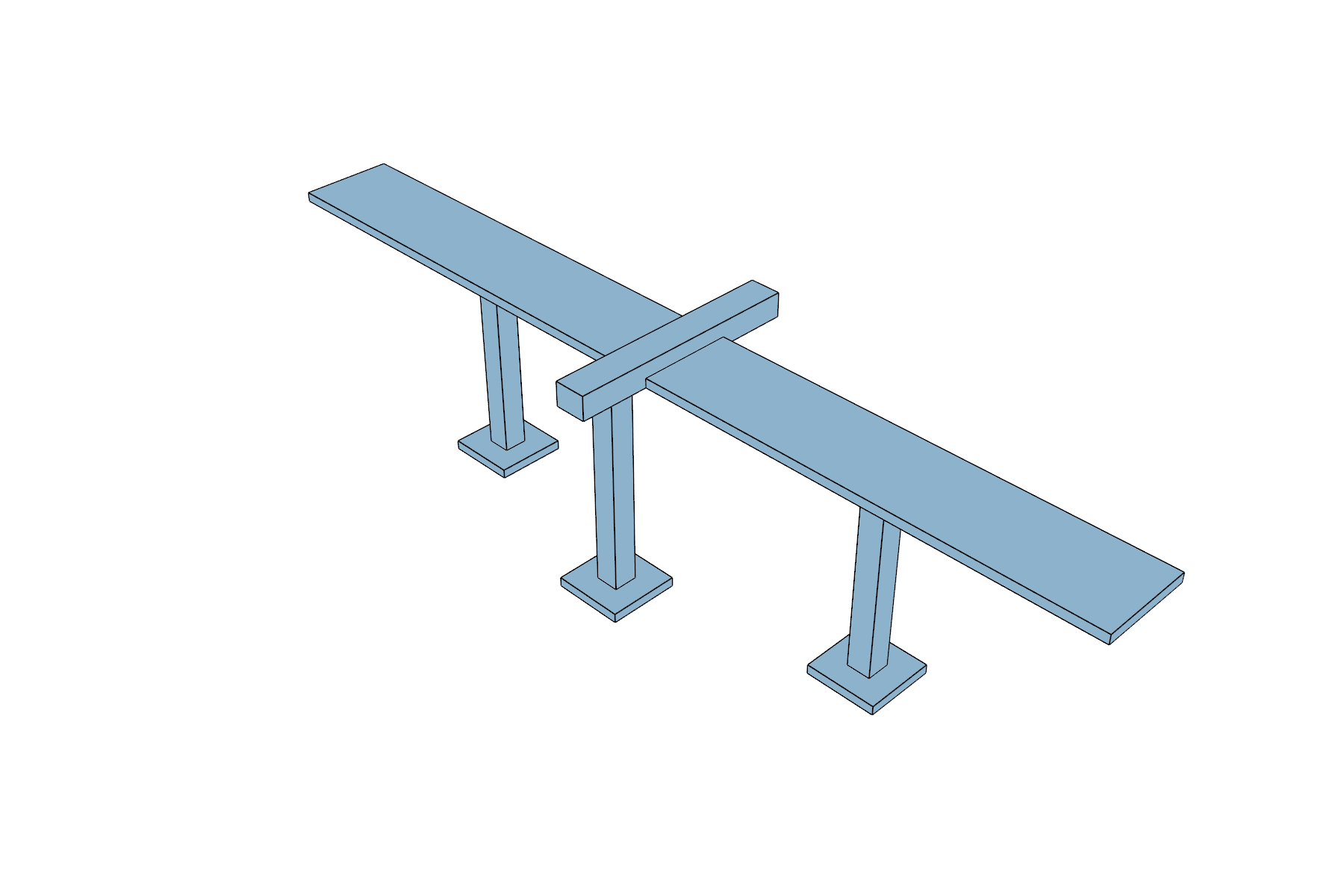}
			\end{subfigure}\par\vspace{3pt}
			\begin{subfigure}{\linewidth}\centering
				\includegraphics[width=\linewidth, trim={8cm 8cm 6cm 7cm}, clip]{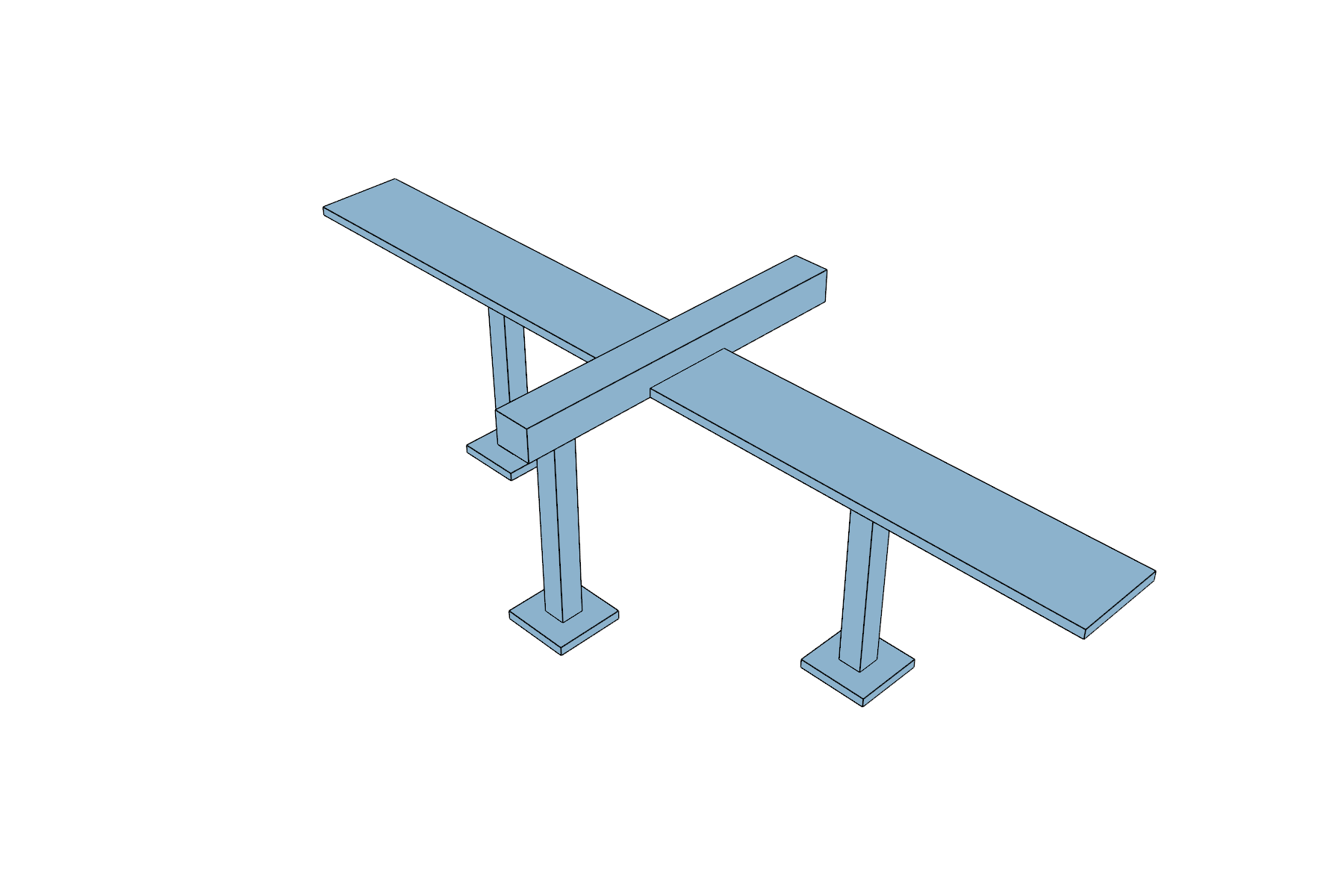}
			\end{subfigure}\par\vspace{3pt}
			\begin{subfigure}{\linewidth}\centering
				\includegraphics[width=\linewidth, trim={8cm 8cm 6cm 7cm}, clip]{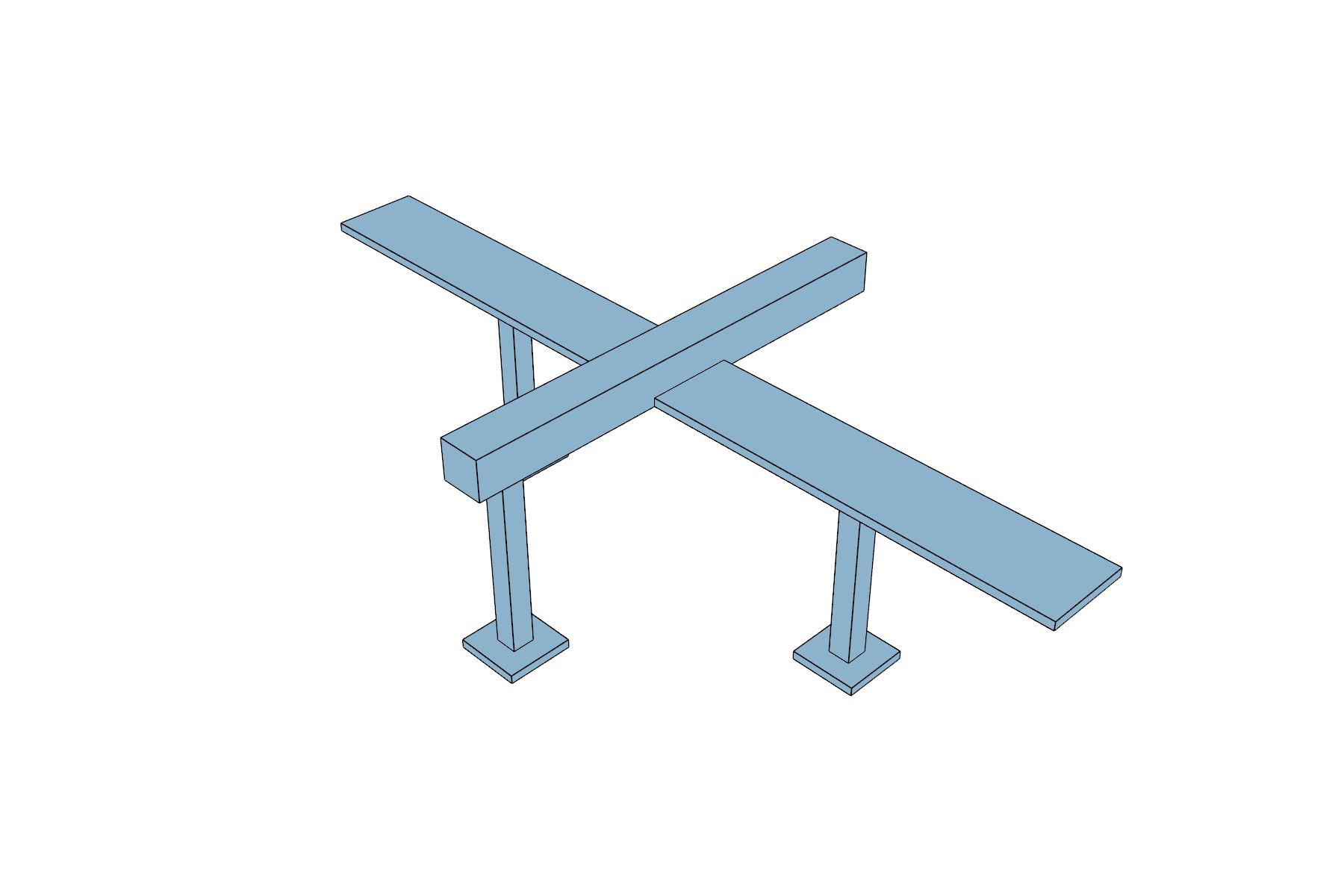}
			\end{subfigure}
		\end{minipage}\hfill
		\begin{minipage}[t]{0.48\textwidth}
			\centering
			\begin{subfigure}{\linewidth}\centering
				\includegraphics[width=\linewidth, trim={8cm 8cm 6cm 7cm}, clip]{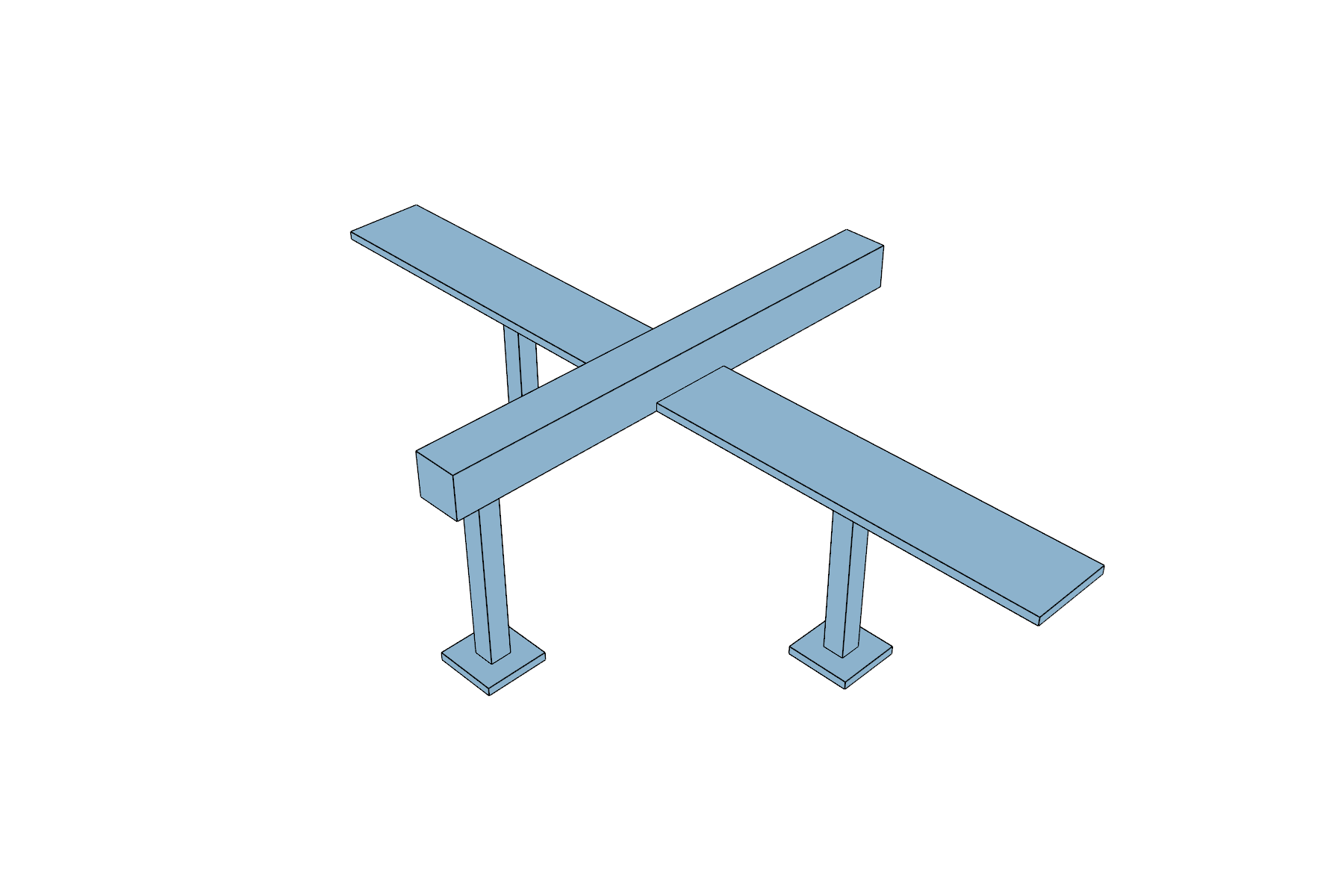}
			\end{subfigure}\par\vspace{3pt}
			\begin{subfigure}{\linewidth}\centering
				\includegraphics[width=\linewidth, trim={8cm 8cm 8cm 7cm}, clip]{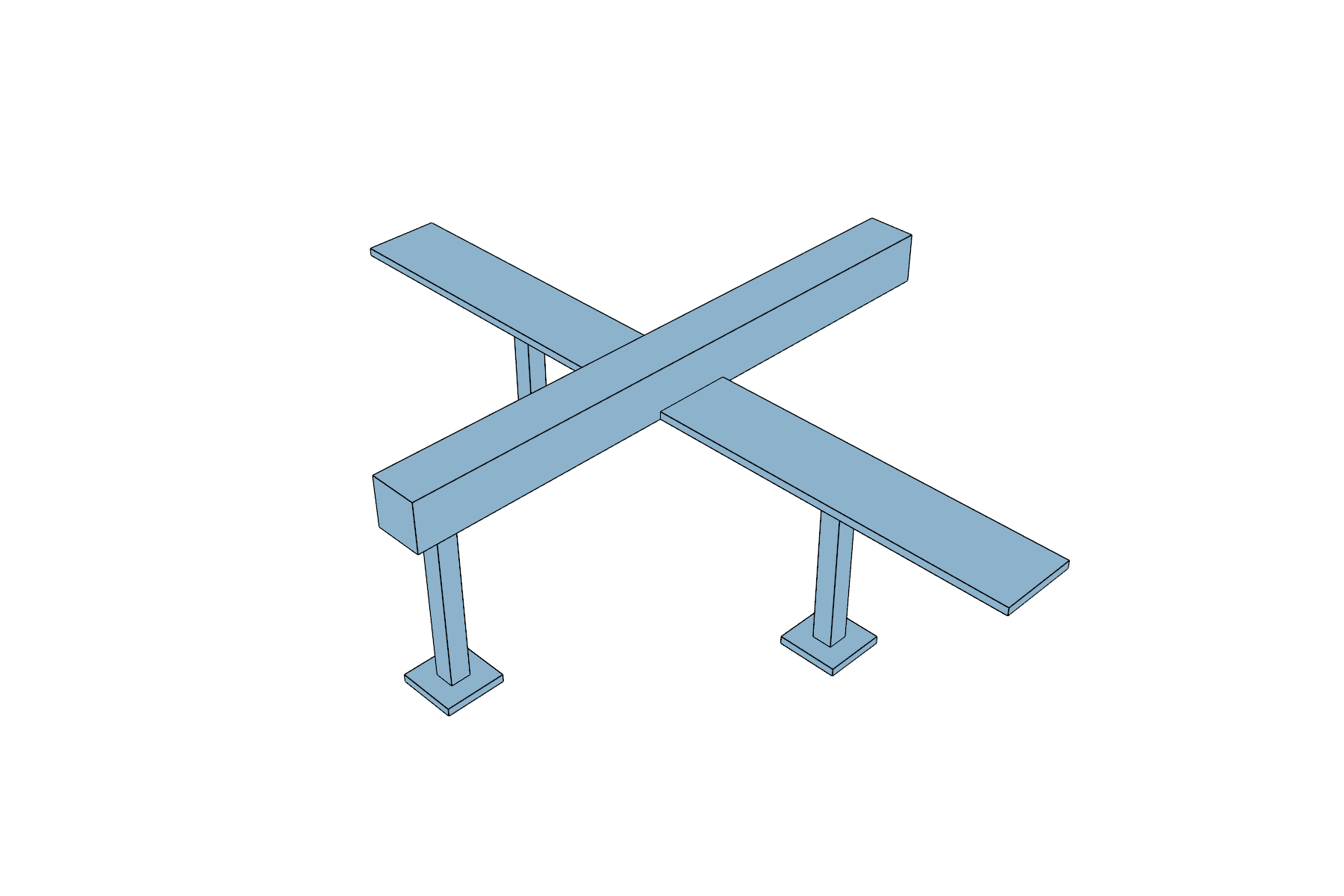}
			\end{subfigure}\par\vspace{3pt}
			\begin{subfigure}{\linewidth}\centering
				\includegraphics[width=\linewidth, trim={8cm 8cm 6cm 7cm}, clip]{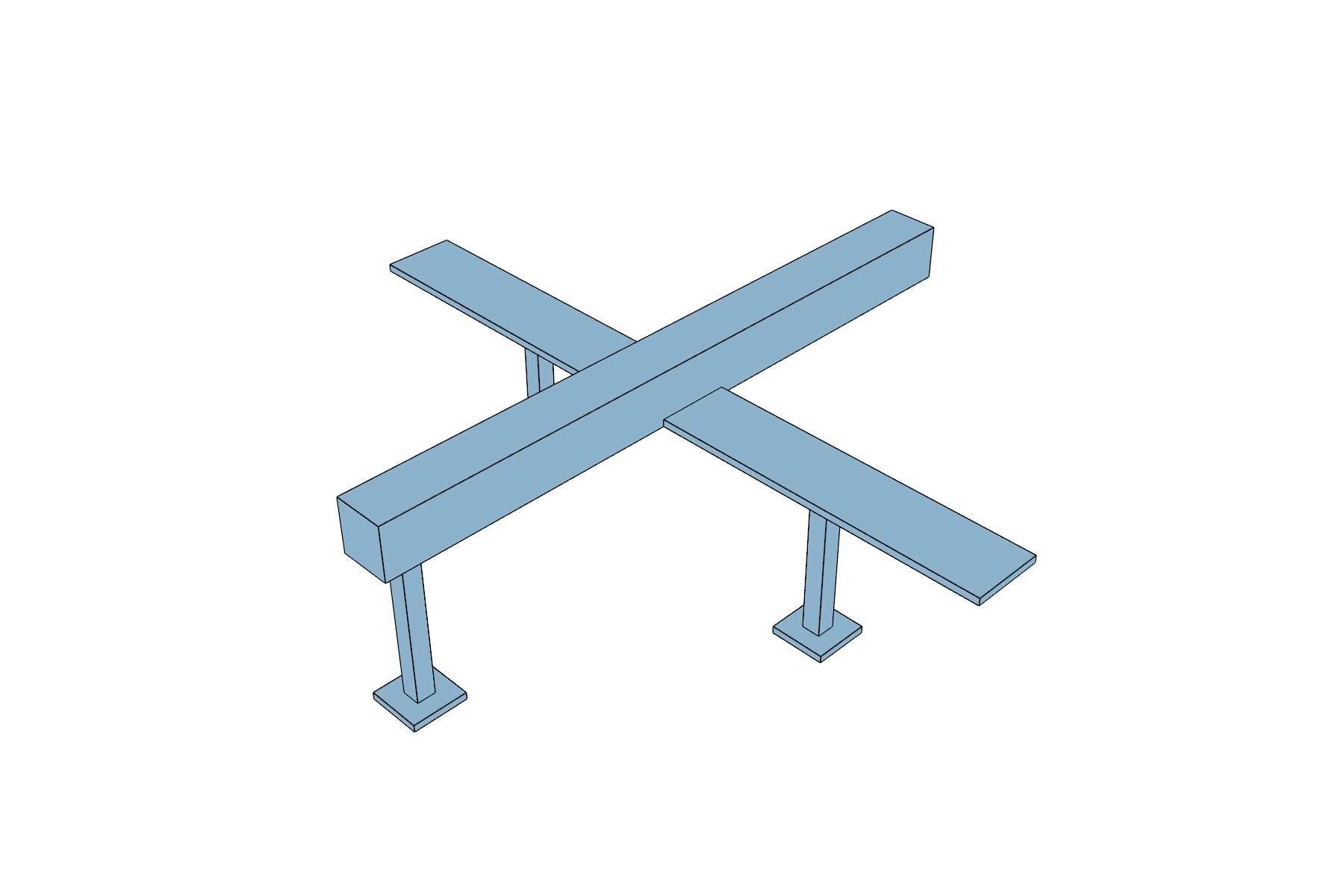}
			\end{subfigure}\par\vspace{3pt}
			\begin{subfigure}{\linewidth}\centering
				\includegraphics[width=\linewidth, trim={8cm 8cm 6cm 7cm}, clip]{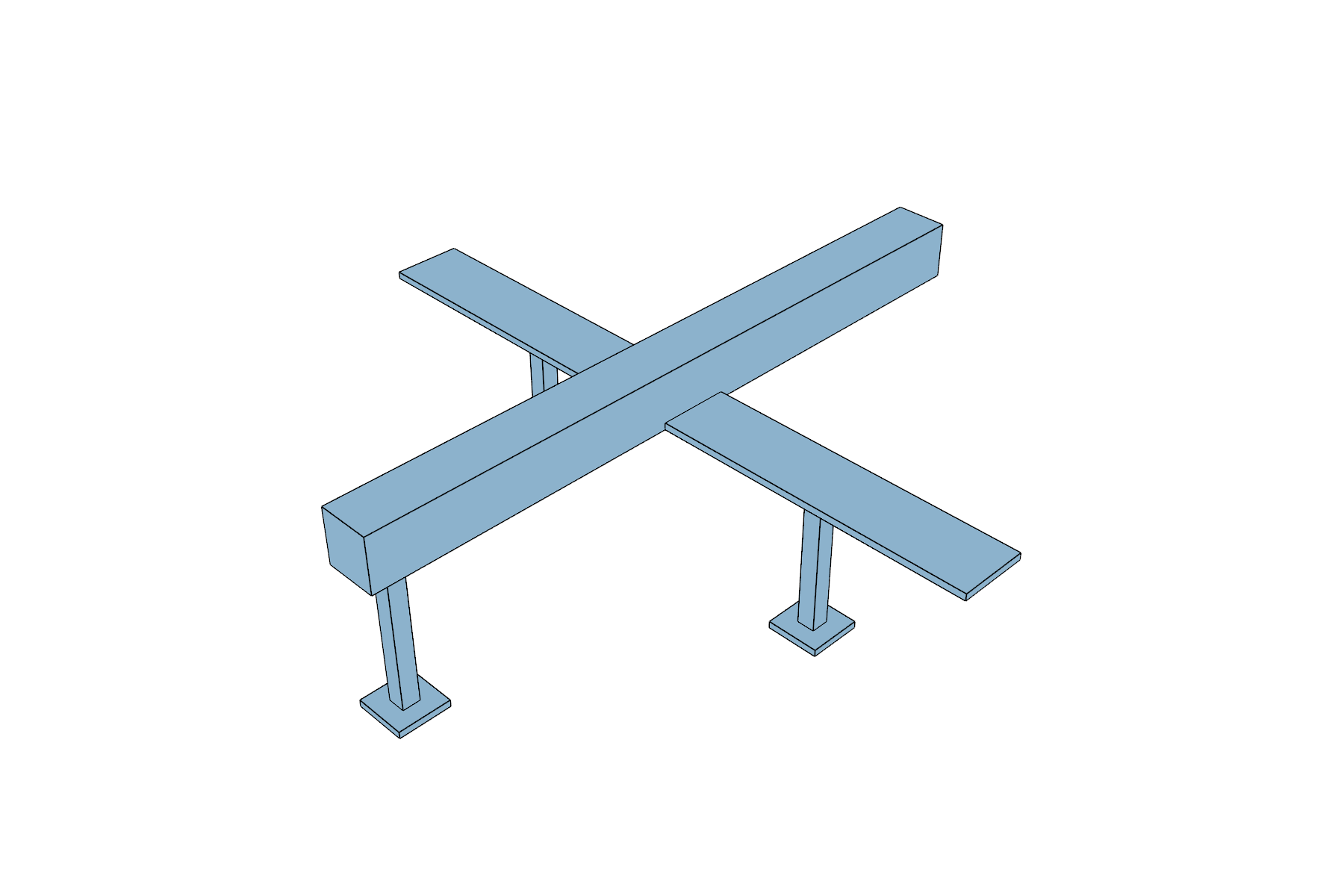}
			\end{subfigure}
		\end{minipage}
		
		\caption{Subset of intermediate geometries (1, 85, 168, 251, 293, 376, 459, and 500). Images are read top-to-bottom in the left column, then top-to-bottom in the right column.}
		\label{fig:intermediates-case-2}
	\end{figure}

	A total of 500 FEMs were generated, one each for the bridge and aeroplane configurations and 498 via progressively varying the fuselage length and the position of the middle support. (Provided that an initial model is available, this process is straightforward to automate, albeit potentially computationally expensive for a large number of intermediates.) FRFs were computed at the sensor locations for all FEMs using the same excitation and response points as in the experiments, and the FRF magnitudes over the selected frequency band were sampled on the same grid as the test data. This procedure ensured that experimental endpoints and FEM intermediates were represented in a consistent feature space. FRFs corresponding to the experimental bridge, a subset of the intermediate models, and the experimental aeroplane are shown in Figure \ref{fig:frfs-case-2}. The corresponding subset of intermediate FEM geometry is shown in Figure \ref{fig:intermediates-case-2}.
	
	\subsection{Transfer and results} \label{transfer-and-results}
	In this case study, only the ends of the chain were experimental: the physical bridge acted as the source domain and the physical aeroplane as the final target. The intermediate structures along the transfer path were provided entirely by the FEMs described in Section~\ref{FEMs}. For each structure (experimental endpoints and FEM intermediates) and each damage state, FRFs were computed at the sensor locations shown in Figure~\ref{fig:sensor-layout} and sampled on the same frequency grid within the band of interest. The magnitudes of these sampled FRFs, stacked over all selected frequencies and sensor locations, formed the feature vector for each test.
	
	The throughput time data were processed using a common pipeline for all structures and states. Around each selected hammer impact, a fixed-length block of 4000 samples was extracted with a pre-trigger of 100 samples. To generate multiple noisy replicates, each block of time data was replicated 10 times. For each repetition, zero-mean Gaussian noise was added to the responses, with the noise level sized to achieve an approximate post-window signal-to-noise ratio of 35 dB in the responses. The choice of 10 replicates provided a compromise between having enough datasets for robust training and keeping the computational cost manageable. After adding noise to the time data, a tapered exponential window was applied. FRFs were computed from the replicates using the H1 estimator and the magnitude of the FRFs was taken. This process resulted in each structure/model having 50 FRFs per sensor location, for each class. They were then used to construct normal- and damage-condition datasets at every structure along the chain. 
	
	Sensor 4 (location 4) was excluded from the transfer analyses because it exhibited a significant, class-specific inconsistency between the experimental and FEM bridges. Cosine alignment was computed between the experimental and FEM bridge FRFs for all sensors. Location 4 had a substantially lower similarity for the D1 class (0.45), whereas the similarity at the remaining locations was consistently high across classes (0.90–0.98). This indicates that, at location 4, the D1 FRFs differ between experiment and model in a way that is not representative of the remaining sensors. Transfer was performed independently for each sensor; as such, removing Sensor 4 did not affect the transfer results for the other channels. Sensor 4 was located adjacent to a support, where the response is particularly sensitive to support compliance and connection details. In the experiment, the supports were bolted to the deck, whereas in the FEMs they were modelled as fully fixed; this modelling simplification would be expected to increase discrepancy between the test and model responses at this location.

	The subsequent preprocessing and transfer pipeline followed the same overall structure as in the simulated bridge example in Case~1; however, NCA was not used to pre-process the data. Instead, a weighted RMS normalisation was used. For each dataset (source and target) and sensor, each FRF was scaled by a combination of its own row-wise RMS and a dataset-level reference RMS (computed over all FRFs in that dataset, including healthy and damaged conditions). Let $\mathrm{RMS}{_{\mathrm{row},n}}$ denote the RMS of the $n$-th FRF over the retained frequency bins, and let $\mathrm{RMS}{_\mathrm{ref}}$ denote the RMS computed over all FRFs in the dataset. The normalised FRF was then computed as,
	
	\begin{equation}
		\mathbf{H_{norm}} = \frac{\mathbf{H}}{\alpha\mathrm{RMS}{_\mathrm{row}}(\mathbf{H}) + (1-\alpha)\mathrm{RMS}{_\mathrm{ref}} + \varepsilon}
	\end{equation}

	\noindent This provided mild correction for the impact-to-impact amplitude variability noted in Section \ref{experiments} without over-normalising the individual FRFs, which would risk discarding class-related amplitude differences. No class information was used. The mixing parameter $\alpha=0.53$ was selected once using a source-only PCA stability criterion, by requiring the dominant PCA direction to remain sufficiently invariant under resampling (threshold $0.95$), and was then held fixed for all transfer experiments.
	
	Seven (7) measurement locations were considered, and the full transfer pipeline was executed independently for each location. This served the  purpose of enabling identification of locations that were more informative for damage identification, avoiding decision boundaries becoming unnecessarily complex as might happen with aggregation of heterogeneous sensor responses, and providing an internal consistency check on the transfer by comparing the performance of the models across locations. 
	
	As in the previous case, at each hop along a chain, classification was performed in two ways. For the first approach, PCA was fitted on the current source domain and a linear SVM was trained on the projected source data, then applied to the projected target data. For the second approach, a GFK was constructed from the source and target PCA subspaces at that hop, and a linear SVM was trained and evaluated in the resulting embedded space. Pseudo-labels and known healthy labels were then propagated hop-by-hop from the bridge to the aeroplane.	
		
\begin{figure}[htbp]
		\centering
		\begin{subfigure}{0.45\linewidth}
			\includegraphics[width=\linewidth]{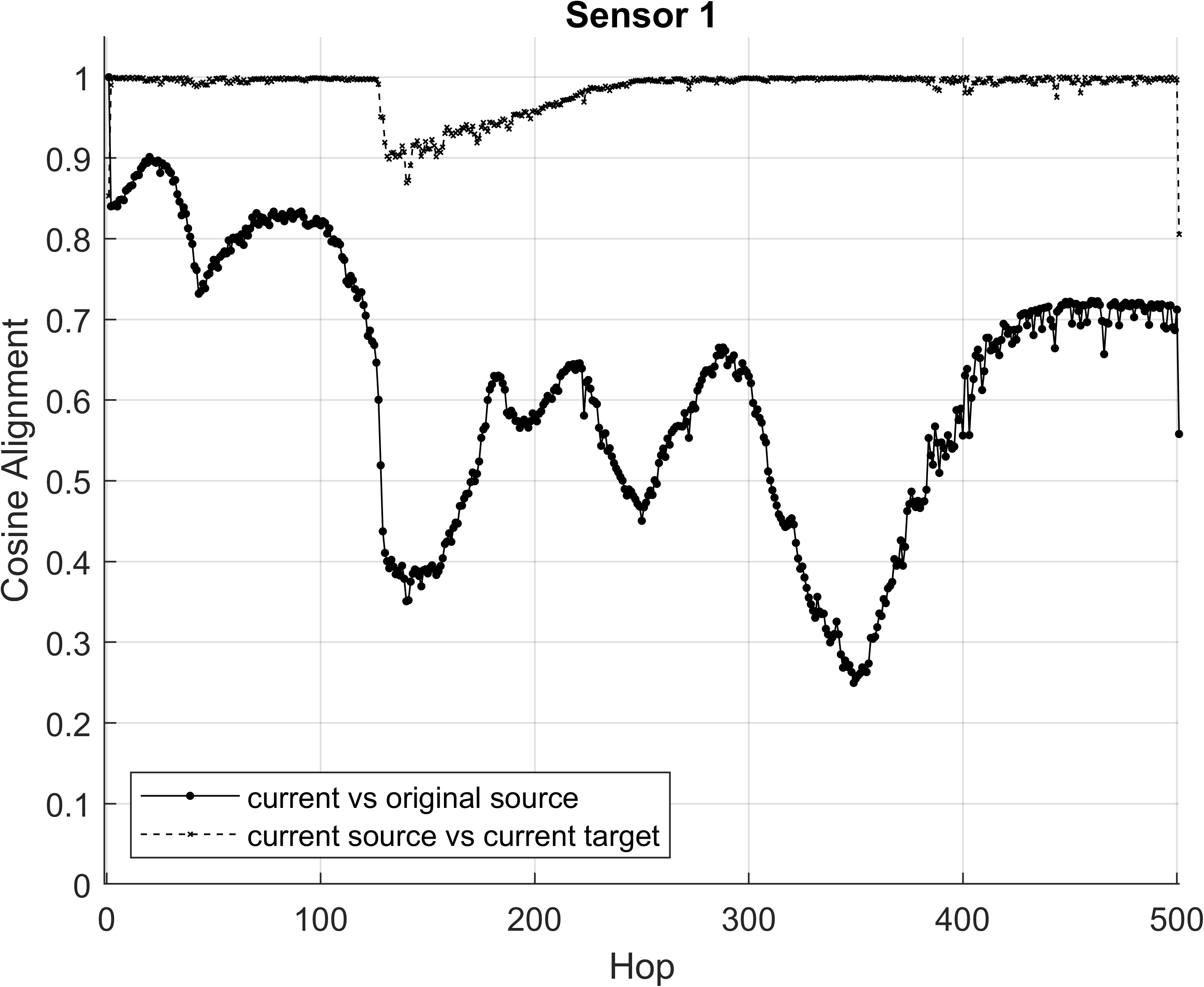}
		\end{subfigure}
		\hfill
		\begin{subfigure}{0.45\linewidth}
			\includegraphics[width=\linewidth]{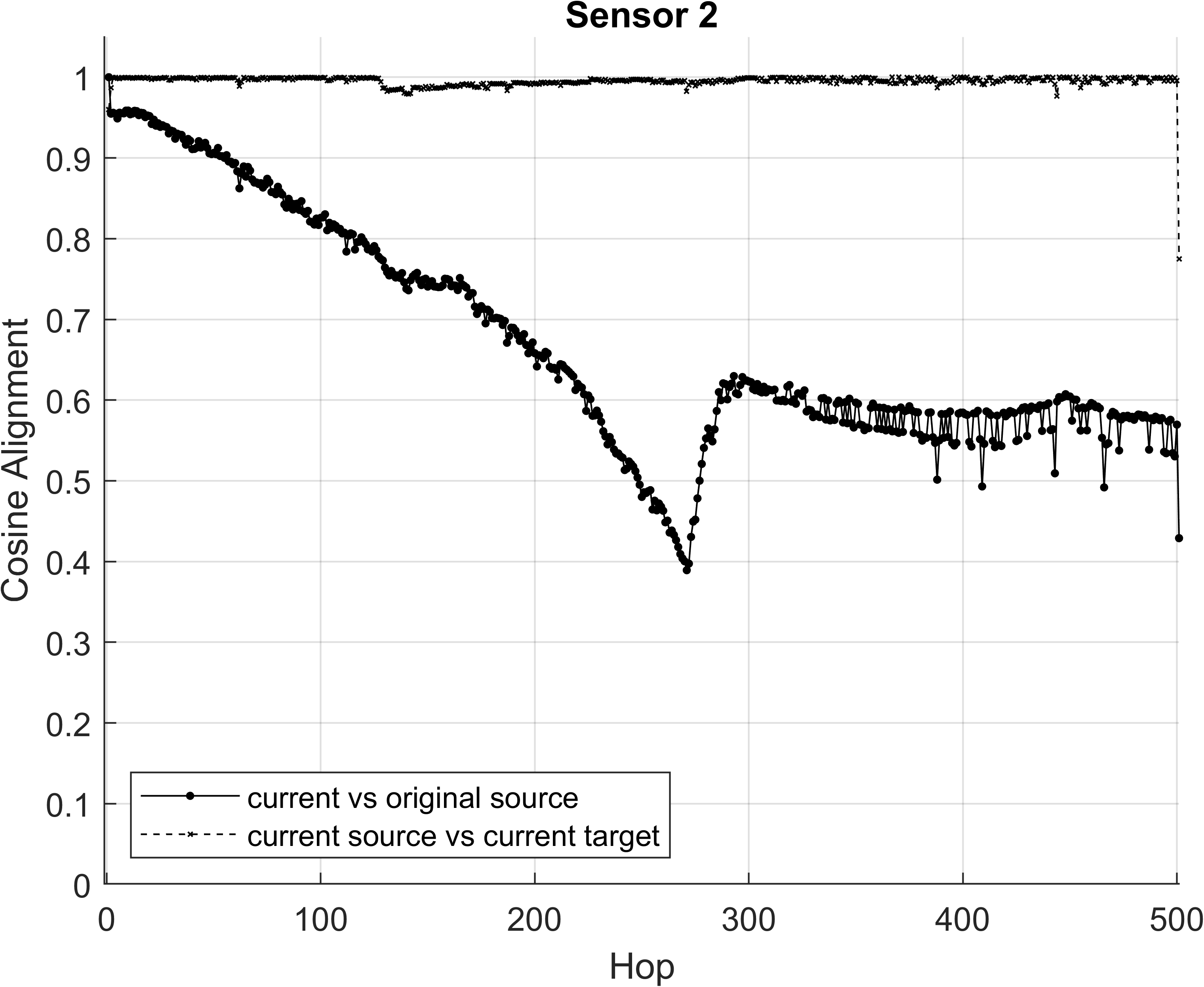}
		\end{subfigure}
		
		\vspace{2pt}
		
		\begin{subfigure}{0.45\linewidth}
			\includegraphics[width=\linewidth]{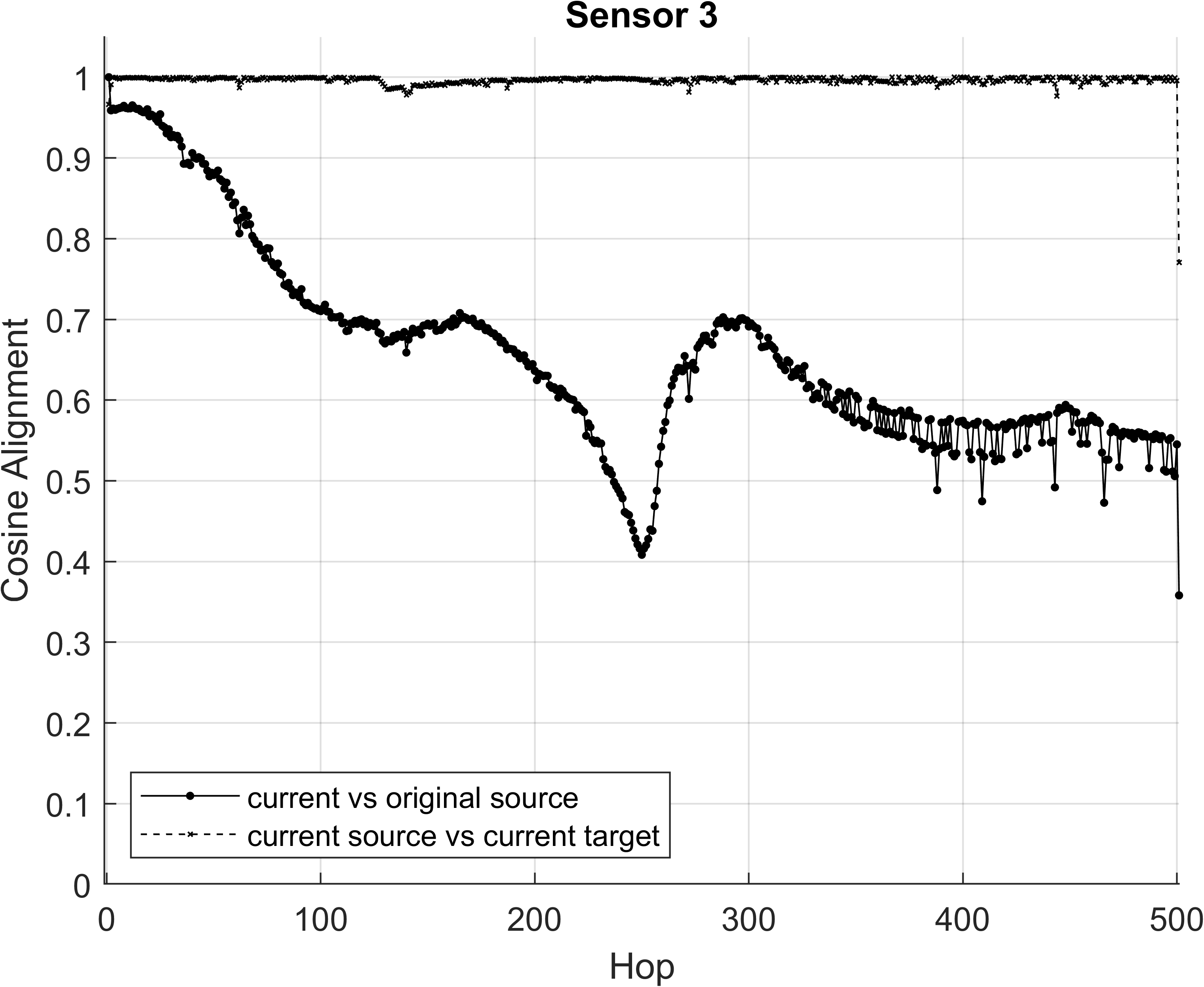}
		\end{subfigure}
		\hfill	
		\begin{subfigure}{0.45\linewidth}
			\includegraphics[width=\linewidth]{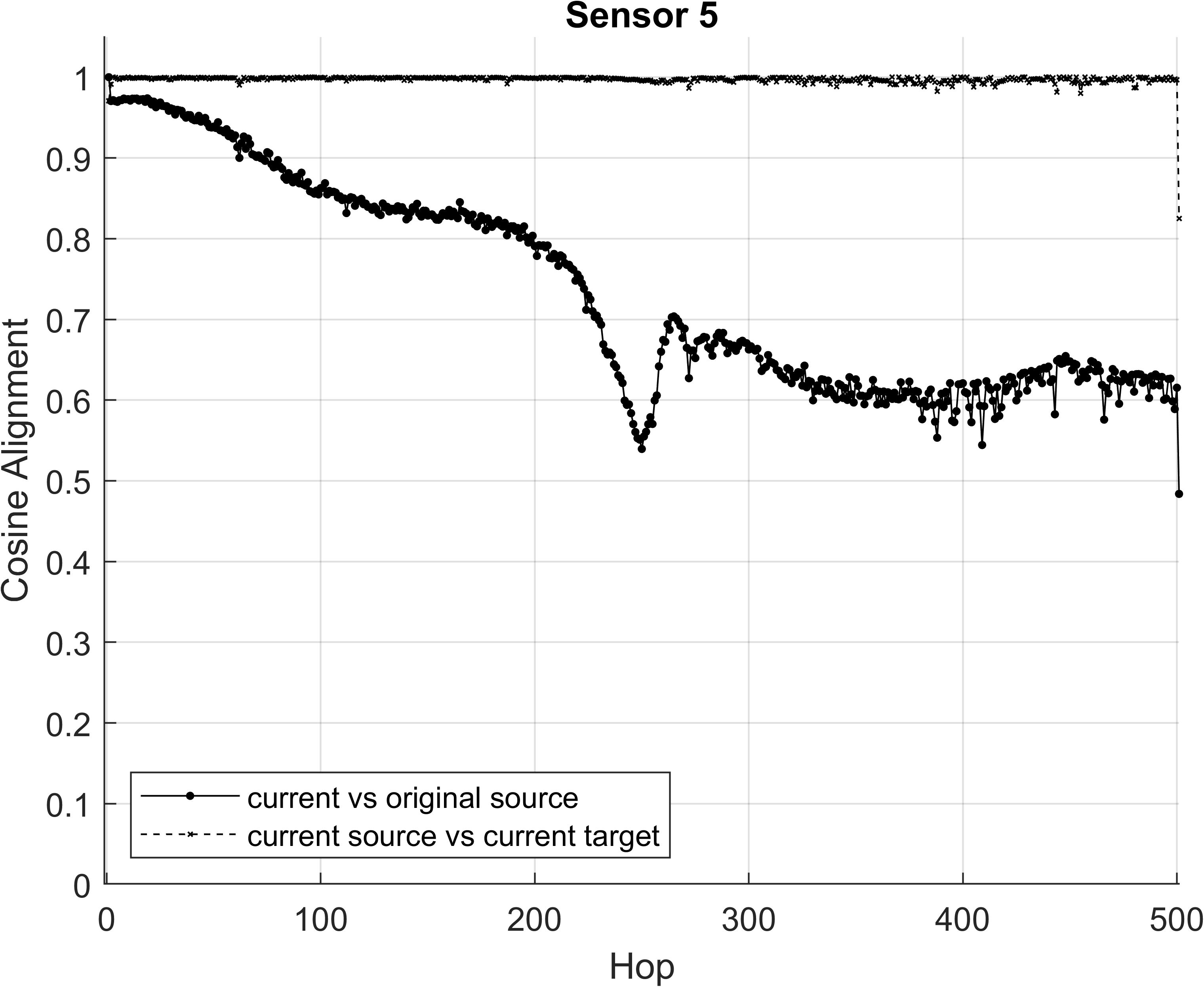}
		\end{subfigure}
	
		\vspace{2pt}
		
		\begin{subfigure}{0.45\linewidth}
			\includegraphics[width=\linewidth]{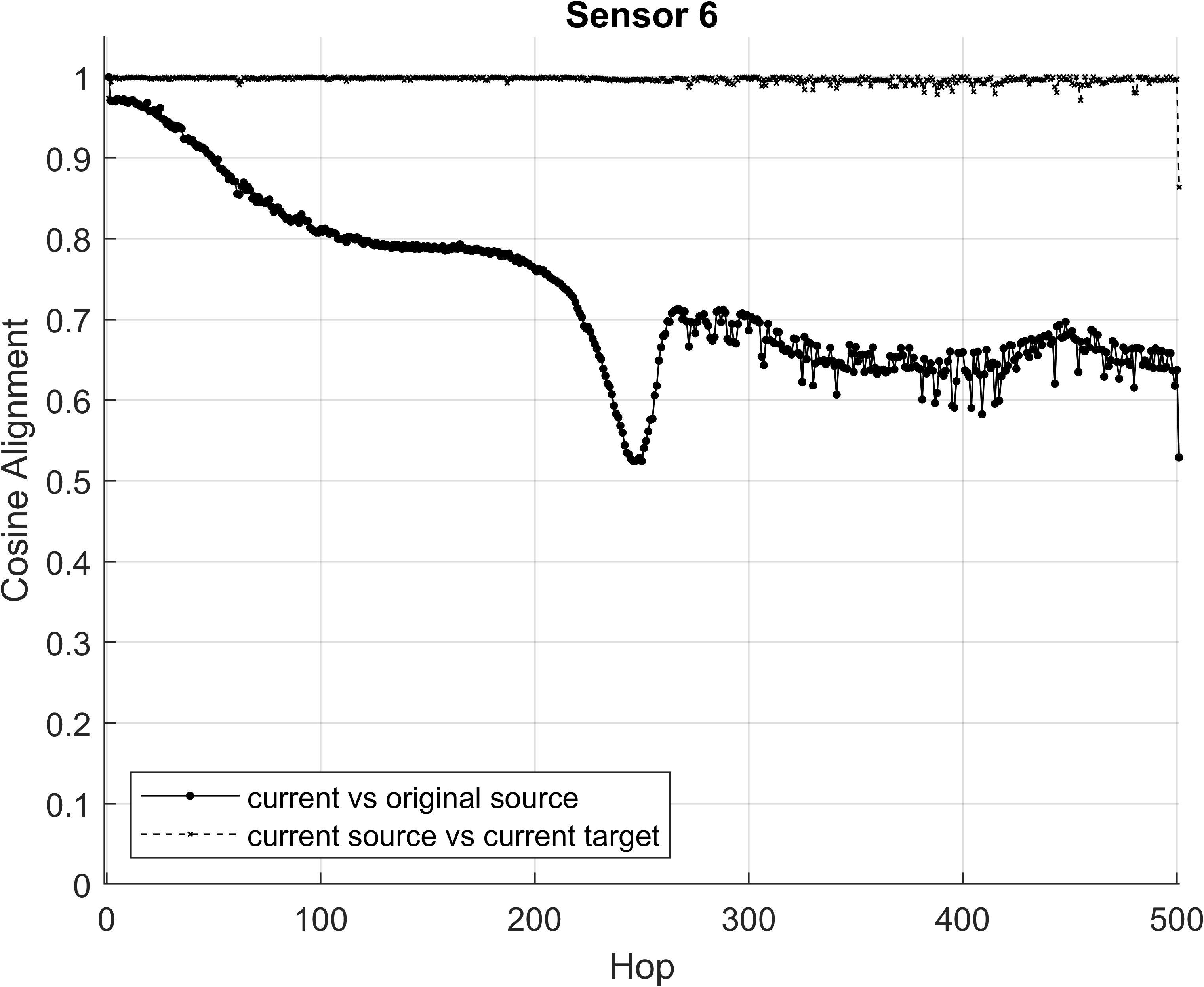}
		\end{subfigure}
		\begin{subfigure}{0.45\linewidth}
			\includegraphics[width=\linewidth]{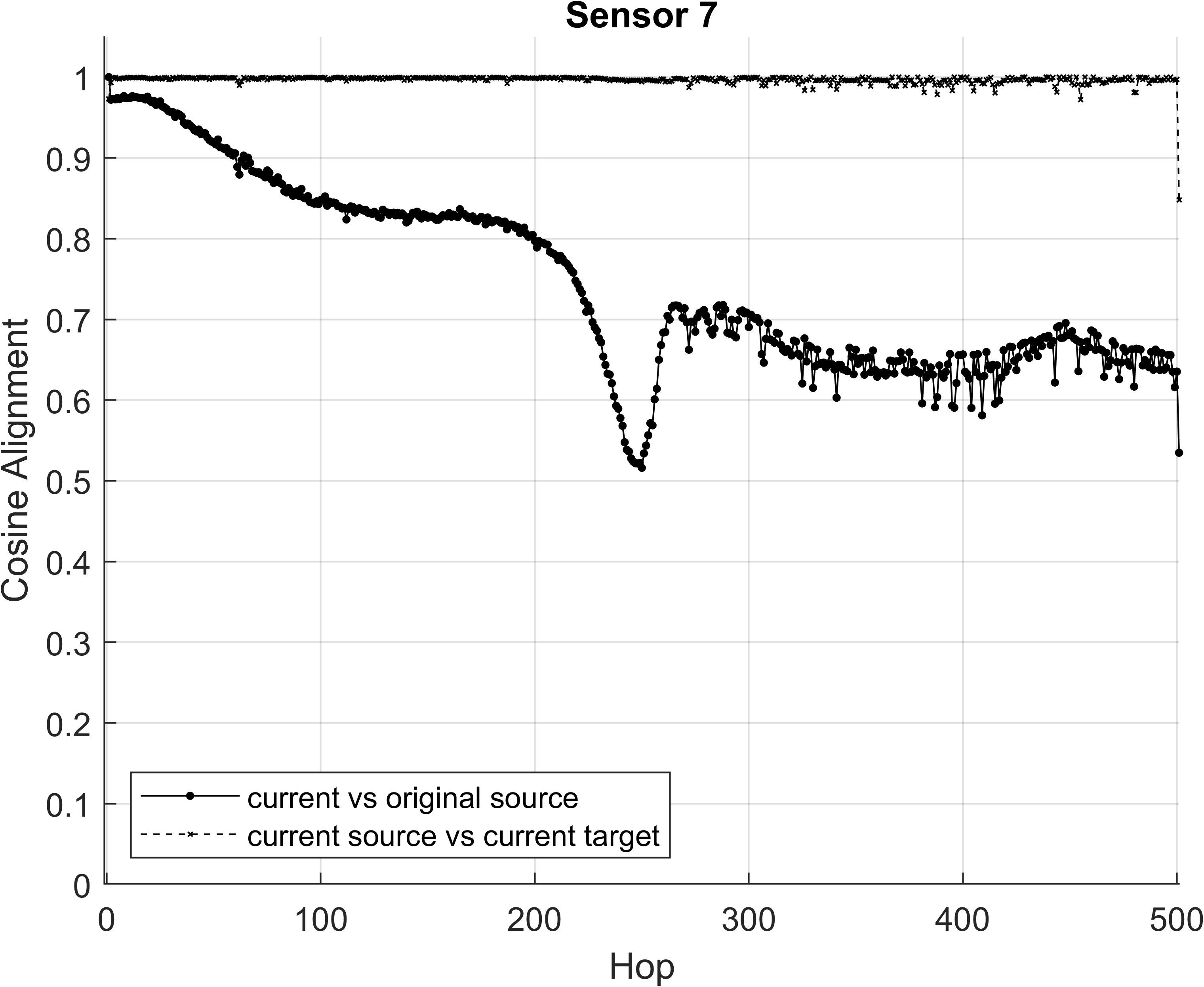}
		\end{subfigure}
	
		\vspace{2pt}
		
		\begin{subfigure}{0.45\linewidth}
			\includegraphics[width=\linewidth]{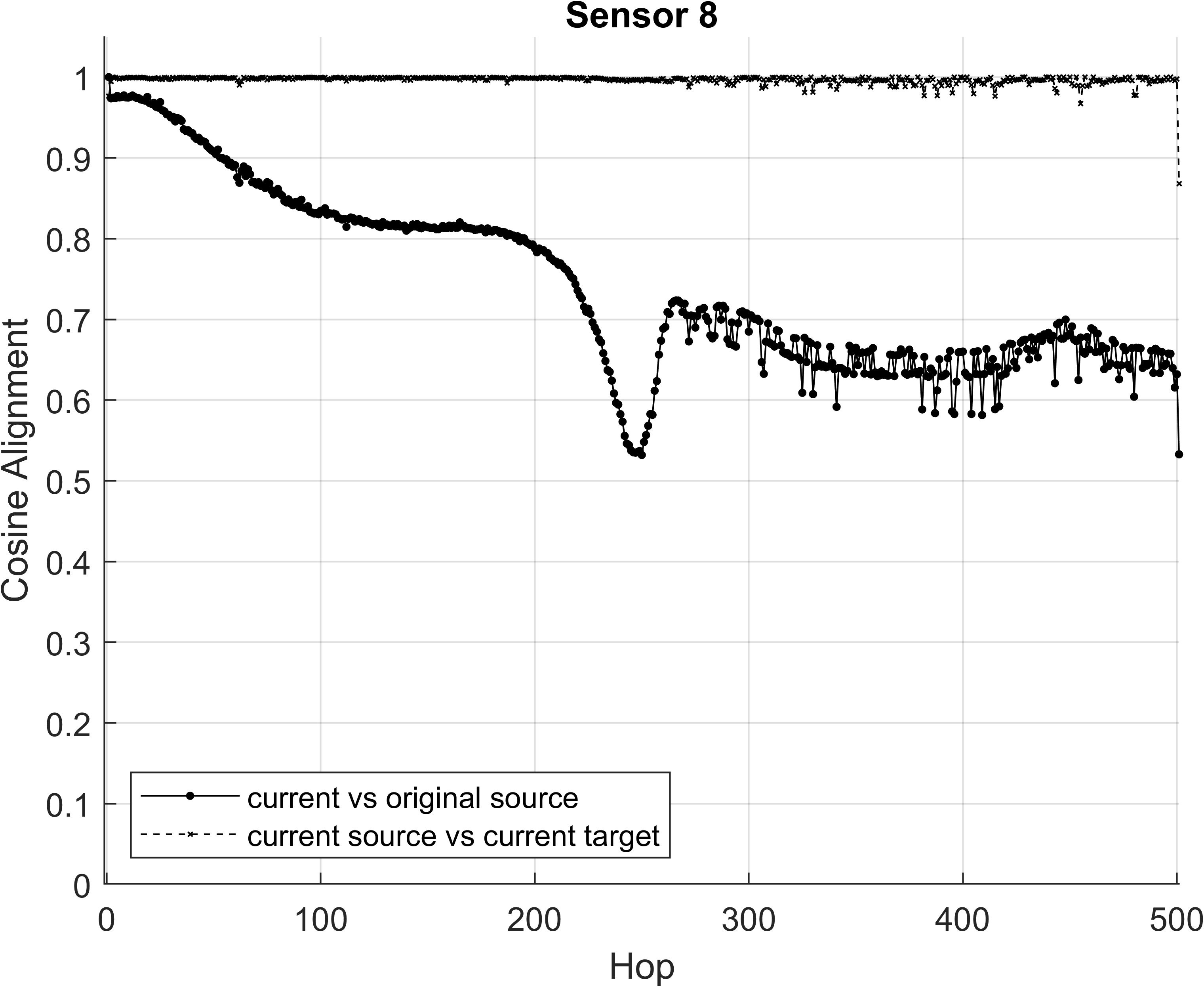}
		\end{subfigure}
		
		\vspace{1pt}
		
		\caption{Cosine alignment along the transfer path at each sensor location for Case 2. The alignment between the current source and target is shown in black with $\mathbf{\times}$ markers, with values close to 1 indicating that adjacent structures in the chain are consistently similar. The alignment between the current source and the original source is shown in black with $\bullet$ markers, with decreasing values showing progressively less similarity as the chain moves farther away from the original source.}
		\label{fig:dot-case-2}
	\end{figure}

	The evolution of the feature subspaces along the full set of intermediates was characterised in the same way as in the first case. At every hop, cosine alignments were computed between the leading principal directions of (1) the current source and current target and (2) the current source and the original source (the experimental bridge; with the final target included as the last point) to gain insight into the evolution of the principal angle directions with progression along the chain. The cosine alignment plots at each sensor location are shown in Figure \ref{fig:dot-case-2}.

	Across most locations, the local alignment (current source vs.\ current target) remained close to unity over the majority of hops, indicating that consecutive domains were typically close in the leading subspace direction and therefore transfer was possible at each hop. In contrast, the cumulative drift relative to the original source was overall monotonically-decreasing, which is expected, as the dominant principal directions gradually rotated from the bridge orientation to the aeroplane. A representative example is Location 7, where the cumulative drift decreased smoothly from near-perfect alignment at early hops to a lower plateau at later hops, with a pronounced transient drop and recovery around the 250th hop. This behaviour is consistent with a local region of the parametric path in which small geometric/material changes produce disproportionately large changes in the response, resulting from (for example) mode re-ordering or changes in modal participation at the measurement point. Importantly, despite this drop, transfer is still feasible provided that the chain is sufficiently refined along this region (i.e., by introducing additional intermediates or by adjusting the growth rates of the various parameters).
	
	Location 1 showed the most pronounced non-monotonic behaviour. This sensor was positioned closest to the portion of the structure undergoing the most substantial geometric growth (the fuselage), and its response is therefore more sensitive to local changes in modal participation as the fuselage length and associated stiffness/mass distribution evolve. As with the drop at the other locations, this behaviour did not prevent transfer along the chain but required enough intermediates to ensure that each model in the chain was sufficiently close to the next.
	
	Effective transfer paths were estimated via inspection of the cosine alignment plots. Intermediates were selected to maintain reasonable spacing along the chain while avoiding those associated with sharp local drops, as these indicate hops where the dominant PCA direction changed more abruptly than in neighbouring parts of the chain. For Sensors 2 through 8, the intermediates to avoid primarily fell between approximately 210 to 300, so a chain was constructed using $K=4$ intermediates 71, 201, 351, and 500 (so, effectively jumping over the difficult region). To verify that this region would be problematic for transfer, a second chain was constructed with $K=5$, with the addition of 251 to the chain listed above. For Sensor~1, sharp drops occurred in multiple regions (approximately 32-60, 110-175, 183-215, 221-284, and 296-395), so a chain was constructed using $K=9$ intermediates: 26, 31, 91, 101, 181, 219, 291, 401, 500. Transfer was also performed across all available ($K=500$) intermediates. 
	
	Performance was evaluated for each selected chain across multiple independent noise realisations. Rather than using an arbitrarily large number of repetitions, the number of realisations for a given chain was increased until the Standard Error of the Mean (SEM) of the accuracy fell below 1\%. This stopping rule was used because repeated runs are computationally expensive, particularly for long chains. Under this criterion, transfer without intermediates required 400 repetitions to achieve $\mathrm{SEM}$ less than 1\%. The $K=4$ chain required 25 repetitions, the $K=5$ chain 350 repetitions, the $K=9$ chain 385 repetitions, and the full chain ($K=500$) required 25 repetitions. 
	
\begin{figure}[htbp]
		\centering
		\subfloat[\label{fig:conf1}]{\includegraphics[width=.45\textwidth]{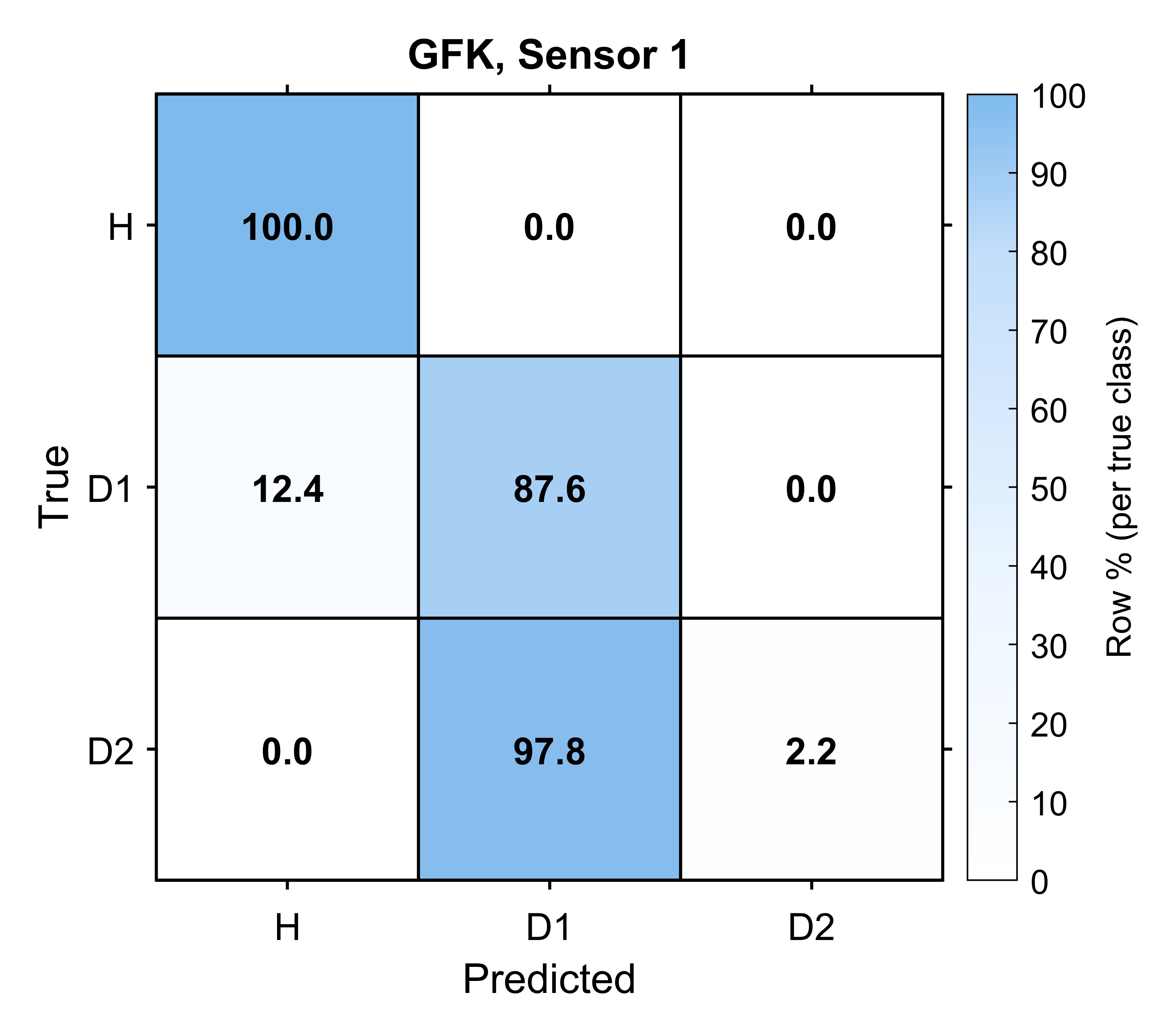}}\hfill
		\subfloat[\label{fig:conf2}]{\includegraphics[width=.45\textwidth]{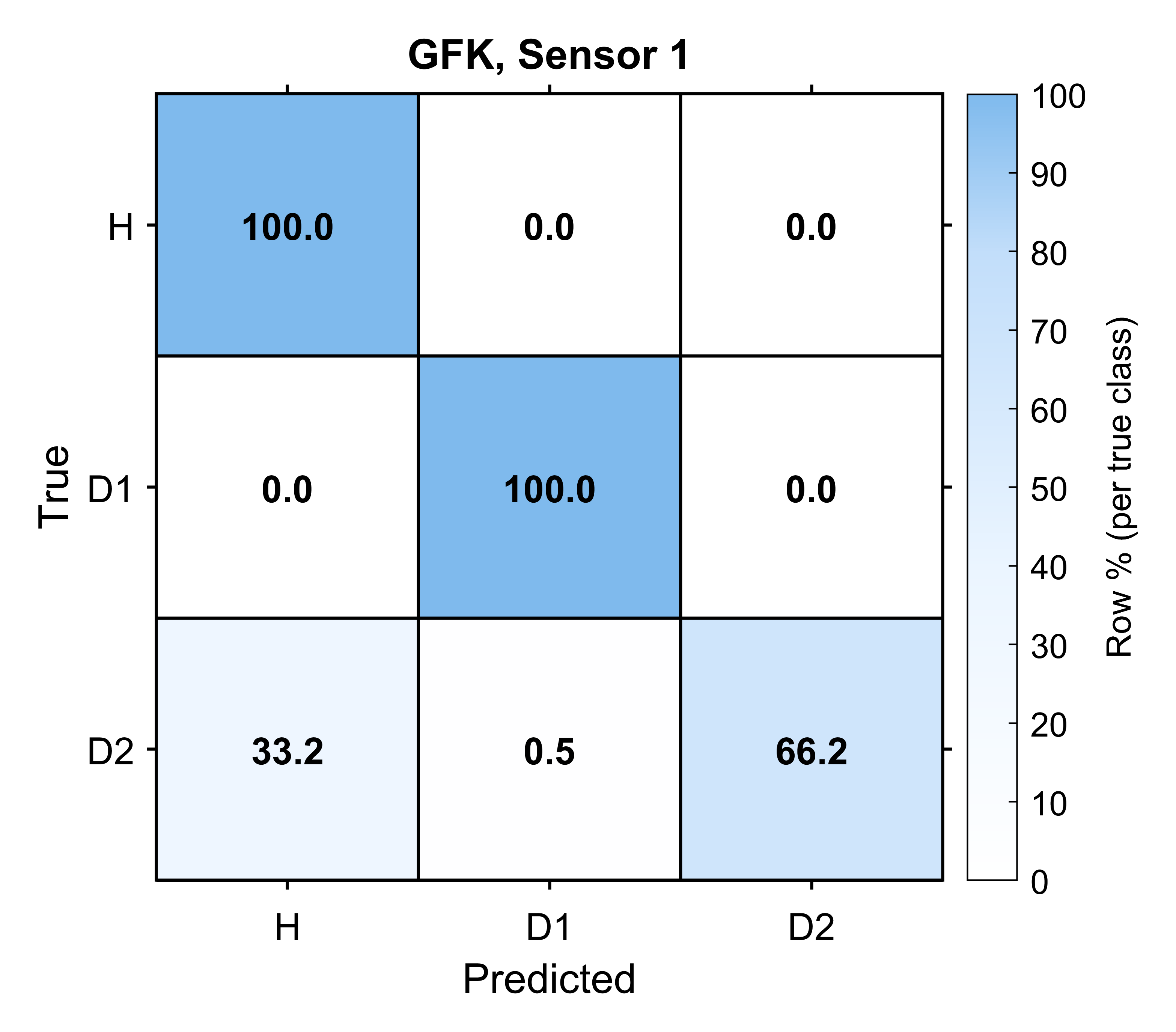}}\\[2pt]
		\subfloat[\label{fig:conf3}]{\includegraphics[width=.45\textwidth]{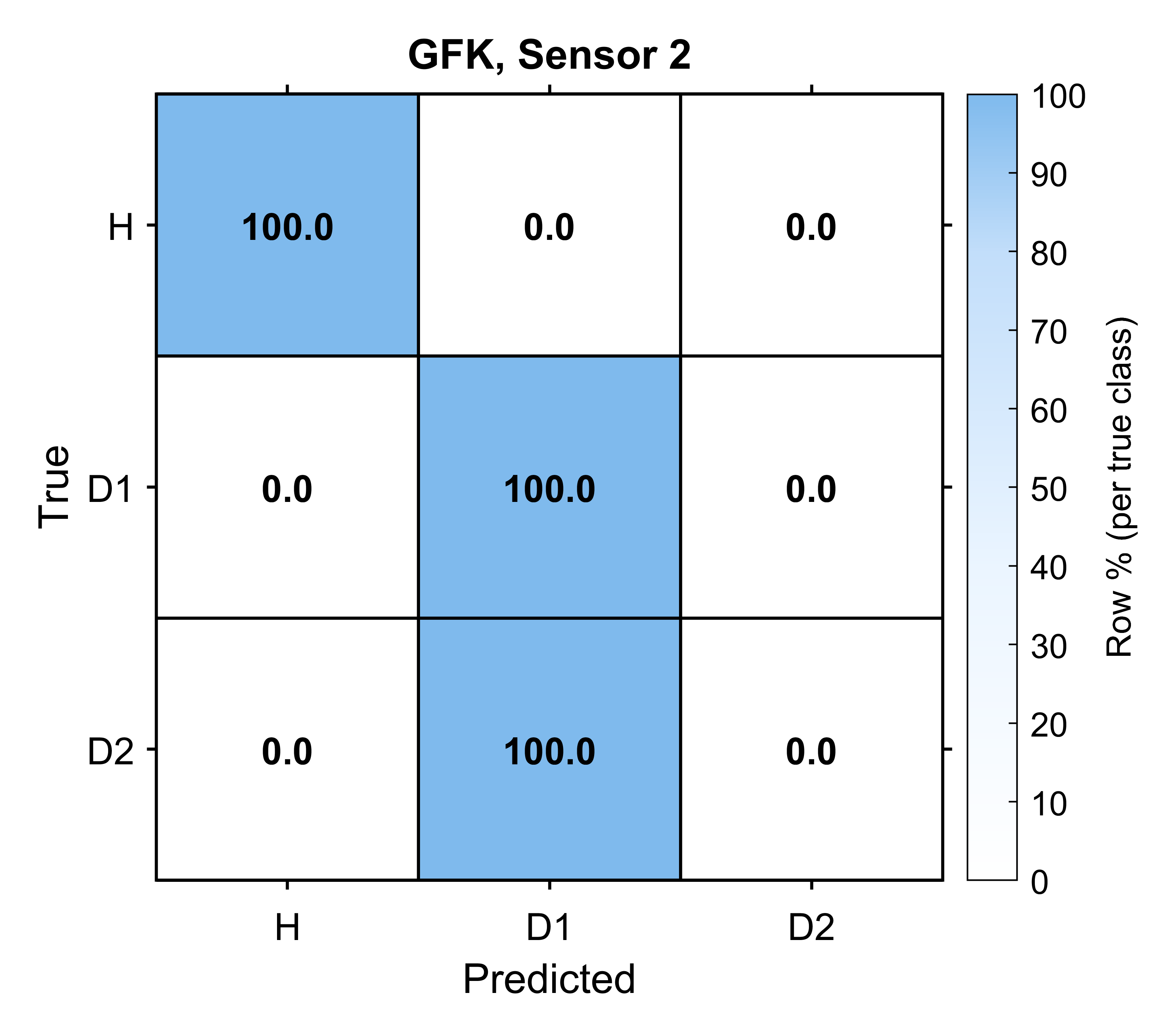}}\hfill
		\subfloat[\label{fig:conf4}]{\includegraphics[width=.45\textwidth]{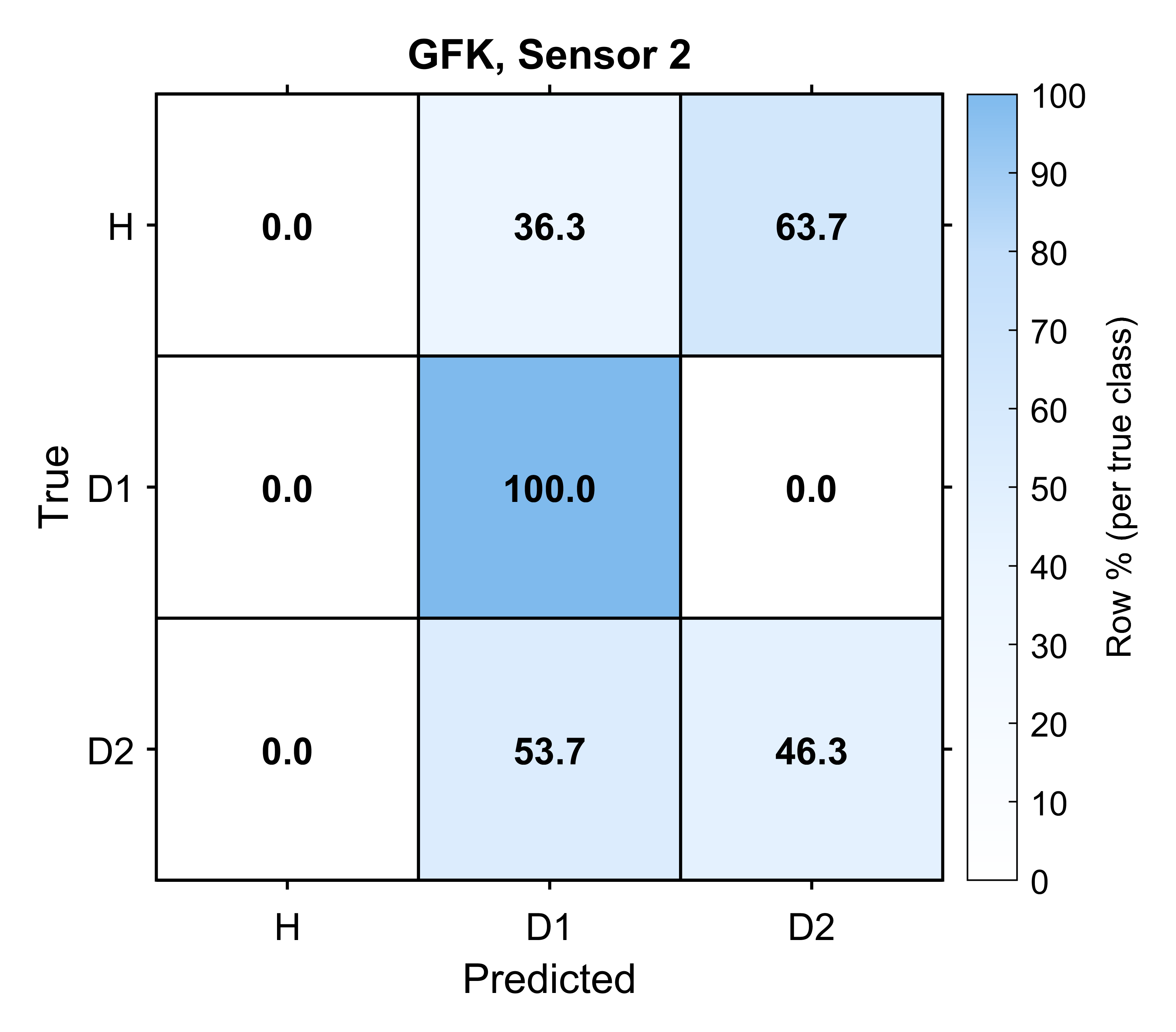}}\\[2pt]
		\subfloat[\label{fig:conf5}]{\includegraphics[width=.45\textwidth]{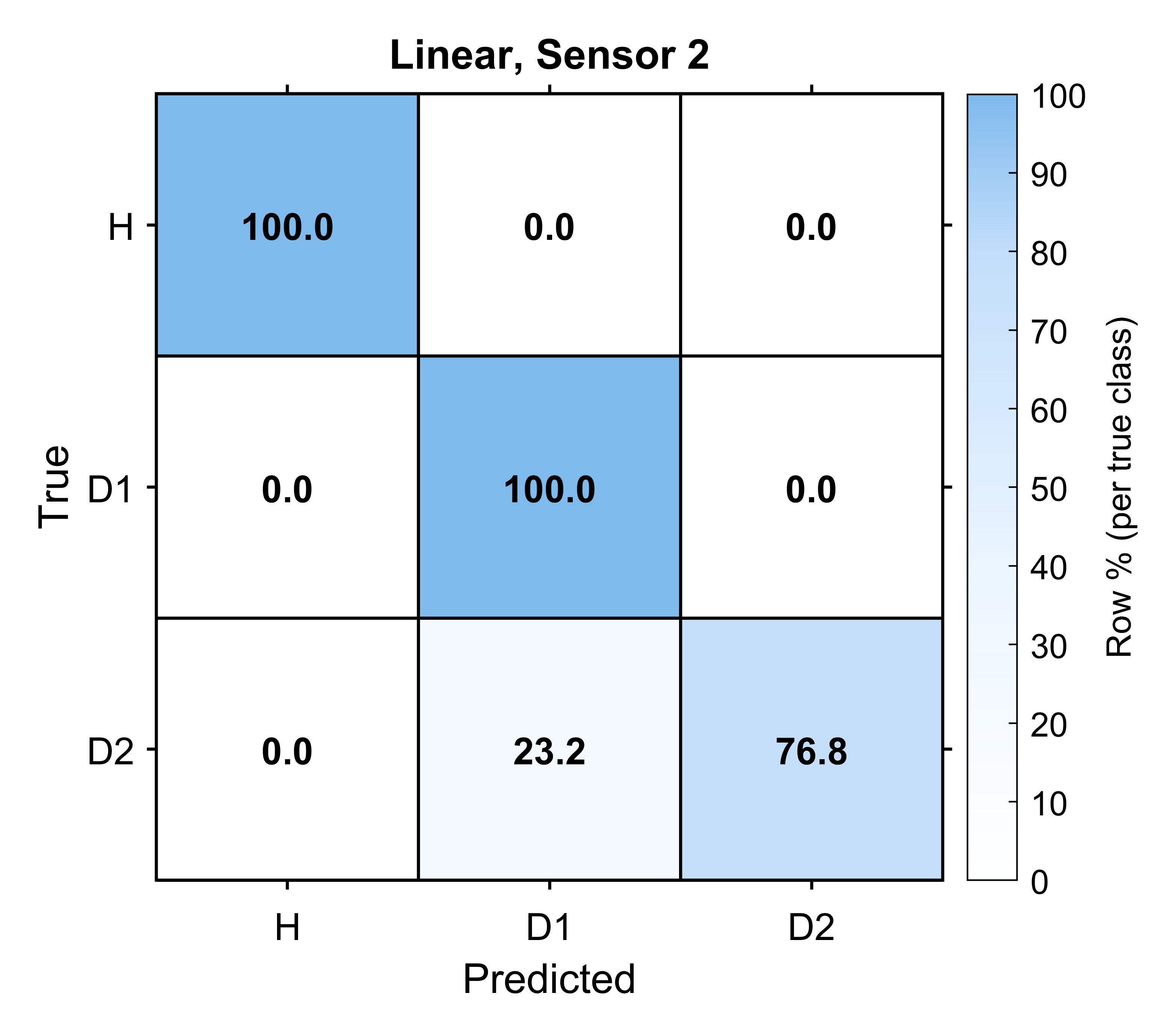}}
		
		\caption{A subset of the confusion matrices for Case~2 transfer configurations. Top row: Sensor~1, (a) GFK with $K=0$ (no intermediates) and (b) $K=9$. Middle row: Sensor~2, (c) GFK with $K=0$ and (d) $K=5$. Bottom: Sensor~2, (e) Linear with the full chain ($K=500$). Matrices are averaged over the corresponding noise realisations reported in Tables~\ref{tab:K0_transfer_case2} to \ref{tab:K500_transfer_case2}.} 
		\label{fig:confmat-case-2}
	\end{figure}

	Subspace dimension was selected from the source healthy data to explain 95\% of the variance and then capped at $d=30$ to keep the subspace size modest for computational efficiency and as mild regularisation. In Case~2 this cap was active (the 95\% threshold typically suggested $d\approx 35$--$45$ per sensor). The same dimension was used for both the linear and GFK approaches. The transfer results are shown in Tables \ref{tab:K0_transfer_case2} to \ref{tab:K500_transfer_case2}. A subset of the confusion matrices averaged over the repetitions are shown in Figure \ref{fig:confmat-case-2}.

	\begin{table}[ht]
		\centering
		\caption{Transfer accuracy at the final hop without intermediates ($K=0$) for Linear and GFK approaches in Case~2. Values are mean $\pm$ standard deviation over 400 noise realisations.}
		\label{tab:K0_transfer_case2}
		\begin{tabular}{c c c}
			\toprule
			Sensor & Linear Kernel (\%) & Geodesic Flow Kernel (\%) \\
			\midrule
			1 & $28.57 \pm 0.00$ & $60.64 \pm 11.90$ \\
			2 & $64.29 \pm 0.00$ & $64.28 \pm 0.05$ \\
			3 & $64.29 \pm 0.00$ & $64.29 \pm 0.00$ \\
			5 & $64.29 \pm 0.00$ & $67.17 \pm 13.74$ \\
			6 & $63.92 \pm 9.41$ & $63.59 \pm 12.51$ \\
			7 & $64.29 \pm 0.00$ & $61.26 \pm 11.89$ \\
			8 & $60.52 \pm 7.61$ & $61.14 \pm 14.64$ \\
			\bottomrule
		\end{tabular}
	\end{table}

	\begin{table}[ht]
		\centering
		\caption{Transfer accuracy at the final hop for Linear and GFK approaches in Case~2 using $K=4$ intermediates: 71, 201, 351, and 500. Values are mean $\pm$ standard deviation over 25 noise realisations.}
		\label{tab:K4_transfer_case2}
		\begin{tabular}{c c c}
			\toprule
			Sensor & Linear Kernel (\%) & Geodesic Flow Kernel (\%) \\
			\midrule
			1 & $35.71 \pm 0.00$ & $35.71 \pm 0.00$ \\
			2 & $35.71 \pm 0.00$ & $100.00 \pm 0.00$ \\
			3 & $100.00 \pm 0.00$ & $100.00 \pm 0.00$ \\
			5 & $100.00 \pm 0.00$ & $100.00 \pm 0.00$ \\
			6 & $64.29 \pm 0.00$ & $100.00 \pm 0.00$ \\
			7 & $64.29 \pm 0.00$ & $100.00 \pm 0.00$ \\
			8 & $64.29 \pm 0.00$ & $100.00 \pm 0.00$ \\
			\bottomrule
		\end{tabular}
	\end{table}

	\begin{table}[ht]
		\centering
		\caption{Transfer accuracy at the final hop for Linear and GFK approaches in Case~2 using $K=5$ intermediates: 71, 201, 251, 351, and 500.  Values are mean $\pm$ standard deviation over 350 noise realisations.}
		\label{tab:K5_transfer_case2}
		\begin{tabular}{c c c}
			\toprule
			Sensor & Linear Kernel (\%) & Geodesic Flow Kernel (\%) \\
			\midrule
			1 & $35.71 \pm 0.00$ & $35.69 \pm 0.38$ \\
			2 & $35.71 \pm 0.00$ & $52.27 \pm 17.82$ \\
			3 & $35.71 \pm 0.00$ & $36.45 \pm 4.53$ \\
			5 & $35.71 \pm 0.00$ & $35.71 \pm 0.00$ \\
			6 & $35.71 \pm 0.00$ & $40.86 \pm 10.99$ \\
			7 & $35.71 \pm 0.00$ & $48.20 \pm 14.19$ \\
			8 & $35.71 \pm 0.00$ & $47.63 \pm 14.11$ \\
			\bottomrule
		\end{tabular}
	\end{table}

	\begin{table}[ht]
		\centering
		\caption{Transfer accuracy at the final hop for Linear and GFK approaches in Case~2 using $K=9$ intermediates: 26, 31, 91, 101, 181, 219, 291, 401, 500. Values are mean $\pm$ standard deviation over 385 noise realisations.}
		\label{tab:K9_transfer_case2}
		\begin{tabular}{c c c}
			\toprule
			Sensor & Linear Kernel (\%) & Geodesic Flow Kernel (\%) \\
			\midrule
			1 & $64.29 \pm 0.00$  & $87.94 \pm 16.91$ \\
			2 & $35.71 \pm 0.00$  & $38.37 \pm 4.33$ \\
			3 & $100.00 \pm 0.00$ & $100.00 \pm 0.00$ \\
			5 & $100.00 \pm 0.00$ & $100.00 \pm 0.00$ \\
			6 & $100.00 \pm 0.00$ & $100.00 \pm 0.00$ \\
			7 & $57.37 \pm 13.86$ & $100.00 \pm 0.00$ \\
			8 & $64.29 \pm 0.00$  & $100.00 \pm 0.00$ \\
			\bottomrule
		\end{tabular}
	\end{table}

	\begin{table}[ht]
		\centering
		\caption{Transfer accuracy at the final hop for Linear and GFK approaches in Case~2 using all intermediates ($K=500$). Values are mean $\pm$ standard deviation over 25 noise realisations.}
		\label{tab:K500_transfer_case2}
		\begin{tabular}{c c c}
			\toprule
			Sensor & Linear Kernel (\%) & Geodesic Flow Kernel (\%) \\
			\midrule
			1 & $100.00 \pm 0.00$ & $100.00 \pm 0.00$ \\
			2 & $91.71 \pm 0.99$  & $100.00 \pm 0.00$ \\
			3 & $100.00 \pm 0.00$ & $100.00 \pm 0.00$ \\
			5 & $100.00 \pm 0.00$ & $100.00 \pm 0.00$ \\
			6 & $100.00 \pm 0.00$ & $100.00 \pm 0.00$ \\
			7 & $100.00 \pm 0.00$ & $100.00 \pm 0.00$ \\
			8 & $100.00 \pm 0.00$ & $100.00 \pm 0.00$ \\
			\bottomrule
		\end{tabular}
	\end{table}

	Across the sensors, transfer without intermediates ($K=0$, Table~\ref{tab:K0_transfer_case2}) was generally poor to moderate, with large variability under GFK for several sensors (e.g., Sensor~1: $60.64\pm11.90$\%, and Sensors~5--8 showing standard deviations of $\approx 12$--$15$\%). Several sensors exhibit zero variance alongside characteristic accuracy values of 28.57\% or 64.29\%. These are consistent with classifier collapse, where all unlabelled samples are assigned to a single class. Given the class distribution of 40 healthy, 50 D1, and 50 D2 unlabelled samples, a classifier that predicts only healthy samples will achieve $40/140 \approx 28.57\%$ accuracy, and a classifier that predicts healthy samples and only one class of damaged samples (D1 or D2) will achieve $90/140 \approx 64.29\%$ accuracy. Note that the labelled healthy samples are excluded from the reported accuracy, so the collapse behaviour is fully determined by the unlabelled class distribution. Introducing a short chain that effectively jumped over the difficult region for the majority of the sensors (Table~\ref{tab:K4_transfer_case2}) improved performance substantially for the GFK approach for Sensors~2, 3, 5, 6, 7, and~8, which reached 100.00\% accuracy. In contrast, inserting an additional intermediate within the identified dip region (the $K=5$ chain including 251; Table~\ref{tab:K5_transfer_case2}) reduced performance across all sensors, with accuracies between around $\approx 28$--$67$\% and increased variability. Overall, these results indicate that transfer performance is sensitive to the placement of intermediates, rather than the number of intermediates alone. 
	
	Sensor~1 behaved differently, consistent with the alignment plots showing multiple sharp drops for this sensor. The $35.71\%$ values seen here are similarly consistent with classifier collapse, corresponding to a classifier predicting only D1 or only D2. The $K=9$ chain (Table~\ref{tab:K9_transfer_case2}) improved Sensor~1 performance substantially relative to the shorter chains, particularly for GFK ($87.94 \pm 16.91$). Finally, using all intermediates (Table~\ref{tab:K500_transfer_case2}) provided the strongest performance overall, achieving 100\% accuracy for all reported sensors under GFK and all but Sensor~2 for the linear approach ($91.71\pm0.99$\%). This is consistent with the idea that increased refinement along difficult parts of the path can facilitate accurate transfer by avoiding large discontinuities between successive intermediates.

	Figure~\ref{fig:confmat-case-2} shows a subset of the confusion matrices for the chains considered. Entries on the diagonal correspond to correct classification (Healthy, D1, and D2), while off-diagonal entries indicate misclassification. For Sensor~1, GFK transfer without intermediates ($K=0$) resulted in substantial misclassification, consistent with the moderate mean accuracy and high variability in Table~\ref{tab:K0_transfer_case2}. In this case the healthy class was predicted accurately (as some healthy data were labelled), while the two damaged datasets were mapped to the same class. Classification at Sensor~1 improved with the $K=9$ chain (this chain was selected in accordance with the cosine alignment plot for that sensor), although some variability and sensitivity to added noise remained.
	
	For Sensor~2, the confusion matrix corresponding to GFK classification with no intermediates ($K=0$) shows that the healthy class was predicted accurately (this is not surprising, as some healthy data were labelled) and the damaged datasets were assumed to belong to a single class. Introducing an intermediate corresponding to a dip in the cosine alignment plot dramatically increased misclassification ($K=5$) compared to the $K=4$ chain, which includes an intermediate within the identified dip region, consistent with the reduced mean accuracy and larger variance in Table~\ref{tab:K5_transfer_case2}. Finally, the confusion matrix corresponding to the linear approach at Sensor~2 using the full set of intermediates ($K=500$) illustrates the remaining, non-zero error for an otherwise near-perfect classification (Table~\ref{tab:K500_transfer_case2}). 
	
	Ongoing work has focussed on several extensions/modifications to the approach. First, chain optimisation will be investigated by considering nonlinear growth schemes for the parameters (as opposed to the linear growth scheme used in the current work), with the goal of keeping consecutive hops close in the feature space and avoiding regions where small parameter changes induce abrupt changes in dynamics. In addition, the sensitivity of the initial and final hops will be examined by quantifying how closely the endpoint FEMs must match the experimental structures for transfer at the chain ends to remain achievable. Second, label-set mismatch will be studied in settings where the taxonomy differs between domains, for example when one domain contains only D1 observations while another contains D1 and D2, or when the definition of the damaged classes is not directly comparable across structures. An interesting approach might be to parameterise the damage itself (with respect to location and severity), so that it evolves along the chain in the same way as the geometry, material properties, and boundary conditions. Third, partial domain adaptation will be explored in scenarios where one or more target classes are missing entirely (i.e., missing data as well as labels). For example, if the target is missing damaged data. In that case, the objective shifts toward aligning normal-condition structure while preserving the ability to detect and separate future damage.

\vspace{12pt} 
\section{Concluding remarks} \label{conclusions}
A novel approach for information transfer between highly-heterogeneous systems was presented. By introducing intermediate domains via parameterisation of the configuration space, a transfer path can be constructed in which consecutive hops are sufficiently similar for accurate transfer to be achieved, with perfect classification demonstrated in both case studies when the chain is appropriately refined. 

In Case 1, the approach was demonstrated on (simulated) two-span and three-span bridges, where the parameterised transfer path involved gradually materialising the second support. This demonstrated that positive transfer could be achieved even when the intermediate structures were not physically realisable. Case 2 demonstrated the approach using an experimental source and final target (a cartoon bridge and an aeroplane) and a chain comprised of finite-element models. PCA-projected FRFs provided the features, and transfer was evaluated independently at seven sensor locations. In both cases, transfer was performed using a linear kernel and SVM and a GFK and SVM. 

It was found that the primary indicator of transfer success was the proximity of adjacent intermediates in the feature space. Provided that the data are separable, deviations from monotonically-decreasing behaviour could be managed by refining the chain in those areas. The GFK was shown to navigate these areas more robustly than the linear kernel, requiring fewer hops to achieve positive transfer. Ongoing work will focus on optimising chain design (including nonlinear parameter growth and quantifying how closely the endpoint FEMs must match the experimental structures for the initial and final hops to be achievable), extending the framework to handle label-set mismatch by parameterising damage location and severity, and evaluating settings where only healthy target data are available during adaptation.

\vspace{12pt} 
\section{Acknowledgements} \label{acknowledgements}
The authors gratefully acknowledge the support of the UK Engineering and Physical Sciences Research Council (EPSRC), via grant reference EP/W005816/\-1. This research made use of The Laboratory for Verification and Validation (LVV), which was funded by the EPSRC (via EP/J013714/1 and EP/N010884/1), the European Regional Development Fund (ERDF), and the University of Sheffield. For the purpose of open access, the authors have applied a Creative Commons Attribution (CC BY) licence to any Author Accepted Manuscript version arising. The authors would like to extend special thanks to Michael Dutchman of the LVV, for fabricating the structures used in the experiments, and Mathew Hall, for assisting with experimental setup. 

\bibliographystyle{elsarticle-num-names} 
\bibliography{references}

\end{document}